\documentclass{article}
\usepackage[final]{neurips_2019}

\usepackage{graphicx}
\usepackage{natbib}
\RequirePackage{fix-cm}

\usepackage{algorithmicx}
\usepackage{algpseudocode}
\usepackage{bm}
\usepackage{enumitem}
\usepackage{multirow}
\usepackage{amssymb}
\usepackage{amssymb}
\usepackage{array}
\usepackage{listings}
\usepackage{xspace}
\usepackage{mathtools}
\usepackage{todonotes}
\usepackage{float}
\usepackage{booktabs}
\usepackage{hyperref}
\usepackage{caption}
\usepackage{subcaption}
\usepackage{dsfont}
\usepackage[section]{placeins}

\mathtoolsset{showonlyrefs=true}

\RequirePackage{natbib}
\setcitestyle{square,aysep={},yysep={;}} 

\definecolor{codegreen}{rgb}{0,0.6,0}
\definecolor{codegray}{rgb}{0.5,0.5,0.5}
\definecolor{codepurple}{rgb}{0.58,0,0.82}
\definecolor{backcolour}{rgb}{0.95,0.95,0.92}

\DeclareMathOperator*{\argmax}{arg\,max}

\def\S{\mathcal{S}}
\def\A{\mathcal{A}}

\newcommand{\suite}{\texttt{realworldrl-suite}\xspace}
\newcommand{\mujoco}{\texttt{MuJoCo}\xspace}
\newcommand{\cartpoleswingup}{\texttt{cartpole:swingup}\xspace}
\newcommand{\walkerwalk}{\texttt{walker:walk}\xspace}
\newcommand{\quadrupedwalk}{\texttt{quadruped:walk}\xspace}
\newcommand{\humanoidwalk}{\texttt{humanoid:walk}\xspace}

\lstdefinestyle{mystyle}{
    backgroundcolor=\color{backcolour},   
    commentstyle=\color{codegreen},
    keywordstyle=\color{magenta},
    numberstyle=\tiny\color{codegray},
    stringstyle=\color{codepurple},
    basicstyle=\ttfamily\footnotesize,
    breakatwhitespace=false,         
    breaklines=true,                 
    captionpos=b,                    
    keepspaces=true,                 
    numbers=left,                    
    numbersep=5pt,                  
    showspaces=false,                
    showstringspaces=false,
    showtabs=false,                  
    tabsize=2
}

\title{An empirical investigation of the challenges of real-world reinforcement learning}

\author{
  Gabriel Dulac-Arnold$^{1,}$\thanks{equally contributed, $^{1}$Google Research, Paris, $^{2}$Deepmind, London, $^{3}$Deepmind, Mountainview, $^{4}$Work done during time at Deepmind}\\
  \texttt{dulacarnold@google.com} \\
  \And
  Nir Levine$^{3,*}$\\
  \texttt{nirlevine@google.com} \\
  \And
  Daniel J. Mankowitz$^{2,*}$\\
  \texttt{dmankowitz@google.com} \\
  \And
  Jerry Li$^{2}$\\
  \And
  Cosmin Paduraru$^{2}$\\
  \And
  Sven Gowal$^{2}$\\
  \And
  Todd Hester$^{4}$\\
  }

\protect
\begin{document}

% \protect\title{An empirical investigation of the challenges of real-world reinforcement learning}

% \protect\author{Gabriel Dulac-Arnold\protect\footnote{equally contributed} \protect\and Nir Levine$^*$ \protect\and Daniel J. Mankowitz$^*$ \protect\and Jerry Li \protect\and Cosmin Paduraru \protect\and Sven Gowal \protect\and Todd Hester}
% \protect\institute{G. Dulac-Arnold \protect\email{dulacarnold@google.com} \protect\at Google Research \protect\and N. Levine \protect\email{nirlevine@google.com}, D. J. Mankowitz \protect\email{dmankowitz@google.com}, J. Li, C. Paduraru, S. Gowal, T. Hester \protect\at DeepMind
% }

% \protect\date{}
% % The correct dates will be entered by the editor

\maketitle

% \protect\maketitle

\protect\begin{abstract}
Reinforcement learning (RL) has proven its worth in a series of artificial domains, and is beginning to show some successes in real-world scenarios.  However, much of the research advances in RL are hard to leverage in real-world systems due to a series of assumptions that are rarely satisfied in practice. 
  In this work, we identify and formalize a series of independent challenges that embody the difficulties that must be addressed for RL to be commonly deployed in real-world systems. 
  For each challenge, we define it formally in the context of a Markov Decision Process, analyze the effects of the challenge on state-of-the-art learning algorithms, and present some existing attempts at tackling it.
  We believe that an approach that addresses our set of proposed challenges would be readily deployable in a large number of real world problems. 
  Our proposed challenges are implemented in a suite of continuous control environments called \suite which we propose an as an open-source benchmark. 
\protect\end{abstract}

% \section{Potential framing}

% Intro:
% Our previous paper setup these challenges... repeat some of the motivation. 
% We want to do two things:
% - validate that these challenges are a problem for existing SOTA methods
% - release a task suite to enable further research and development on these challenges.

% some text on why we're focused on this subset of the challenges rather than the whole 9.

% Task setup:
% explain the task suite, why we chose these tasks, general principle around wrapping them to add the different challenges so we can easily add/remove challenges and vary them from easy to difficult. 

% challenge sections:
% For each challenge, give the explanation and related work as in the previous paper. Then explain how we have implemented the challenge as an environment wrapper. Then show the baseline results that this challenge does cause problems for sota algorithms. give some ideas of the obvious next things we could do (recurrent models, lagrange multipliers, etc.)

% ? should we have a section summarizing the other challenges?

% conclusion: again re-iterate that each challenge causes problems for sota methods. Also show a plot with results when all the challenges are turned on at once. 

\section{Introduction}

\label{intro}

Reinforcement learning (RL) \citep{sutton2018reinforcement} is a powerful algorithmic paradigm encompassing a wide array of contemporary algorithmic approaches \citep{mnih2015human, silver2016mastering, hafner2018learning}.  RL methods have been shown to be effective on a large set of simulated environments \citep{mnih2015human, silver2016mastering, lillicrap2015continuous, OpenAI_dota}, but uptake in real-world problems has been much slower.  We posit that this is primarily due to a large gap between the casting of current experimental RL setups and the generally poorly defined realities of real-world systems.

We are inspired by a large range of real-world tasks, from control systems grounded in the physical world~\citep{vecerik2019practical, kalashnikov2018qt} to global-scale software systems interacting with billions of users~\citep{gauci2018horizon, covington2016deep, ie2019recsim}. 

Physical systems can range in size from a small drone~\citep{abbeel2010autonomous} to a data center~\citep{DM_Datacenter}, in complexity from a one-dimensional thermostat~\citep{hester2018predictively} to a self-driving car, and in cost from a calculator to a spaceship. 
Software systems range from billion-user recommender systems~\citep{covington2016deep} to on-device controllers for individual smart-phones, they can be scheduling millions of software jobs across the globe or optimizing the battery profile of a single device, and the codebase might be millions of lines of code to a simple kernel module.
In all these scenarios, there are recurring themes: the systems have inherent latencies, noise, and non-stationarities that make them hard to predict.  They may have large and complicated state \& action spaces, safety constraints with significant consequences, and large operational costs both in terms of money and time.
This is in contrast to training on a perfect simulated environment where an agent has full visibility of the system, zero latency, no consequences for bad action choices and often deterministic system dynamics.

We posit that these difficulties can be well summarized by a set of nine challenges that are  holding back RL from real-world use. At a high level these challenges are:
\begin{enumerate}
    \item \label{ch1} Being able to learn on live systems from limited samples.
    \item \label{ch2} Dealing with unknown and potentially large delays in the system actuators, sensors, or rewards.
    \item \label{ch3} Learning and acting in high-dimensional state and action spaces. 
    \item \label{ch4} Reasoning about system constraints that should  never or rarely be violated.
    \item \label{ch5} Interacting with systems that are partially observable, which can alternatively be viewed as systems that are non-stationary or stochastic.
    \item \label{ch6} Learning from multi-objective or poorly specified reward functions.
    \item \label{ch7} Being able to provide actions quickly, especially for systems requiring low latencies.
    \item \label{ch8} Training off-line from the fixed logs of an external behavior policy.
    \item \label{ch9} Providing system operators with explainable policies.
\end{enumerate}

\subsection{Illustrative Examples}
These challenges can present themselves in a wide array of task scenarios.  We choose three examples, from robotics, healthcare and software systems to illustrate how these challenges can manifest themselves in various ways.  

A common robotic challenge is autonomous manipulation, and has potential applications ranging from manufacturing to healthcare.  Such a robotic system is affected by nearly all of the proposed challenges. 
% Robot time is often costly and therefore learning must be data-efficient (Challenge \ref{ch1}). A robot's actuators and sensor introduce varying amounts of delay, and the task reward might be delayed relative to the system's state (Challenge \ref{ch2}).  
% Robotic systems almost always have some form of constraints either in their movement space, or on their joints in terms of velocity and acceleration constraints (Challenge \ref{ch4}).
% As the system manipulates the space around it, things will react in unexpected, stochastic ways, and the system will not be fully observable (Challenge \ref{ch5}).  
% System operators may want to optimize for a certain performance on the task, but also want to encourage fast operation, energy efficiency, and reduce wear \& tear (Challenge \ref{ch6}).  
% Response time necessary for good control requires real-time control and fast response times (Challenge \ref{ch7}).
% There are likely logs of the system operating either through tele-operation, or simpler black-box controllers, both of which could be leveraged to learn offline without costing system time (Challenge \ref{ch8}).

\begin{itemize}
\item Robot time is costly and therefore learning should be data-efficient (Challenge \ref{ch1}).
\item Actuators and sensor introduce varying amounts of delay, and the task reward can be delayed relative to the system state (Challenge \ref{ch2}).  
\item Robotic systems almost always have some form of constraints either in their movement space, or directly on their joints in terms of velocity and acceleration constraints (Challenge \ref{ch4}).
\item As the system manipulates the space around it, things will react in unexpected, stochastic ways, and the robot's environment will not be fully observable (Challenge \ref{ch5}).  
\item System operators may want to optimize for a certain performance on the task, but also want to encourage fast operation, energy efficiency, and reduce wear \& tear (Challenge \ref{ch6}).  
\item A performant controller requires low latency for both smooth and safe control (Challenge \ref{ch7}).
\item There are generally logs of the system operating either through tele-operation, or simpler black-box controllers, both of which can be leveraged to learn offline without costing system time (Challenge \ref{ch8}).
\end{itemize}

In the case of a healthcare application, we can imagine a policy for assisted diagnostic that is trained from electronic health records (EHRs).  This policy could work hand-in-hand with doctors to help in treatment approaches, and would be presented with many of our described challenges:
\begin{itemize}
    \item EHR data is not necessarily plentiful, and therefore learning from limited samples is essential to finding good policies from the available data (Challenge 
    \ref{ch1}).
    \item The effects of a particular treatment may be observable hours to months after it takes place.  These strong delays will likely pose a challenge to any current RL algorithms (Challenge 2).
    \item Certain constraints, such as dosage strength or patient-specific allergies, must be respected to provide pertinent treatment strategies (Challenge 4).
    \item Biological systems are inherently complex, and both observations as well as patient reactions are inherently stochastic (Challenge \ref{ch5}).
    \item Many treatment approaches balance aggressivity towards a pathology with sensitivity to the patients' reaction.  Along with other constraints such as time and drug availability, these problems are often multi-objective (Challenge \ref{ch6}).
    \item EHR data is naturally off-line, and therefore being able to leverage as much information from the data before interacting with patients is essential (Challenge \ref{ch7}).
    \item For successful collaboration between an algorithm and medical professionals, explainability is essential.  Understanding the policy's long-term intended goals is essential in deciding which strategy to take (Challenge \ref{ch9}).
\end{itemize}

Recommender systems are amongst the most solicited large-scale software systems, and RL proposes an enticing framework for optimizing them \citep{covington2016deep, chen2019top}.  However, there are many difficulties to be dealt with in large user-facing software systems such as these:
\begin{itemize}
    \item Interactions with the user can be strongly delayed, either from users reacting to recommendations with high latency, or recommendations being sent to users at different points in time (Challenge \ref{ch2}).
    \item The set of possible actions is generally very large (millions to even potentially billions), which becomes particularly difficult when reasoning about action selection (Challenge \ref{ch3}).
    \item Many aspects of the user's interactions with the system are unobserved: Does the user see the recommendation? What is a user currently thinking? Does the user choose not to engage due to poor recommendations? (Challenge \ref{ch5})
    \item Optimization goals are often multi-objective, with recommender systems trying to increase engagement, all while driving revenue, reducing costs, maintaining diversity and ensuring fairness (Challenge \ref{ch6}).
    \item Many of these systems interact in real-time with a user, and need to provide recommendations within milliseconds (Challenge \ref{ch7}).
    \item Although some degree of experimentation is possible on-line, large amounts of information are available in the form of interaction logs with the system, and need to be exploited in an off-line manner (Challenge \ref{ch8}).
    \item Finally, as a recommender system has a potential to significantly affect the user's experience on the platform, its choices need to be easily understandable and interpretable (Challenge \ref{ch9}).
\end{itemize}

This set of examples shows that the proposed challenges appear in varied types of applications, and we believe that by identifying, replicating and solving these challenges, reinforcement learning can be more readily used to solve many of these important real-world problems.
% In this paper, we introduce a real-world RL benchmarking suite (\suite) that incorporates a subset of the above challenges into a set of control environments. We focus primarily on challenges which are direct modifications to the Markov Decision Process (MDP) environment \cite{sutton2018reinforcement}, rather than the experimental setup itself. As a result, we focus on the first \textbf{six} challenges and plan to investigate challenges related to off-line learning from logs (7), explainability (8), and real-time interaction (9) in future work.

\subsection{Contributions}
This paper presents four main contributions:
\begin{itemize}
    \item \textbf{Identification and definition of real-world challenges:} Our main goal is to more clearly define the issues reinforcement learning is having when dealing with real systems. 
    By making these problems identifiable and well-defined, we hope they can be dealt with more explicitly, and thus solved more rapidly.  We structure the difficulties of real-world systems in the aforementioned 9 challenges.  For each of the above challenges, we provide some intuition on where it arises and discuss potential solutions present in the literature.
    
    \item \textbf{Experiment design and analysis for each challenge:}
    For all challenges except explainability, we provide a formal definition of the challenge and implement a set of environments exhibiting this challenge's characteristics. This allows researchers to easily observe the effects of this challenge on various algorithms, and evaluate if certain approaches seem promissing in dealing with the given challeng.  To both illustrate the extent of each challenge's difficulty, and provide some reference results, we train two state-of-the-art RL agents on each defined environment, with varying degrees of difficulty, and analyze the challenge's effects on learning.  With these analyses we provide insights as to which challenges are more difficult and propose calibrated parameters for each challenge implementation. 
     
    \item \textbf{Define and baseline RWRL Combined Challenge Benchmark tasks:} After careful calibration, we combine a subset of our proposed challenges into a single environment and baseline the performance of two state-of-the-art learning agents on this setup in Section \ref{sec:combined_challenges}. We show that state-of-the-art agents fail quickly, even for mild perturbations applied along each challenge dimension. We encourage the community to work on improving upon the combined challenges' baseline performance.  We believe that in doing so, we will take large steps towards developing agents that are implementable on real world systems.
    
    \item \textbf{Open-source \suite codebase:}  We present the set of perturbed environments in a parametrizable suite, called \suite which extends the DeepMind Control Suite \citep{tassa2018deepmind} with various perturbations representing the aforementioned challenges. The goal of the suite is to accelerate research in these areas by enabling RL practitioners and researchers to quickly, in a principled and reproducible fashion, test their learning algorithms on challenges that are encountered in many real-world systems and settings.  The \suite is available for download here: \url{https://github.com/google-research/realworldrl_suite}.
    A user manual, found in Appendix \ref{app:codebase}, explains how to instantiate each challenge and also provides code examples for training an agent.

    % \item \textbf{Baseline combined challenges:}
    % We also provide experiments that combine all challenges together. The hyperparameters for each combined challenge can be found in Appendix~\ref{app:parameters}, Table~\ref{app:challengehyperparams}. Pre-defined configurations are available in the codebase to run agents on each combined challenge. These challenges are termed ``Easy'', ``Medium'' and ``Hard'', and correspond to hyperparameter settings in increasing order of difficulty. We baseline the performance of state-of-the-art learning algorithms in these combined challenges. We believe that solving these combined challenges will provide a meaningful step forward towards realizing RL in real-world applications.

\end{itemize}

\section{Analysis of the Real-World Challenges}

In this section, for each of the challenges presented in the introduction we discuss its importance and present current research directions that attempt to tackle the challenge, providing starting points for practitioners and newcomers to the domain. We then define it more formally, and analyse its effects on state-of-the-art learning algorithms using the \suite, to provide insights on how these challenges manifest themselves in isolation.
%An in-depth overview of the techniques mentioned in each challenge can be found in Appendix \ref{app:related_work}. 
While not all of these challenges are present together in every real system, for many systems they are all present together to some degree.  For this reason, in Section \ref{sec:combined_challenges} we also present a set of combined reference challenges, varying in difficulty, that emulate a complete system with all of the introduced challenges.  We believe that a learner able to tackle these combined challenges would be a good candidate for many real-world systems.

\paragraph{Notation}
Environments are formalised as Markov Decision Processes (MDPs). A MDP can be defined as a tuple $\langle \S, \A , p, r, \gamma \rangle$, where an agent is in a state $s_t \in \S$ and takes an action $a_t \in \A$ at timestep $t$.  When in state $s_t$ and taking an action $a_t$, an agent will arrive in a new state $s_{t+1}$ with probability $p(s_{t+1} | s_t, a_t)$, and receive a reward $r(s_t, a_t, s_{t+1})$.  Our environments are episodic, which is to say that they last a finite number of timesteps, $1 \leq t \leq T$. The value of $\gamma$, the discount factor, reflects the agent's planning horizon. The full state of the process, $s_t$, respects the Markov property: $p(s_{t+1} | s_t, a_t, \cdots, s_0, a_0) = p(s_{t+1} | s_t, a_t)$, i.e. all necessary information to predict $s_{t+1}$ is contained in $s_t$ and $a_t$.  In many of the environments in this paper the \textit{observed} state does not include the full internal state of the \mujoco physics simulator.  It has nevertheless been shown empirically that the observed state is sufficient to control an agent, so we interchange the notion of state and observation unless otherwise specified.

Ultimately, the goal of a RL agent is to find an optimal policy $\pi^*: \S \rightarrow \A$ which maximizes its expected return over a given MDP: $$\pi^* = \argmax_\pi \mathbb{E^\pi}\left[\sum_{t=0}^{\infty} \gamma^t r(s_t, \pi(s_t), s_{t+1} \sim p(s_t, \pi(s_t)))\right]$$

There are many ways to find this policy~\citep{sutton2018reinforcement}, and we will use two \textit{model-free} methods described in the following section.
% $$J(\pi)=\inf_{p \in \mathcal{P}}\mathbb{E}^p \biggl[\sum_{t=0}^\infty \gamma^t r_t \vert \mathcal{P}, \pi \biggr].$$

\paragraph{Learning algorithms:} For each challenge, we present the results of two state-of-the-art (SOTA) RL learning algorithms: Distributional Maximum a Posteriori Policy Optimization  (DMPO)~\citep{Abdolmaleki2018} and Distributed Distributional Deterministic Policy Gradient (D4PG)~\citep{barthmaron2018d4pg}. We chose these two algorithms for benchmarking performance as they  (1) yield SOTA performance on the dm-control suite (see e.g., \cite{hoffman2020acme,barthmaron2018d4pg}); (2) they are both fundamentally different algorithms (DMPO is an EM-style policy iteration algorithm with a stochastic policy and D4PG is a deterministic policy gradient algorithm). Note that we also tested the original non-distributional algorithm MPO and found performance to be similar to DMPO. As such we did not include the results. It was important that our algorithms were both strong in terms of performance and diverse in terms of algorithmic implementation to show that SOTA algorithms struggle on many of the challenges that we present in the paper. We could have included more algorithms such as SAC and PPO. However, we felt that the environmental cost of running thousands of additional experiments would not justify the additional insights gained. One of our main motivations in this work is to show that SOTA algorithms do suffer from these challenges to encourage more research on these topics.

D4PG is a modified version of Deep Deterministic Policy Gradients (DDPG)~\citep{lillicrap2015continuous}, an actor-critic algorithm where state-action values are estimated by a critic network, and the actor network is updated with gradients sampled from the critic network. D4PG makes four changes to improve the critic estimation (and thus the policy): evaluating $n$-step rather than 1-step returns, performing a \textit{distributional} critic update~\citep{BellemareDM17}, using prioritized sampling of the replay buffer, and performing distributed training. These improvements give D4PG state of the art results across many DeepMind control suite~\citep{tassa2018deepmind} tasks as well as manipulation and parkour tasks~\citep{heess2017emergence}. The hyperparameters for D4PG can be found in Appendix \ref{app:algorithms}, Table \ref{table:d4pg_hyperparameters}.  

MPO~\citep{abbas2018mpo} is an RL method that combines the sample efficiency of off-policy methods with the scalability and hyperparameter robustness of on-policy methods. It is an EM style method, which alternates an E-step that re-weights state-action samples with an M step that updates a deep neural network with supervised training. MPO achieves state of the art results on many continuous control tasks while using an order of magnitude fewer samples when compared with PPO~\citep{Schulman2017ppo}. Distributional MPO (DMPO) is an extension of MPO that uses a distributional value function and achieves superior performance. The hyperparameters for DMPO can be found in Appendix \ref{app:algorithms}, Table \ref{table:mpo_hyperparameters}. The hyperparameters were found by doing a grid-search on each algorithm, based on parameters used in the original papers. The algorithms achieved optimal reported performance in each case using these parameters in the `no challenge' setting (i.e., when none of the challenges are present in the environment).

Each algorithm is run for $30$K episodes on $5$ different seeds on \cartpoleswingup, \walkerwalk, \quadrupedwalk and \humanoidwalk tasks from the \suite. Unless stated otherwise, the mean value reported in each graph is the mean performance of the last $100$ episodes of training with the corresponding standard deviation. All hyperparameters for all experiments can be found in Table~\ref{app1:hyperparameters_sweeps}. To make experiments more easily reproducible we did not use distributed training for either D4PG or DMPO.  Additionally, unless otherwise noted, evaluation is performed on the same policy as used for training, to be consistent with the notion that there is no train/eval dichotomy. We refer to average reward and average return interchangeably in this paper.

\subsection{Challenge 1: Learning On the Real System from Limited Samples}

\paragraph{Motivation \& Related Work} Almost all real-world systems are either slow-moving, fragile, or expensive enough to operate, that data they produce is costly and therefore learning algorithms must be as data-efficient as possible. 
Unlike much of the research performed in RL~\citep{mnih2015human,Espeholt2018,dqfd,Tessler2017}, real systems do not have separate training and evaluation environments, therefore the agent must quickly learn to act reasonably and safely.
In the case where there are off-line logs of the system, these might not contain anywhere near the amount of data or data coverage that current RL algorithms expect. In addition, as all training data comes from the real system, learning agents cannot have an overly aggressive exploration policy during training, as these exploratory actions are rarely without consequence.  This results in training data that is low-variance with very little of the state and action space being covered.  

Learning iterations on a real system can take a long time, as slower systems' control frequencies can range from hours in industrial settings, to multiple months in cases with infrequent user interactions such as healthcare or advertisement. Even in the case of higher-frequency control tasks, the learning algorithm needs to learn \textit{quickly} from potential mistakes without having to repeat them multiple times.   In addition, since there is often only one instance of the system, approaches that instantiate hundreds or thousands of environments to accelerate training through distributed training~\citep{whoops, impala, adamski2018distributed} nevertheless require as much data and are rarely compatible with real systems.  For all these reasons, learning on a real system requires an algorithm to be both sample-efficient and quickly performant.

There are a number of related works that deal with RL on real systems and, in particular, focus on sample efficiency. One body of work is Model Agnostic Meta-Learning (MAML)~\citep{finn2017model}, which focuses on learning within a task distribution and, with few-shot learning, quickly adapting to solving a new in-distribution task that it has not seen previously. Bootstrap DQN~\citep{osband16} learns an ensemble of Q-networks and uses Thompson Sampling to drive exploration and improve sample efficiency. Another approach to improving sample efficiency is to use expert demonstrations to bootstrap the agent, rather than learning from scratch. This approach has been combined with DQN~\citep{mnih2015human} and demonstrated on Atari~\citep{dqfd}, as well as combined with DDPG~\citep{lillicrap2015continuous} for insertion tasks on robots~\citep{insertion}. Recent Model-based deep RL approaches~\citep{hafner2018learning, chua2018deep, nagabandi2019deep}, where the algorithm plans against a learned transition model of the environment, show a lot of promise for improving sample efficiency. \cite{haarnoja2018soft} introduce soft actor-critic algorithms which achieve state-of-the-art performance in terms of sample efficiency and asymptotic performance. \cite{riedmiller2018learning} propose Schedule Auxiliary Control (SAC-X) that enables an agent to learn complex behaviours from scratch using multiple sparse reward signals. This leads to efficient exploration which is important for sparse reward RL. \cite{levine2013guided} use trajectory optimization to direct policy learning and avoid poor local optima. This leads to sample efficient learning that significantly outperforms the state of the art. \cite{yahya2017collective} build on this work to perform distributed learning with multiple real-world robots to achieve better sample efficiency and generalization performance on a door opening task using four robots. Another common approach is to learn ensembles of transition models and use various sampling strategies from those models to drive exploration and improve sample efficiency~\citep{MLJ12-hester,chua2018deep,buckman2018}.

\begin{figure}
    \begin{center} 
    \begin{subfigure}{.45\textwidth}
        % \centering
  \includegraphics[width=\textwidth]{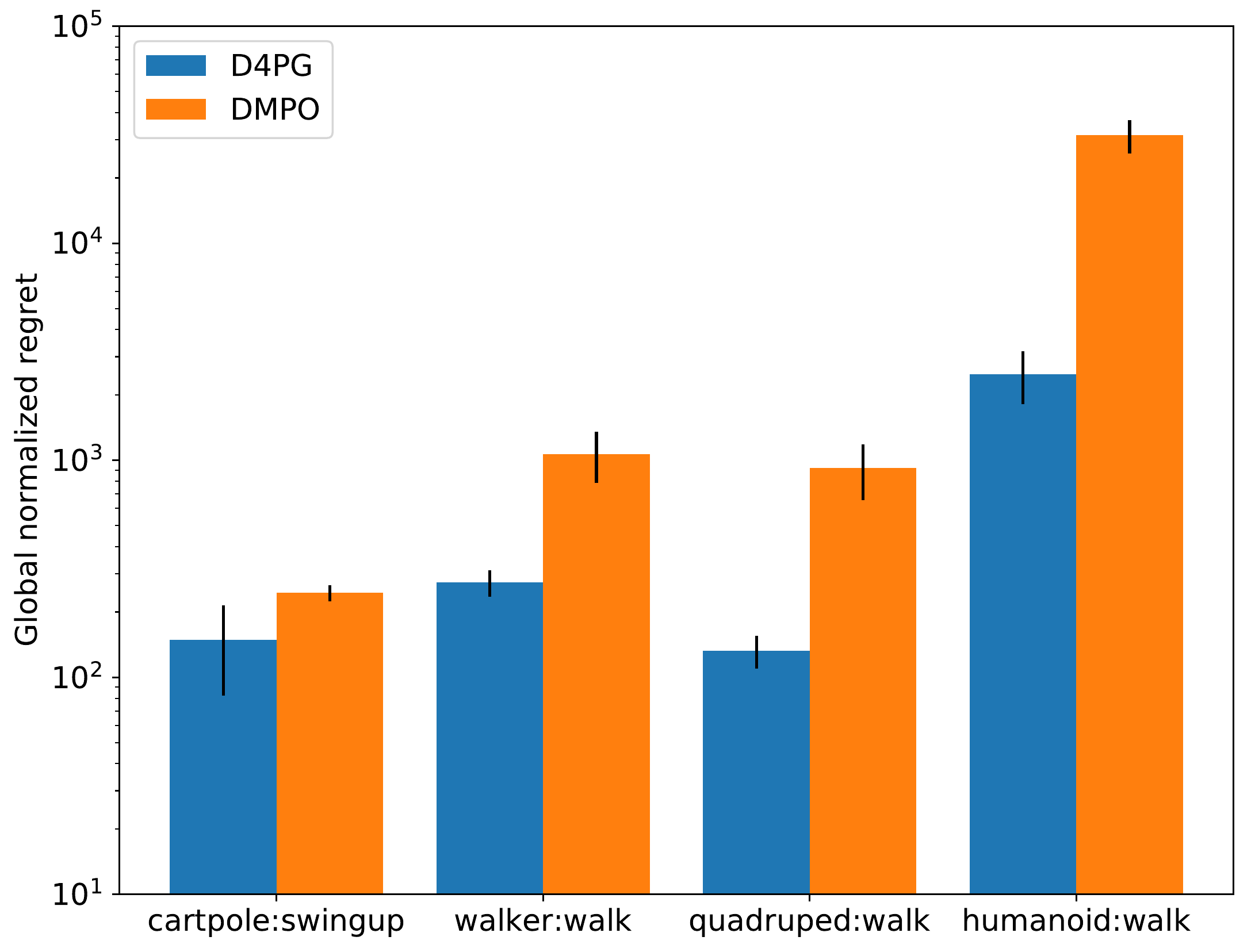}
  \caption{Pre-convergence regret}\label{fig:regret}
\end{subfigure}
\begin{subfigure}{.45\textwidth}
        % \centering
  \includegraphics[width=\textwidth]{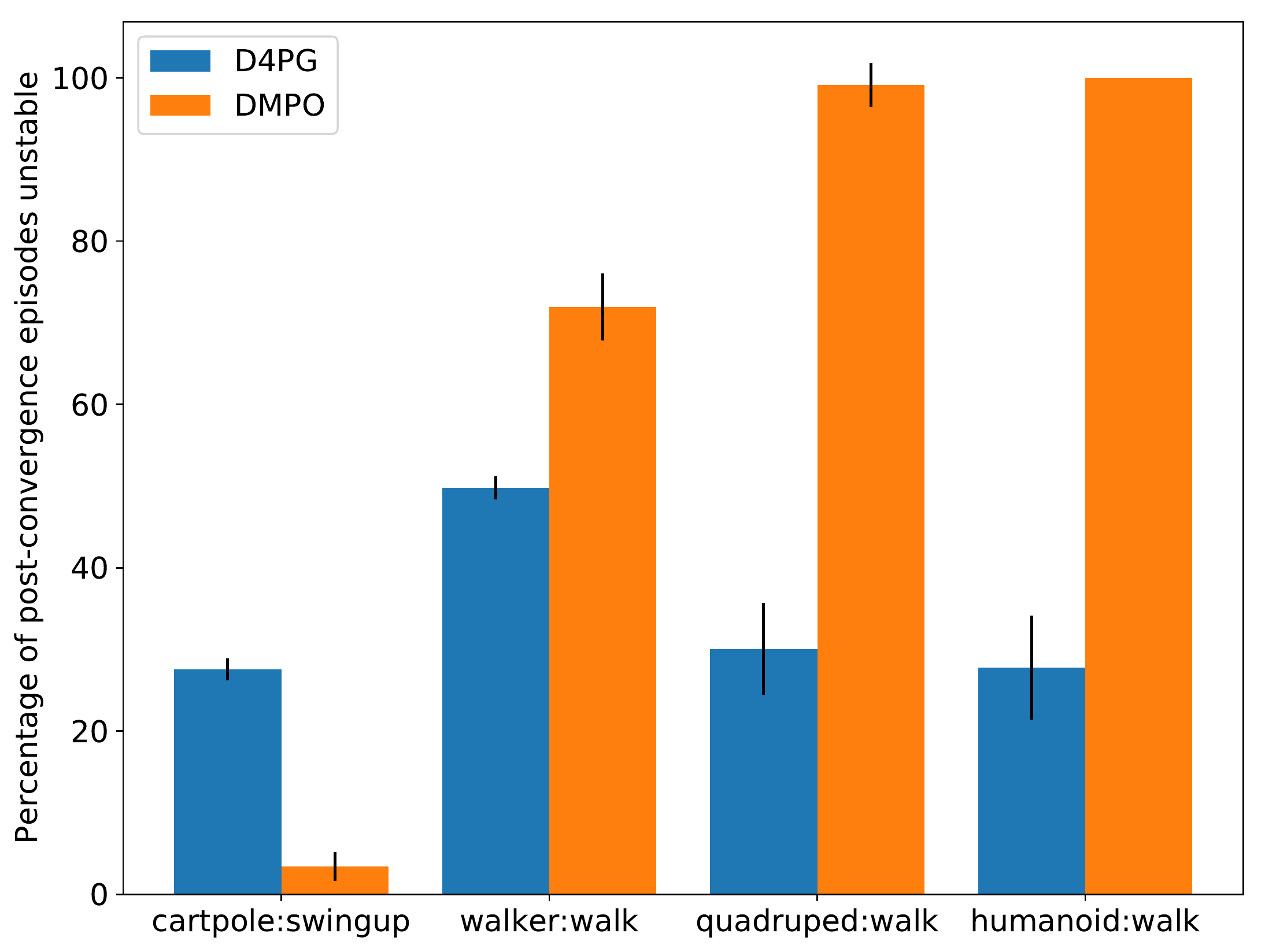}
  \caption{Post-convergence instability}\label{fig:stability}
  \end{subfigure}
 \caption{Sample efficiency metrics. (a) Pre-convergence global normalized regret measures how much the total reward is lost before convergence to the level of final performance reached by the best policy for that task. This is normalized by the average episodic return for the best policy. (b) Post-convergence stability measures what percentage of episodes are suboptimal after convergence. If an algorithm never converges this is measured using the last $window\_size$ episodes, where $window\_size$ is the size of the sliding window used for determining convergence.} % OLD: Assuming the final policy has an average return of $R_{mean}$ with a 95\% confidence interval $[R_{lower}, R_{upper}]$, and that the agent first reaches a return of $R_{lower}$ in episode k, the normalized regret is $\frac{1}{R_{mean}} \sum_{i=0}^K L_i$, where $L_i$ is the regret for episode $i$ with respect to the final policy performance.}
 \label{fig:regret_and_stability}
 \end{center}
\end{figure}

\paragraph{Experimental Setup \& Results} To evaluate this challenge, we measure the global normalized regret with respect to the performance of the best converged policy (across algorithms). Let $window\_size$ be the size of a sliding window $w_k$ across episodes where $k$ is the index of the earliest episode contained in the window. We calculate the highest average return across all algorithms using the final $window\_size$ steps of training and denote this value
%We denote the average return obtained by the best policy (across algorithms) during the final $window\_size$ steps of training by 
$R^*_{mean}$.  We also calculate the $95\%$ confidence interval for this window: $[R^*_{lower}, R^*_{upper}]$. 
We denote $w_K$ as the sliding window for which more than 50\% of episodes have a return higher than $R^*_{lower}$, and consider an agent to have converged at episode $K$.
% We consider the agent to have converged at episode $K$, if $K$ is the first episode of the earliest sliding window during which more than 50\% of the episodes have a higher return than $R^*_{lower}$. 
If this condition is not satisfied during training, then $K = M - window\_size$, where $M$ is the total number of episodes. We can then define the \textbf{global normalized regret} as $$\mathcal{L}_{pre-converge}(\pi) =\frac{1}{R^*_{mean}} \left[ K * R^*_{mean} - \sum_{i=0}^K   R_i  \right], $$ which can be read as sum of regrets for each episode $i$,  i.e., the return that would have been achieved by the best final policy minus the actual return that was achieved. The normalized regret for each of the evaluation domains is shown in Figure \ref{fig:regret}. The normalized regret can effectively be interpreted as the amount of actual return lost, prior to convergence, due to poor policy performance. We can observe that DMPO has higher normalized regret than D4PG on all tasks.
 
Another interesting aspect to measure upon convergence is the instability of the converged policy during training. To do so, we define the \textbf{post-convergence instability}, which measures the percentage of post-convergence episodes for which the return is below $R^*_{lower}$. This can be written as:
$\mathcal{L}_{post-converge}(\pi) = 100 * \frac{\sum_{i=K}^M \mathbf{1}(R_i \geq R^*_{lower})}{M - K},$
where $\mathbf{1}(.)$ is an indicator function.

 The average post-convergence instability for each of the domains\footnote{Note that there are no error bars for humanoid because none of the runs converge to the best performance across algorithms.} is shown in Figure \ref{fig:stability}. As can be seen in the figure, DMPO also has higher instability than D4PG, except for \cartpoleswingup.
 
 The regret and instability metrics together can be used to summarize the sample efficiency of different algorithms. Note that they are both computed with respect to the best known performance for each task. This means that, if a new algorithm is developed that has better performance, the values of these metrics will change as a result. This is by design: when a better method comes along, it should heighten the regret of the previous ones. Note that we could have used the best possible performance for each task instead of the performance of the best known policy, but if we did that we would have run the risk that no algorithm converged to that value, making the regret potentially unbounded. We could also have normalized each algorithm by its own final performance, but that would have made it hard to compare across algorithms.
 
 The results not only show D4PG to be generally more sample efficient, but can also be used to compare the difficulty of achieving sample efficient learning across domains. For instance, it is interesting that while D4PG takes longer to get to a policy on \humanoidwalk, the policy it eventually converges to is more stable than the one for \walkerwalk. We hope that analysing algorithms in this way will enable a practitioner to (1) develop algorithms that are sample efficient and reduce the regret until convergence; and (2) ensure that, once converged, the algorithm is stable. These two properties are highly desirable in many industrial systems.

% \afterpage{\clearpage}

% Old figure for stability:
%  \begin{figure}[h]
%     \centering
%     \begin{subfigure}{.45\textwidth}
%         \centering
%   \includegraphics[width=\textwidth]{figures/D4PG/perc_unstable_D4PG.pdf}
%   \caption{D4PG}\label{fig:d4pg_stability}
% \end{subfigure}
% \begin{subfigure}{.45\textwidth}
%         \centering
%   \includegraphics[width=\textwidth]{figures/DMPO/perc_unstable_DMPO.pdf}
%   \caption{DMPO}\label{fig:dmpo_stability}
%   \end{subfigure}
%   \caption{Once converged, we measure the normalized instability regret for D4PG and DMPO for each domain.}
% \end{figure}

% \begin{figure}
%   \centering
%   \includegraphics[width=0.6\textwidth]{figures/MPO/regret_MPO.pdf}
  
%   \caption{Normalized regret and standard deviation with respect to final policy performance. Assuming the final policy has an average return of $R_{mean}$ with a 95\% confidence interval $[R_{lower}, R_{upper}]$, and that the agent first reaches a return of $R_{lower}$ in episode k, the normalized regret is $\frac{1}{R_{mean}} \sum_{i=0}^K L_i$, where $L_i$ is the regret for episode $i$ with respect to the final policy performance.}
%   \label{fig:efficiency}
% %   \vspace{-1.0cm}
% \end{figure}

\FloatBarrier
\subsection{Challenge 2: System Delays}
\label{subsec:delays}

\paragraph{Motivation \& Related Work} Most real systems have delays in either sensing, actuation, or reward feedback. These might occur because of low-frequency sensing and actuation, because of safety checks or other transformations performed on the selected action before it is actually implemented, or because it takes time for an action's effect to be fully manifested.  

\cite{MLJ12-hester} focus on controlling a robot vehicle with significant delays in the control of the braking system. They incorporate recent history into the state of the agent so that the learning algorithm can learn the delay effects itself. \cite{delayed-outcomes} look at delays in recommender systems, where the true reward is based on the user's interaction with the recommended item, which may take weeks to determine. 
They both present a factored learning approach that is able to take advantage of intermediate reward signals to improve learning in these delayed tasks.
\cite{hung2018optimizing}  introduce a method to better assign rewards that arrive significantly after a causative event.  They use a memory-based agent, and leverage the memory retrieval system to properly allocate credit to distant past events that are useful in predicting the value function in the current timestep. They show that this mechanism is able to solve previously unsolveable delayed reward tasks.
\cite{arjona2018rudder} introduce the RUDDER algorithm, which uses a backwards-view of a task to generate a return-equivalent MDP where the delayed rewards are re-distributed more evenly throughout time. This return-equivalent MDP is easier to learn, is guaranteed to have the same optimal policy as the original MDP, and the approach shows improvements  in Atari tasks with long delays.

\paragraph{Experimental Setup \& Results} The \suite implements delays in observation, action and reward with an  $n$-step buffer between the environment and the agent. An action delay is defined here as delaying the agent's action execution for $n$ timesteps, whereas an observation/reward delay is defined as withholding an agent's observation/reward for $n$ timesteps. We can evaluate the effects of the delay on an agent's performance by looking at the episodic return upon convergence.

\begin{figure}
  \centering
  \begin{subfigure}{.9\textwidth}
  \centering
  \includegraphics[width=\textwidth]{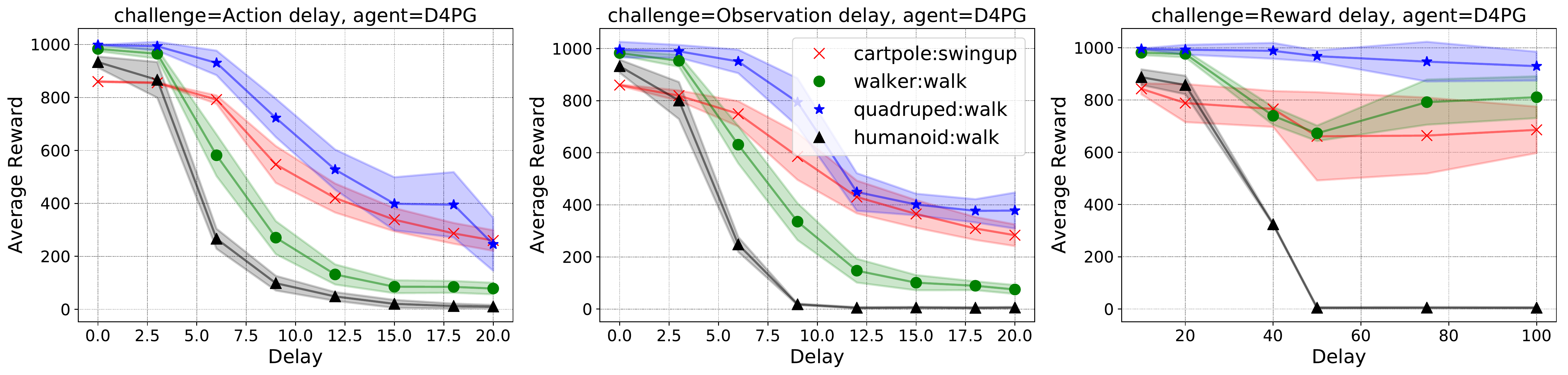}
  \caption{D4PG}\label{fig:d4pg_all_delays}
  \end{subfigure}
  \begin{subfigure}{.9\textwidth}
  \centering
  \includegraphics[width=\textwidth]{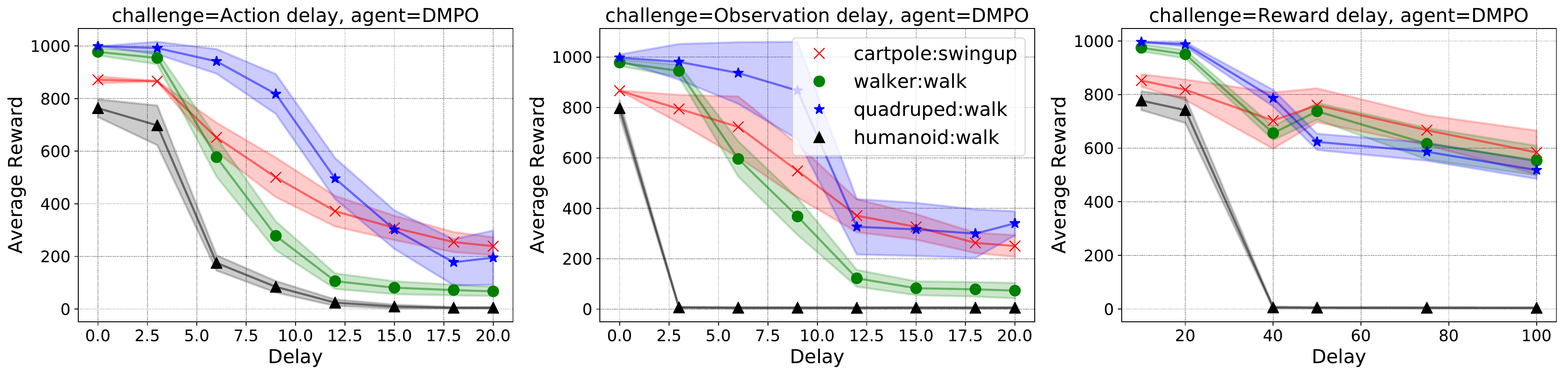}   
  \caption{DMPO}\label{fig:dmpo_all_delays}
  \end{subfigure}
  \caption{Average performance on the four tasks under varying action (left) and observation (middle) delays from a delay of $0$ to a delay to $20$ timesteps. Reward delays (right) include delays from $0$ to $100$ timesteps}
\end{figure}

Figures~\ref{fig:d4pg_all_delays} and \ref{fig:dmpo_all_delays} show the performance of D4PG and DMPO respectively under increasing levels of action, observation and reward delay. As expected, when delays increase, the performance of the algorithm decreases. Both algorithms appear to be less sensitive to reward delay compared to delays in observations or actions. This can be seen in the right-most plot of Figures~\ref{fig:d4pg_all_delays} and \ref{fig:dmpo_all_delays}, where the reward delay (x-axis) has to be increased to $100$ timesteps to see a significant drop in performance. The reason the agent may be more robust to reward delay is that even though the reward is delayed, it can ultimately be credited to an action that led to achieving that reward, even for relatively large delays. However, for more complicated tasks such as \humanoidwalk, where action credit assignment is less obvious for large delays, performance degrades quickly.
It should also be noted that the performance for observation delays is similar to that of action delays. The subtle difference between these settings is the reward that the agent receives at timestep $t$. In the case of action delays, an agent receives the reward $r(s_t, a_{t-n})$ whereas for observation delays, the reward is $r(s_{t-n}, a_t)$. 
%This means that the sum of the two delays can be seen as the total delay between when the environment is in a certain state and when the action computed from that state has an effect on the environment. 

% \begin{figure}
%   \centering
%   \includegraphics[width=0.45\textwidth]{figures/MPO/efficiency_curves_MPO.pdf}
%   \caption{Learning performance on walker walk as the state observation dimension increases. The graph has been cropped to $2000$ episodes for better visualization to highlight the effect that increasing the observation dimensionality has on the learning algorithm. }
%   \label{fig:efficiency_learning}
% \end{figure}

% \begin{figure*}
%   \centering
%   \includegraphics[width=\textwidth]{figures/MPO/all_delays_MPO.pdf}
%   \vspace{-0.6cm}
%   \caption{Average MPO performance on the four tasks under varying action (left) and observation (middle) delays from a delay of $0$ to a delay to $20$ timesteps. Reward delays (right) include delays from $0$ to $100$ timesteps}
%   \label{fig:all_delays}
%   \vspace{-0.5cm}
% \end{figure*}

% \begin{figure*}
%   \centering
%   \includegraphics[width=\textwidth]{figures/MPO/noise_plus_dimensions_MPO.pdf}
%   \vspace{-0.6cm}
%   \caption{Average MPO performance and standard deviation on the four tasks when adding Gaussian action noise (left), Gaussian observation noise (middle) and increasing the dimensionality of the state space with dummy variables (right).}
%   \label{fig:noise}
%   \vspace{-0.5cm}
% \end{figure*}

\FloatBarrier

\subsection{Challenge 3: High-Dimensional Continuous State and Action Spaces}

\paragraph{Motivation \& Related Work} Many practical real-world problems have large and continuous state \& action spaces. For example, consider the huge action spaces of recommender systems~\citep{covington2016deep}, or the number of sensors and actuators to control cooling in a data center~\citep{DM_Datacenter}. These large state and action spaces can present serious issues for traditional RL algorithms, (e.g., see ~\citep{dulac2015deep,Tessler2019b}). 

There are a number of recent works focused on addressing this challenge. \citet{dulac2015deep} look at situations involving a large number of discrete actions, and present an approach based on generating a vector for a candidate action and then doing nearest neighbor search to find the closest applicable action.  For systems with action cardinality that is particularly high ($|\mathcal{A}|>1e5$), it can be practical to decompose the action selection process into two steps: action candidate generation and action ranking, as detailed by~\citet{covington2016deep}.  \citet{zahavy2018learn} propose an Action Elimination Deep Q Network (AE-DQN) that uses a contextual bandit to eliminate irrelevant actions. \citet{he2015deep} present the Deep Reinforcement Relevance Network (DRRN) for evaluating continuous action spaces in text-based games. \citet{Tessler2019b} introduce compressed sensing as an approach to reconstruct actions in text-based games with combinatorial action spaces.

\begin{table}[h]
\centering
\begin{tabular}{|l|c|c|}
\hline
\textbf{Task}             & \textbf{Observation Dimension} & \textbf{Action Dimension} \\ \hline
\textbf{Cartpole:Swingup} & 5                        & 1                         \\ \hline
\textbf{Walker:Walk}      & 18                       & 6                         \\ \hline
\textbf{Quadruped:Walk}   & 78                       & 12                          \\ \hline
\textbf{Humanoid:Walk}    & 67                       & 21                        \\ \hline
\end{tabular}
\caption{The observation and action dimensions for each task.}
\label{app:state_dimensions}
\vspace{-0.4in}
\end{table}

\begin{figure}[h]
    \centering
    \begin{subfigure}{.45\textwidth}
        \centering
  \includegraphics[width=\textwidth]{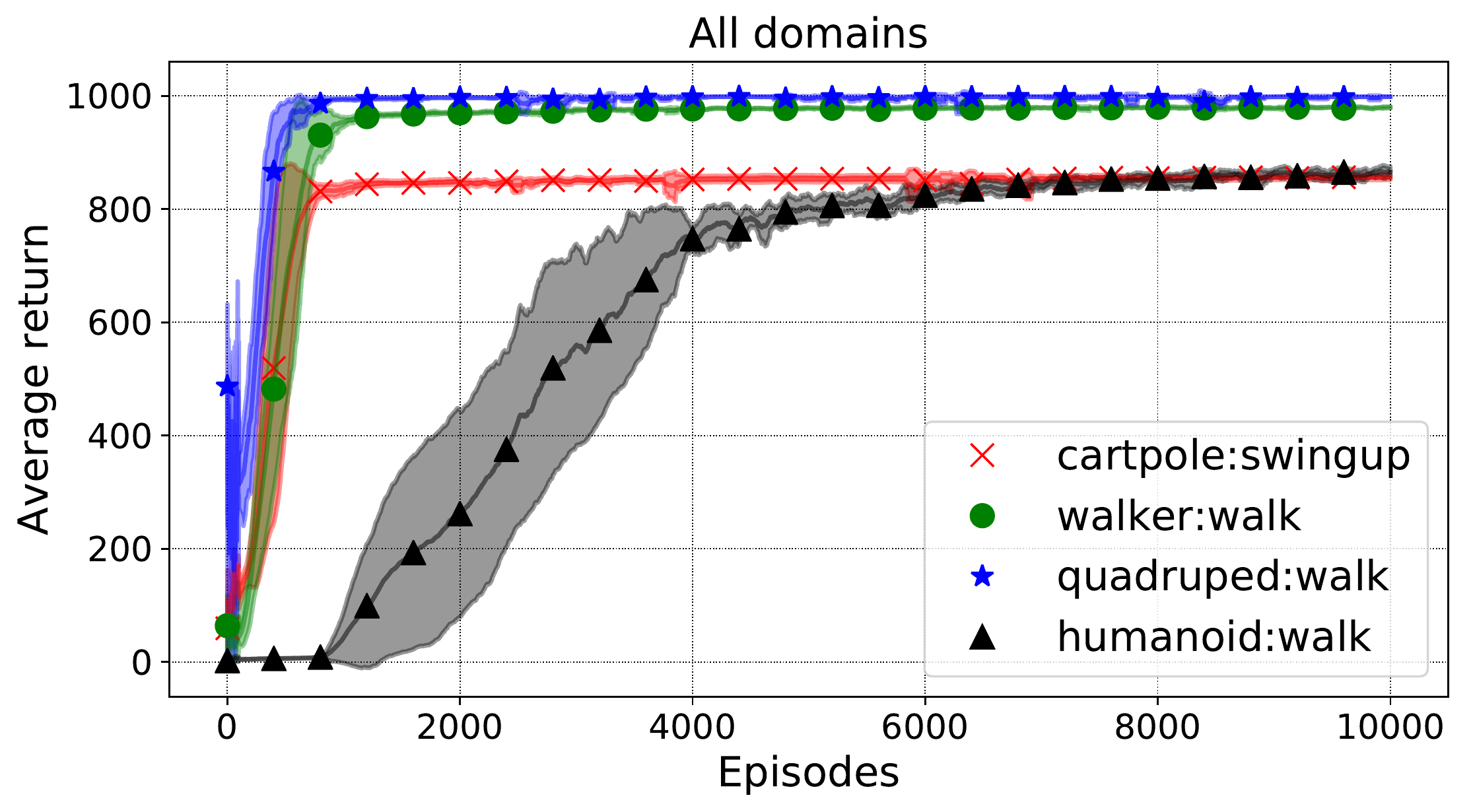}
    \caption{D4PG}\label{fig:d4pg_regular_training_curves}  
    \end{subfigure}
     \begin{subfigure}{.45\textwidth}
        \centering
  \includegraphics[width=\textwidth]{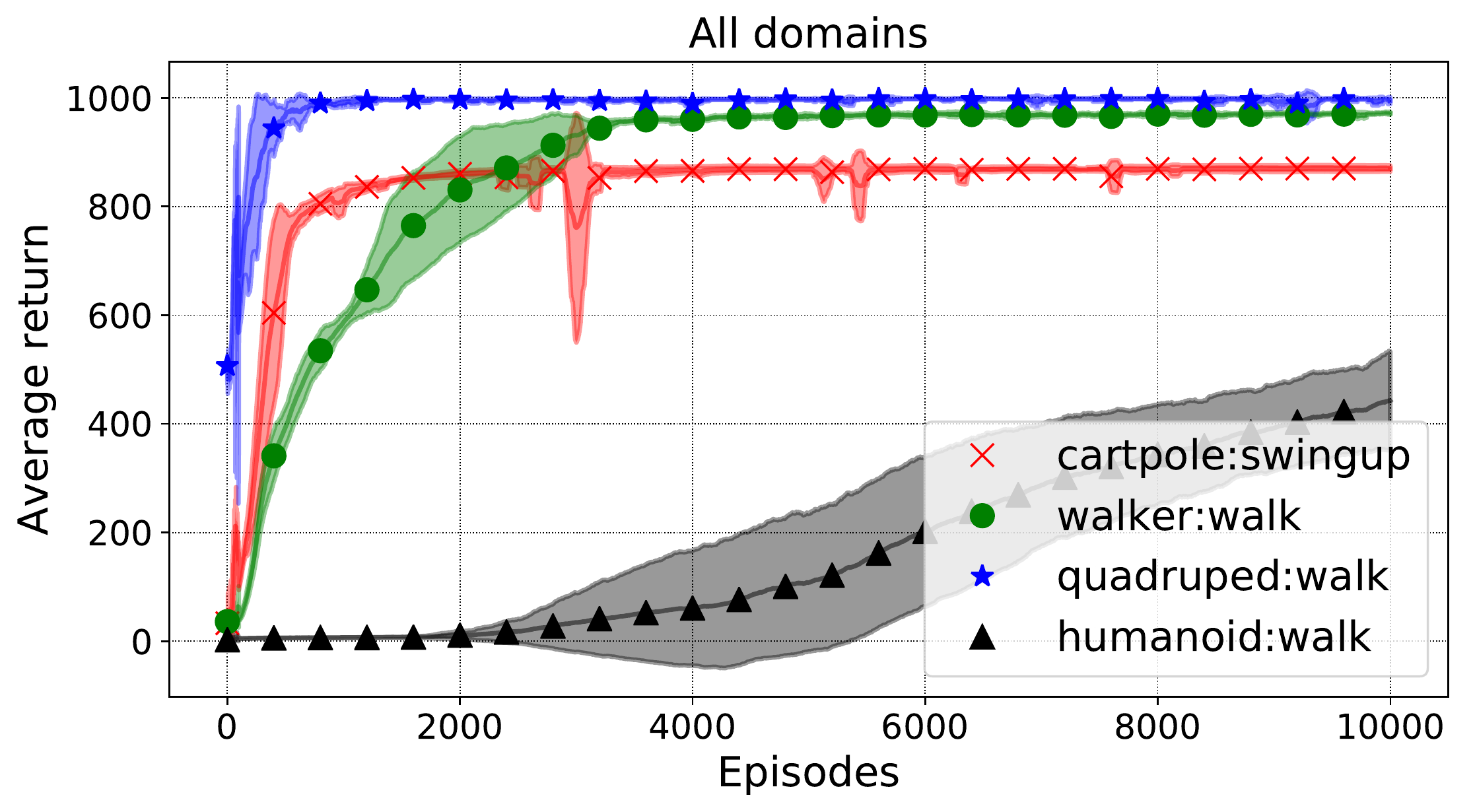}
    \caption{DMPO}\label{fig:dmpo_regular_training_curves}    
  \end{subfigure}
    \caption{Learning performance on all domains as a function of number of episodes, truncated to $10$K episodes for better visualization.}
    \vspace{-0.05in}
\end{figure}

\paragraph{Experimental Setup \& Results} Given the continuous nature of the \suite we chose to simulate a high-dimensional state space, although increasing the action space with dummy dimensions could be interesting for further work.  For readers interested in experiments dealing with large discrete action spaces, please refer to \citep{dulac2015deep} for various experimental setups evaluating large discrete actions spaces.  For this challenge, we first compared results across all the tasks in an unperturbed manner. The state and action dimensions for each task can be found in  Table~\ref{app:state_dimensions}. Both stability of the overall system and the dimensionality affect learning progress. For example, as seen in Figures \ref{fig:d4pg_regular_training_curves} and \ref{fig:dmpo_regular_training_curves} for D4PG and DMPO respectively, \texttt{quadruped} is higher dimensional than \texttt{walker}, yet converges faster since it is a fundamentally more stable system. On the other hand, dimensionality is also a factor as \texttt{cartpole}, which is significantly lower-dimensional than \texttt{humanoid}, converges significantly faster.

% \begin{figure}[h]
%   \centering
%   \includegraphics[width=0.45\textwidth]{figures/MPO/regular_training_curves_MPO.pdf}
%   \caption{Learning performance of MPO on all domains as a function of number of episodes, truncated to $10$K episodes for better visualization. Note that the \humanoidwalk task only converged after approximately 20k steps.}
%   \label{fig:regular_training_curve_mpo}
% \end{figure}

\begin{figure}[h]
  \centering
  \begin{subfigure}{.95\textwidth}
  \centering  
  \includegraphics[width=\textwidth]{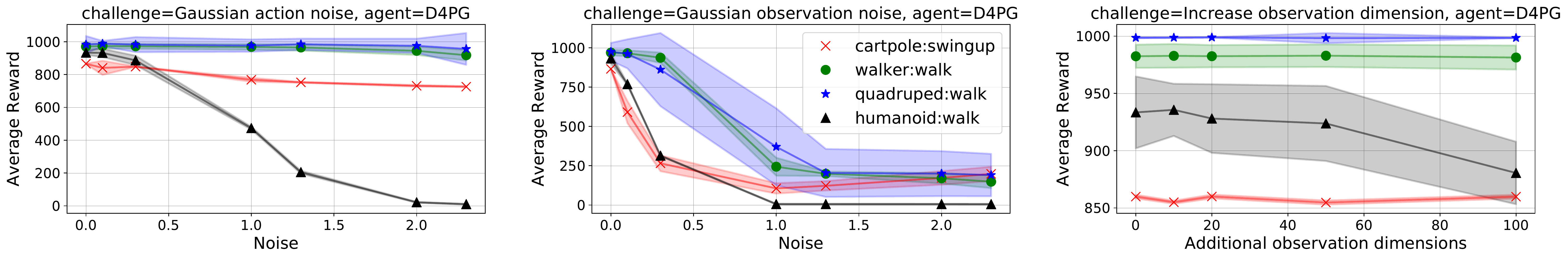}
  \caption{D4PG}\label{fig:d4pg_noise}
  \end{subfigure}  
  \begin{subfigure}{.95\textwidth}
  \centering
  \includegraphics[width=\textwidth]{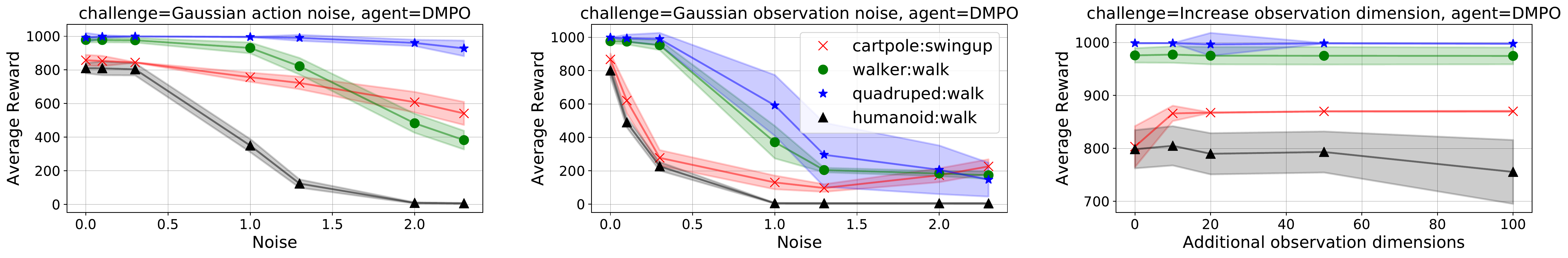}
  \caption{DMPO}\label{fig:dmpo_noise}  
  \end{subfigure}
  \caption{Average performance and standard deviation on the four tasks when adding Gaussian action noise (left), Gaussian observation noise (middle) and increasing the dimensionality of the state space with dummy variables (right).}
\end{figure}

\begin{figure}[H]
    \centering
    \begin{subfigure}{.45\textwidth}
        \centering
  \includegraphics[width=\textwidth]{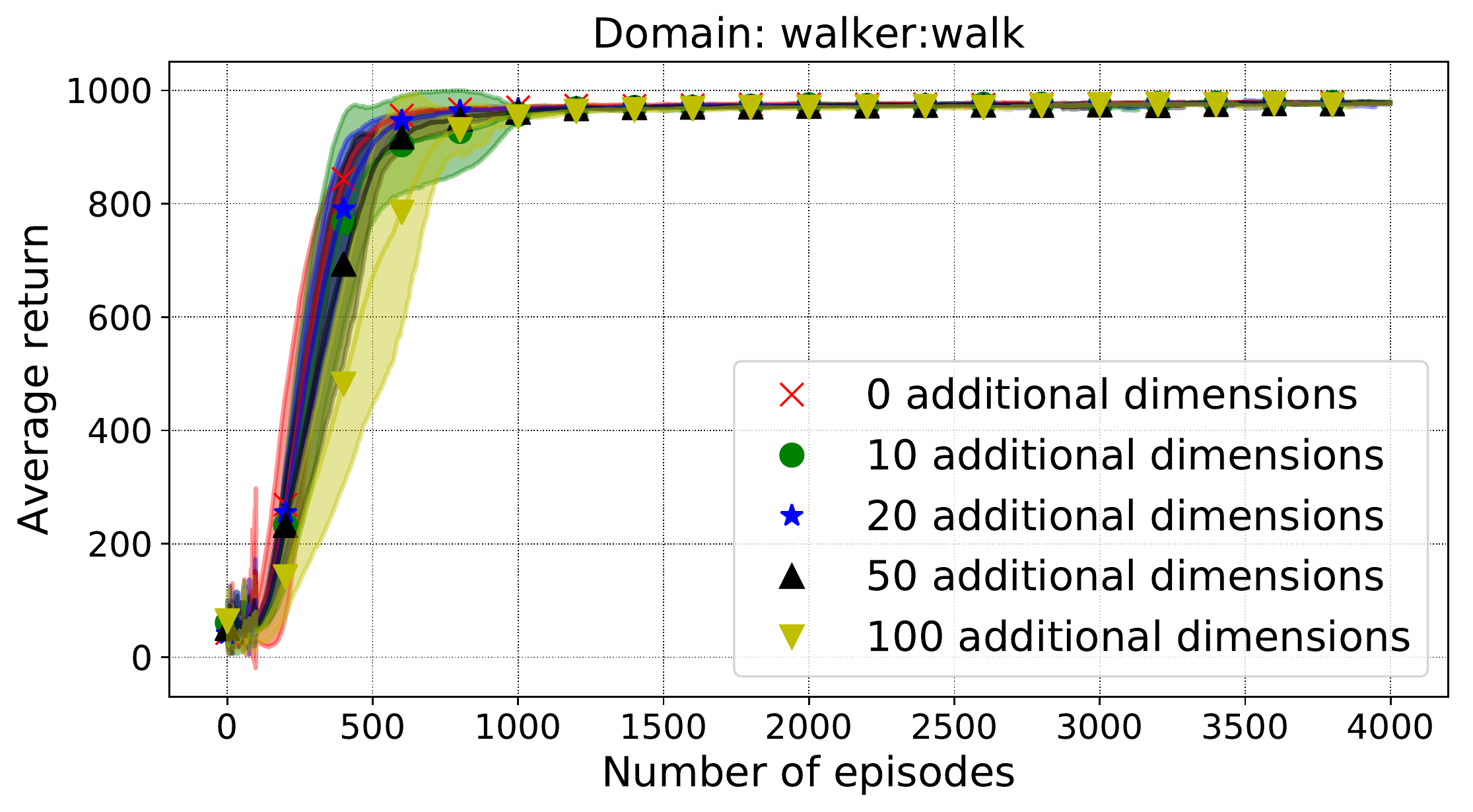}
  \caption{D4PG}\label{fig:d4pg_efficiency_learning}
\end{subfigure}
    \begin{subfigure}{.45\textwidth}
        \centering
  \includegraphics[width=\textwidth]{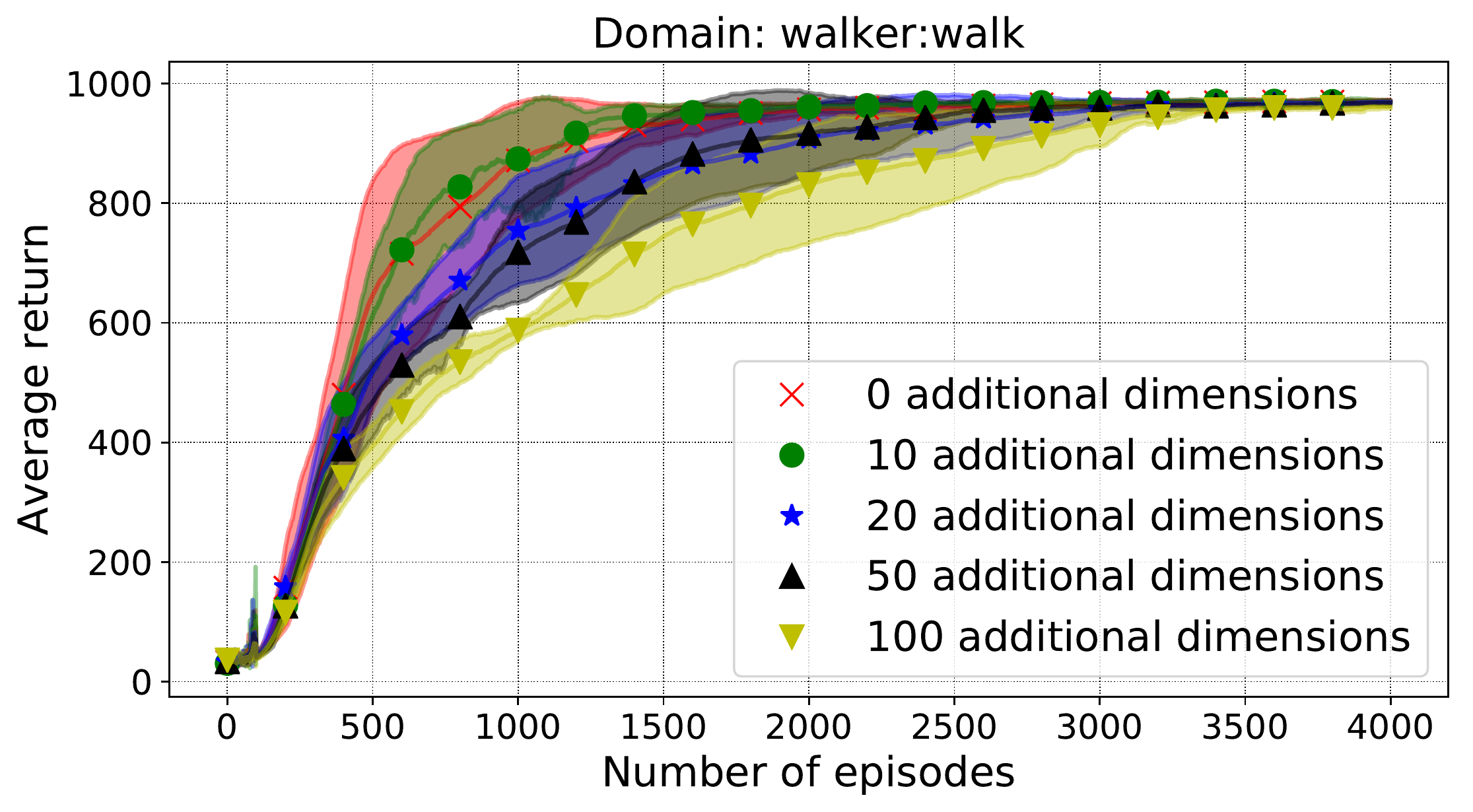}
  \caption{DMPO}\label{fig:dmpo_efficiency_learning}
  \end{subfigure}
    \caption{Learning performance of D4PG (left) and DMPO (right) on walker walk as the state observation dimension increases. The graph has been cropped to $4000$ episodes for better visualization to highlight the effect that increasing the observation dimensionality has on the learning algorithm.}
\end{figure}

We subsequently increased the number of state dimensions of each task with dummy state variables sampled from a zero mean, unit variance normal distribution. We then compare the average return for each task as we increase the state dimensionality. Figures \ref{fig:d4pg_noise} and \ref{fig:dmpo_noise} (right) show the converged average performance of the learning algorithm on each task for D4PG and DMPO respectively. Since the added states were effectively injecting noise into the system, the algorithm learns to deal with the noise and converges to the optimal performance for the cases of \cartpoleswingup, \quadrupedwalk and \walkerwalk. In some cases, e.g. Figures~\ref{fig:d4pg_efficiency_learning} and \ref{fig:dmpo_efficiency_learning} for \walkerwalk, the additional dummy dimensions slightly affect convergence speed indicating that the learning algorithm learns to deal with noise efficiently, but it does slow down learning progress. 
%\todd{interesting, so the dummy dimensions have little effect, which makes sense that the network can learn to just ignore them. Is there something different we could do to illustrate this challenge? Maybe scaling cart-pole up to 2d, 3d, 4d, etc? or discussing going from 2d walker to 3d walker?}
% In the case of \humanoidwalk, the instability of the system and sheer size of the original state space ($67$ dimensions), coupled with the added dummy dimensions appears to degrade performance when adding more than $50$ dummy dimensions as seen in Figure \ref{fig:noise} (right).

% \clearpage
\FloatBarrier
\subsection{Challenge 4: Satisfying Environmental Constraints}
\label{sec:constraints}

\paragraph{Motivation \& Related Work} Almost all physical systems can destroy or degrade themselves and the environment around them if improperly controlled.  Software systems can also significantly degrade their performance or crash, as well as provide improper or incorrect interactions with users. As such, considering constraints on their operation is fundamentally necessary to controlling them. Constraints are not only important during system operation, but also during exploratory learning phases as well.  Examples of physical constraints include limits on system temperatures or contact forces for safe operation, maintaining minimum battery levels, avoiding dynamic obstacles, or limiting end effector velocities. Software systems might have constraints around types of content to propose to users or system load and throughput limits to respect. 

Although system designers may often wrap the learnt controller in a safety watchdog controller, the learnt controller needs to be aware of the constraints to avoid degenerate solutions which lazily rely on the watchdog. We want to emphasize that constraints can be put in place for varying reasons, ranging from monetary costs, to system up-time and longevity, to immediate physical safety of users and operators. Due to the physically grounded nature of our suite, our proposed set of constraints are physically bound and are intended to avoid self-harm, but the suite's framework provides options for users to define any constraints they wish.

Recent work in RL safety~\citep{galdalal, AchiamHTA17,Tessler2018,satija2020constrained} has cast safety in the context of Constrained MDPs (CMDPs)~\citep{altman1999constrained}, and we will concentrate on pre-defined constraints on the environment in this context. Constrained MDPs define a constrained optimization problem and can be expressed as:
\begin{equation}\vspace{-0.2cm}
\max_{\pi \in \Pi} R(\pi) \mbox{ subject to } C^k(\pi) \leq V_k, k  = 1, \ldots,K.
\end{equation}

Here, $R$ is the cumulative reward of a policy $\pi$ for a given MDP, and $C^k(\pi)$ describes the incurred cumulative cost of a certain policy $\pi$ relative to constraint $k$.  The CMDP framework describes multiple ways to consider cumulative cost of a policy $\pi$: the total cost until task completion, the discounted cost, or the average cost. Specific constraints are defined as $c_k(s,a)$.

The CMDP setup allows for arbitrary constraints on state and action to be expressed.  In the context of a physical system these can be as simple as box constraints on a specific state variable, or more complex such as dynamic collision-avoidance constraints. One major challenge with addressing these safety concerns in real systems is that safety violations will likely be very rare in logs of the system. In many cases, safety constraints are assumed and are not even specified by the system operator or product manager.

An extension to CMDPs is budgeted MDPs~\citep{bmdp,Carrara2018AFA}. While for a CMDP, the constraint level $V_k$ is given, for budgeted MDPs, it is unknown. Instead, the policy is learned as a function of constraint level. The user can examine the trade-offs between expected return and constraint level and choose the constraint level that best works for the data. This is a good match for common real-world scenario where the constraints may not be absolute, but small violations may be allowed for a large improvement in expected returns.

Recently, there has a been a lot of work focused on the problem of safety in reinforcement learning. One focus has been the addition of a safety layer to the network \citep{galdalal,optlayer}. These approaches focus on safety during training, and have enabled an agent to learn a task with zero safety violations during training.  There are other approaches~\citep{AchiamHTA17,Tessler2018, bohez2019value} that learn a policy that violates constraints during training but produce a \textit{trained} policy that respects the safety constraints. \cite{stooke2020responsive} introduce the concept of lagrangian damping which leads to improved stability by performing PID control on the lagrangian parameter.
Additional RL approaches include using Lyapunov functions to learn safe policies~\citep{NIPS2018_chow} and exploration strategies that predict the safety of neighboring states~\citep{TurchettaB016,WachiSYO18}. \cite{satija2020constrained} introduce the concept of a backward value function for a more conservative optimization algorithm.
A Probabilistic Goal MDP~\citep{Mankowitz2016,Xu2011} is another type of objective that encourages an agent to achieve a pre-defined reward level irrespective of the time it takes to complete the task. This objective encourages risk-averse behaviour leading to safer and more robust policies. \cite{thomas2015safe} proposes a safe RL algorithm that searches for new and improved policies while ensuring that the probability of selecting bad policies is low. \cite{calian2020balancing} provide a meta-gradient solution to  balancing the trade-off between maximizing rewards and minimizing constraint violations. This D4PG variant learns the learning rate of the lagrange multiplier in a soft-constrained optimization procedure. \cite{thomas2017ensuring} propose a new framework for designing machine learning algorithms that simplifies the problem of specifying and regulating undesired behaviours. There have also been approaches to learn a policy that satisfies constraints in the presence of perturbations to the dynamics of an environment \citep{mankowitz2020robust}. 
%There has also been work on budgeted MDPs where the constraint level $V_k$ is unknown \citep{bmdp,Carrara2018AFA}.

\paragraph{Experimental Setup \& Results} To demonstrate the complexity of system constraints, we leverage the CMDP formalism to include a series of binary safety-inspired constraints to our challenge domains.  These constraints can be either considered passively, as a measure of an agent's behavior, or they can be included in the agent's observation so that the agent may learn to avoid them. 
%By combining the constraints with the multi-objective challenge, one can encourage the agent to respect them using a combined reward objective as demonstrated in Section \ref{sec:multiobj}.   

% \begin{figure}
%   \centering
% %   \vspace{-0.5cm}
%   \includegraphics[width=\textwidth]{figures/D4PG/d4pg_cartpole_swingup_safety.pdf}

%   \caption{Left: Evolution of the number of constraint violations for a naive learner as \texttt{safety\_coeff} is varied.  We can see the number of violations increasing as \texttt{safety\_coeff} approaches 0.  Right: Average number of violations during training for each timestep of an episode.  The strong peak in \texttt{slider\_pos} and \texttt{slider\_accel} correlates with the initial pole swing-up.?? TODO: Kill this figure.}
%   \label{fig:safety_cp}
% \end{figure}

As an example, our \texttt{cartpole} environment with variables $x,\theta$ (cart position and pole angle) includes three boolean constraints:
\begin{enumerate}
    \item \texttt{slider\_pos}, which restricts the cart's position on the track: $x_l < x < x_r$.
    \item \texttt{slider\_accel}, which limits cart acceleration: $ \ddot{x} < A_\text{max}$.
    \item \texttt{balance\_velocity}, a slightly more complex constraint, which limits the pole's angular velocity when it is close to being balanced: $ \left| \theta \right| > \theta_L \vee \dot{\theta} < \dot{\theta}_V$.
\end{enumerate}
The full set of available constraints across all tasks is described in Table~\ref{tab:safe_envs_full}.  Each constraint can be tuned by modifying a parameter $\texttt{safety\_coeff} \in [0,1]$ where $0$ is harder and $1$ is easier to satisfy.

\begin{figure}
\begin{subfigure}{\textwidth}
\centering
  \includegraphics[width=0.48\textwidth]{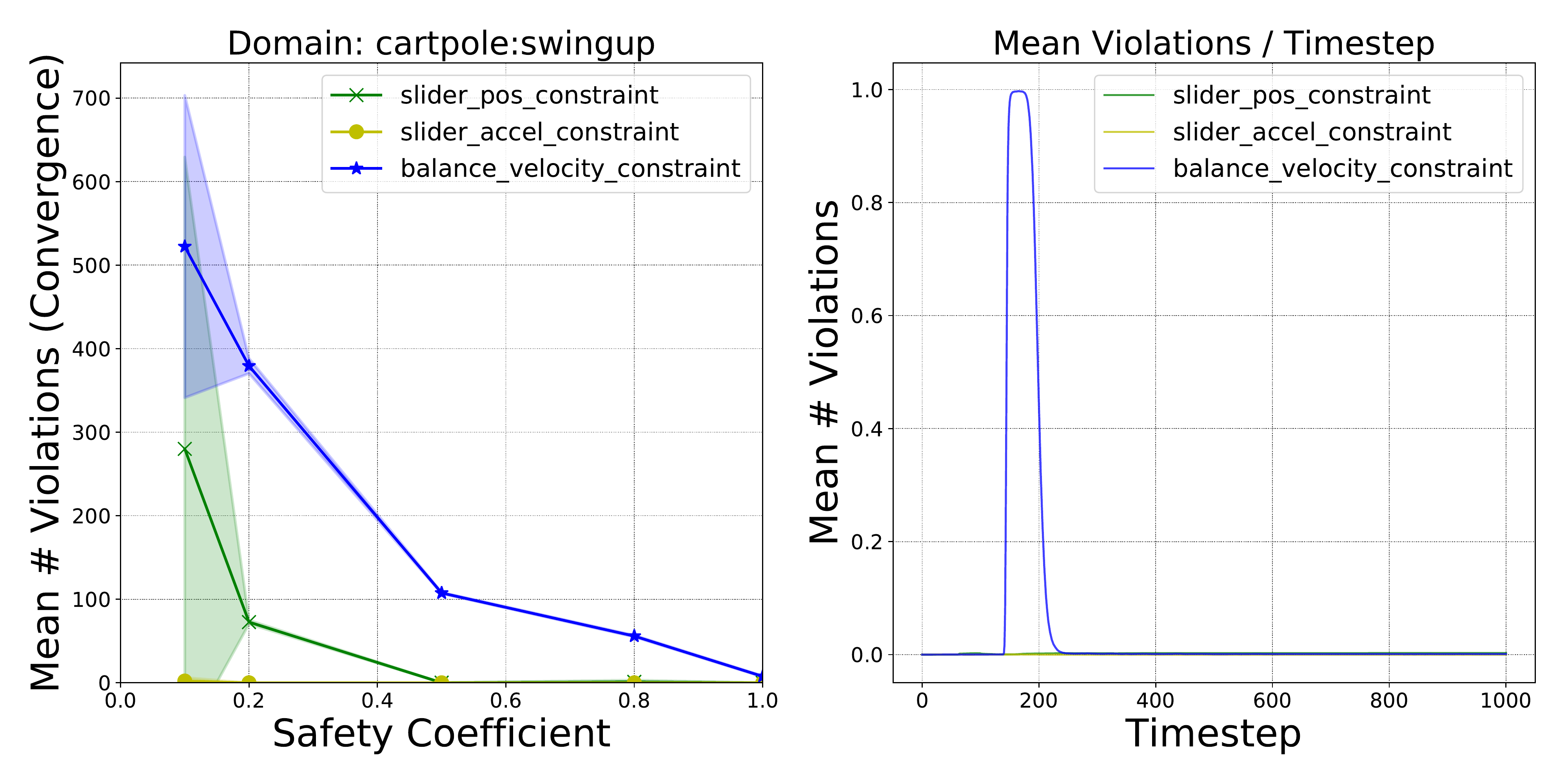}
  \includegraphics[width=0.48\textwidth]{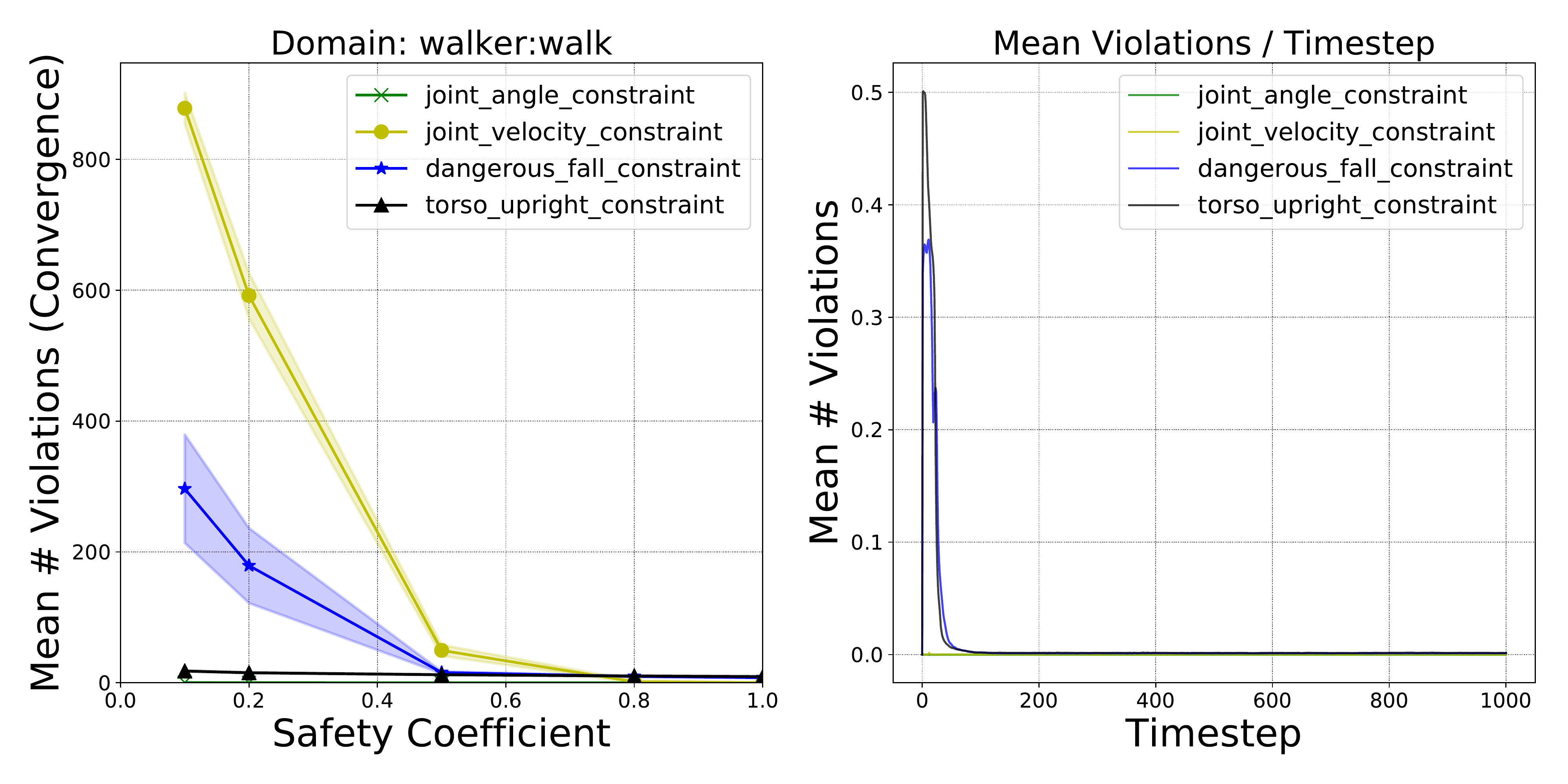}
  \includegraphics[width=0.48\textwidth]{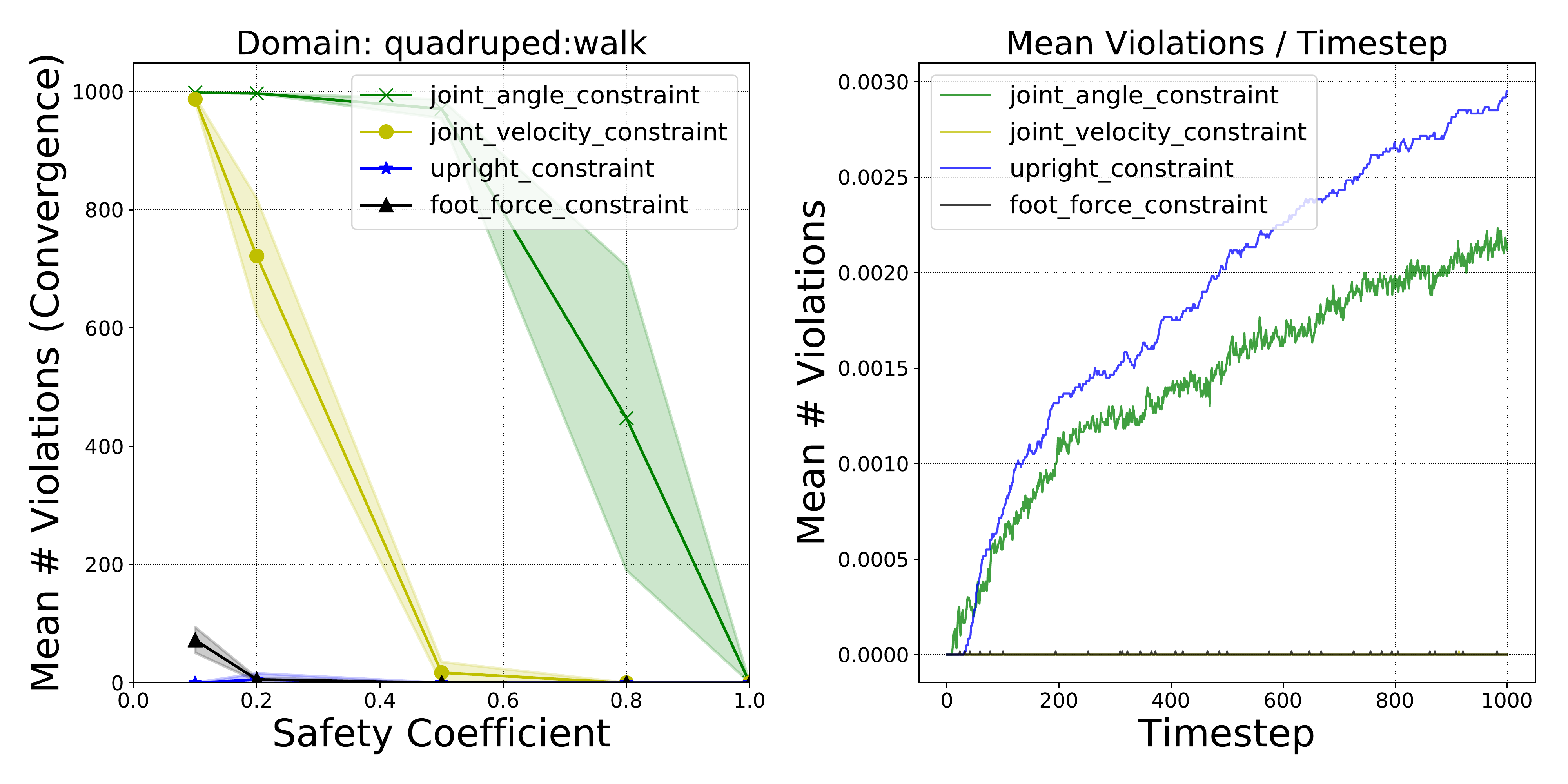}
  \includegraphics[width=0.48\textwidth]{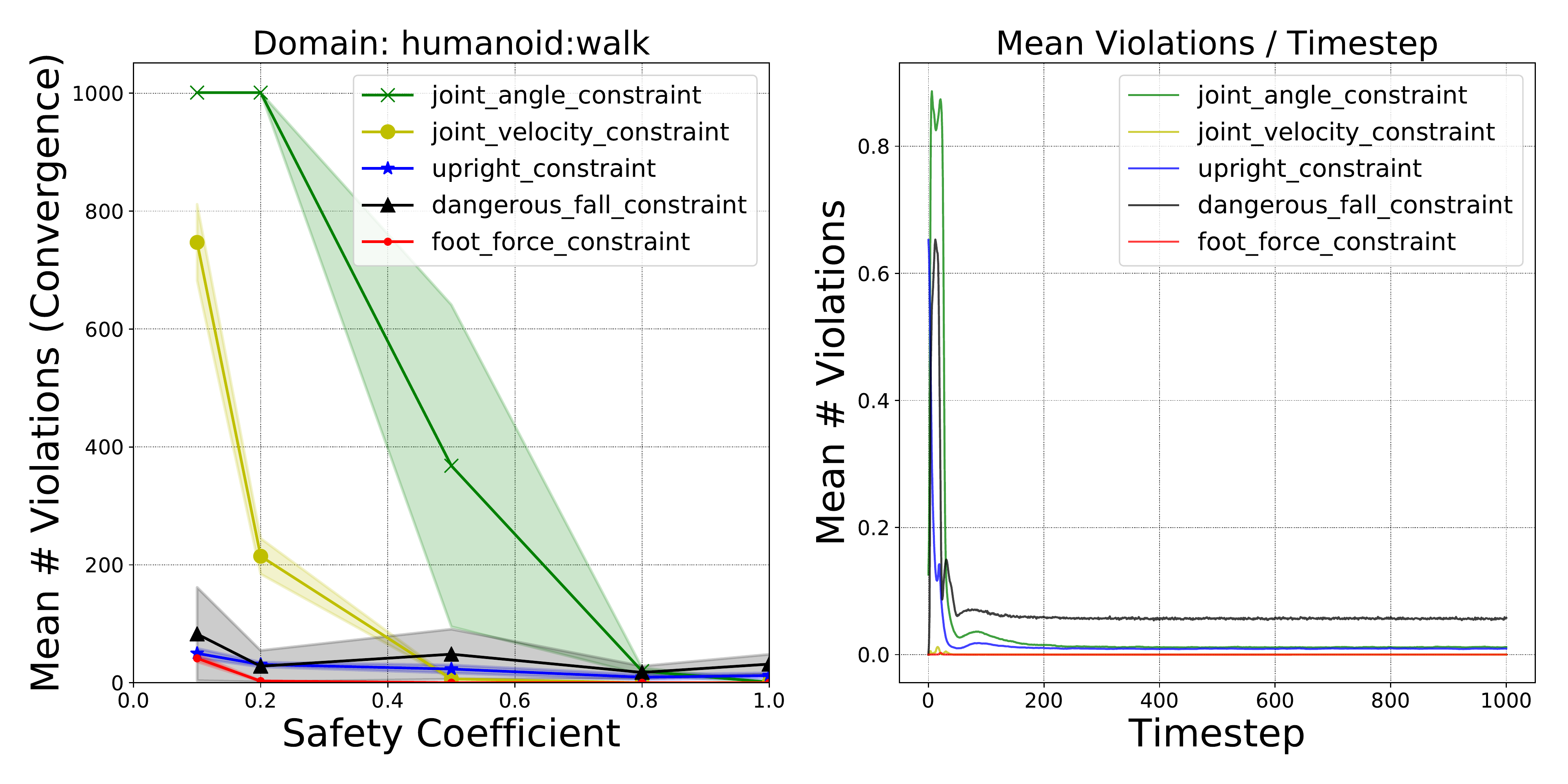}
  \caption{Safety violations for a D4PG learner.}
\end{subfigure}

\begin{subfigure}{\textwidth}
\centering
  \includegraphics[width=0.48\textwidth]{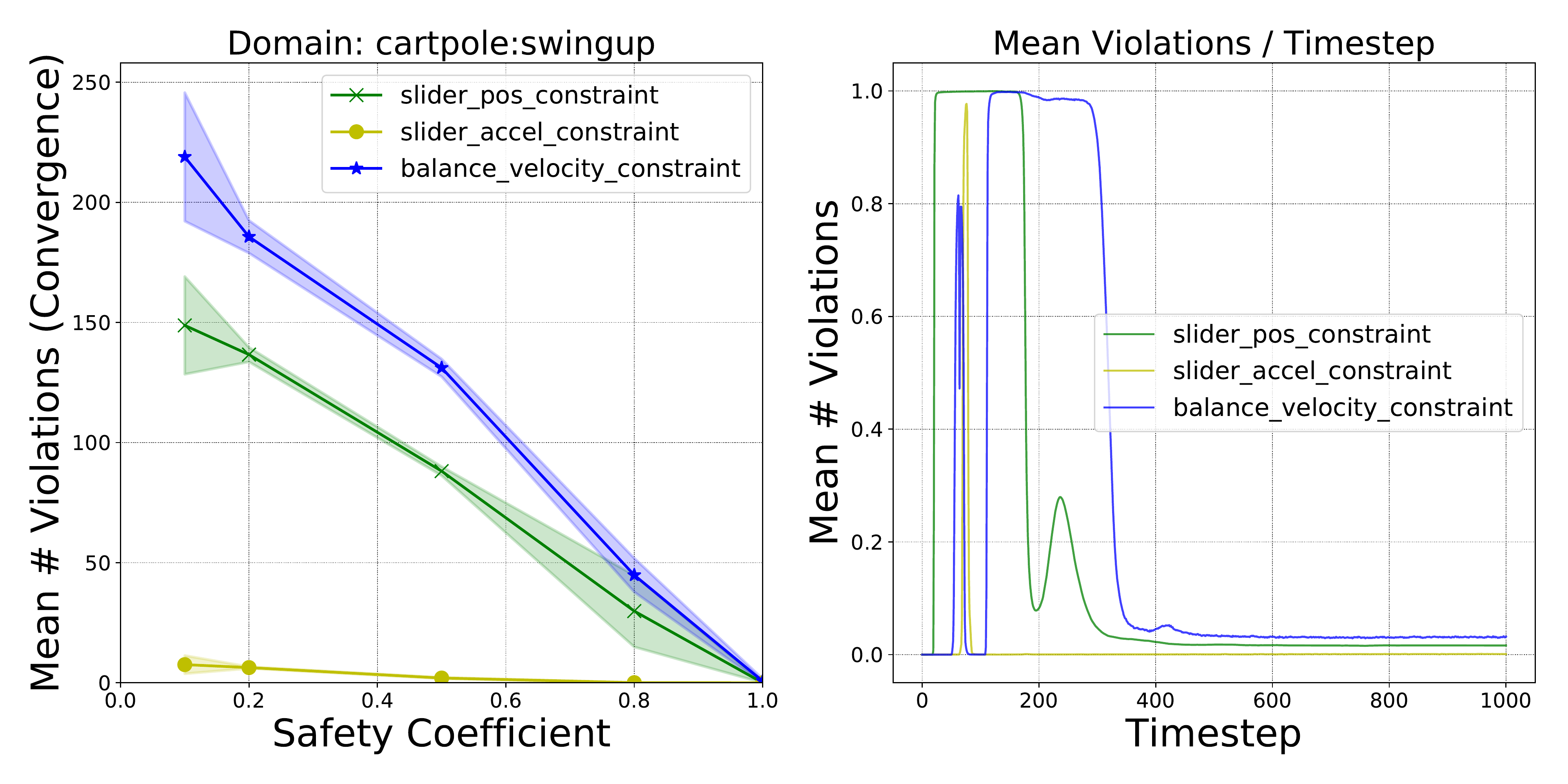}
  \includegraphics[width=0.48\textwidth]{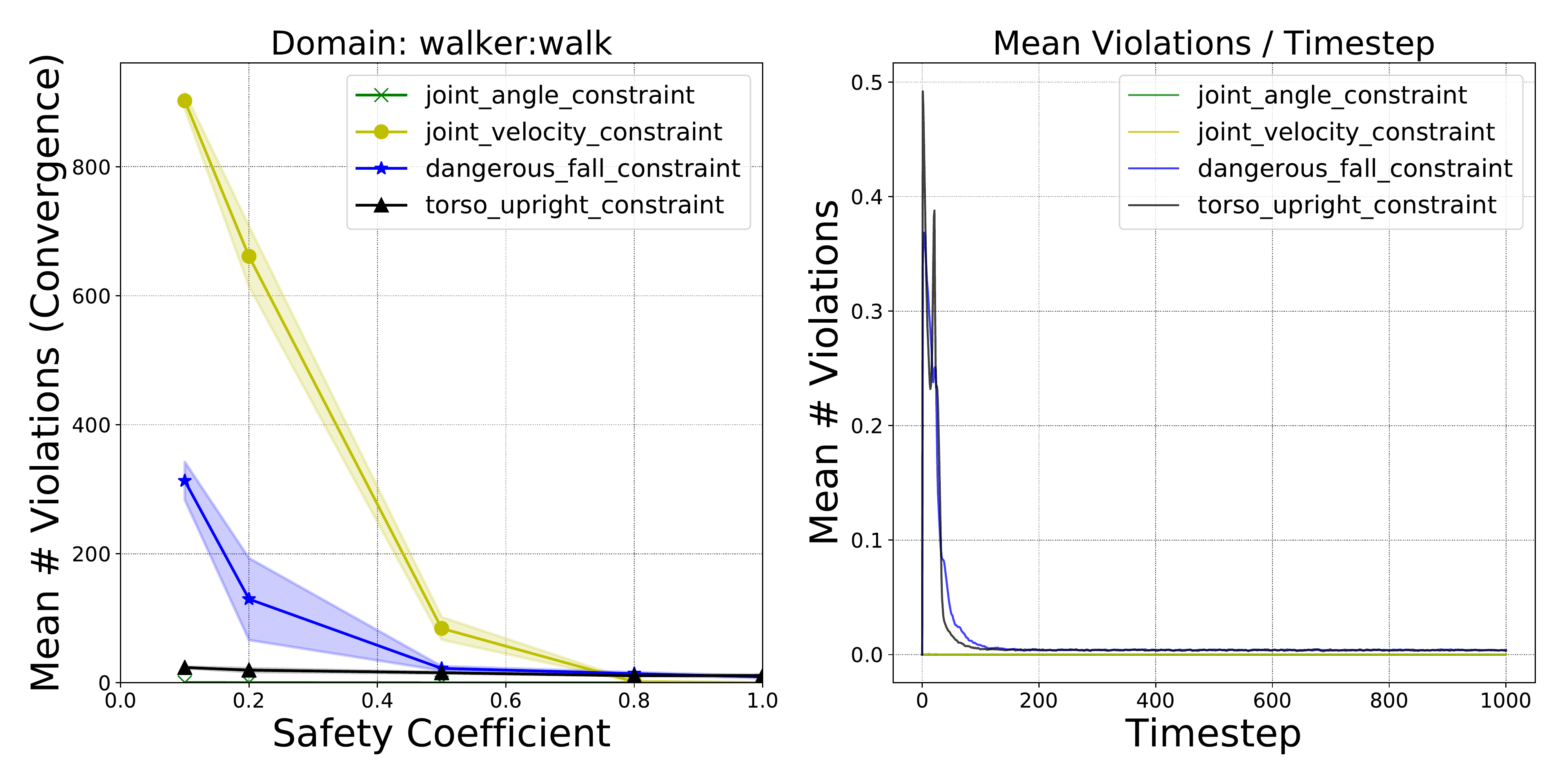}
  \includegraphics[width=0.48\textwidth]{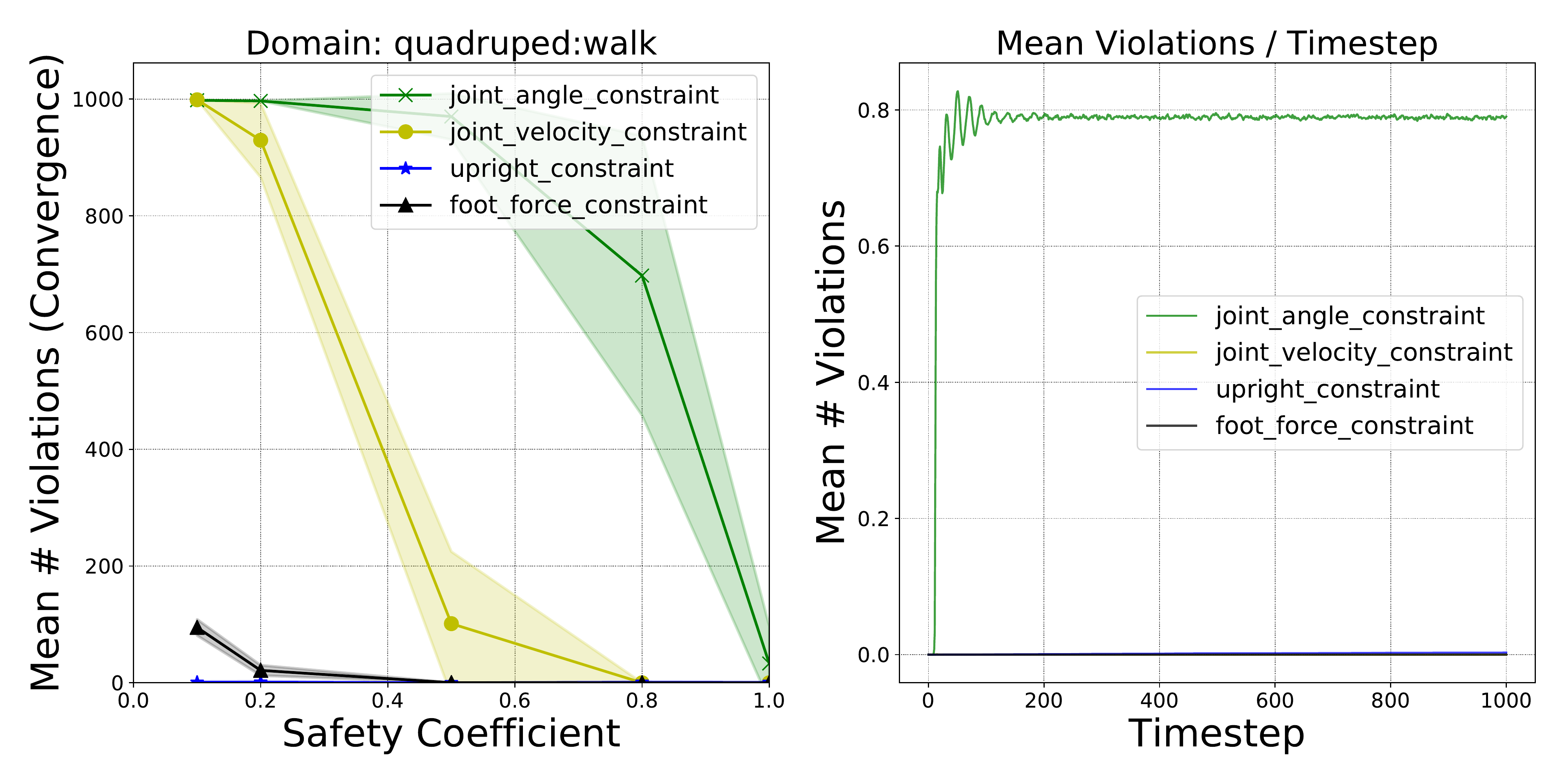}
  \includegraphics[width=0.48\textwidth]{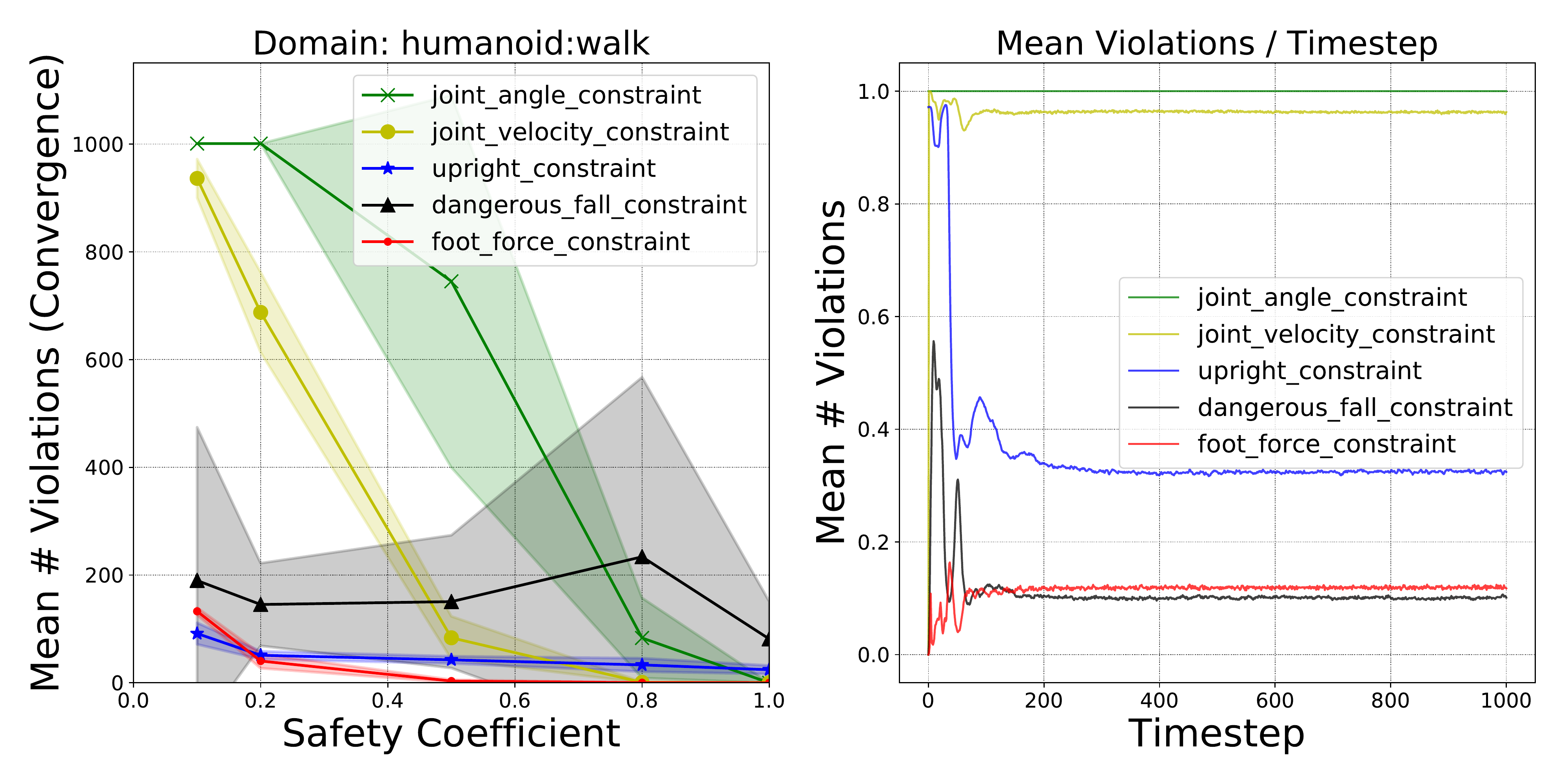}
  \caption{Safety violations for a DMPO learner.}
\end{subfigure}
\caption{For each task, the left plot shows the evolution of the number of safety constraints upon convergence for various values of the safety coefficient. The right plot shows, for a safety coefficient of 1, the evolution of safety violations over an episode on average.  This is to illustrate how different violations get triggered at different points in an episode.}
  \label{fig:safety_hyperparameters_sweeps}
\end{figure}

To evaluate this challenge, we track the number of constraint violations by the agent, for each constraint, throughout training.   We present the effects of \texttt{safety\_coeff} on all four environments in Figure \ref{fig:safety_hyperparameters_sweeps}. For each task, we illustrate both the effects of \texttt{safety\_coeff} as a function of the average number of constraint violations upon convergence (left) as well as the average number of violations throughout an episode of \texttt{cartpole\_swingup} (right). We can see that \texttt{safety\_coeff} makes the task more difficult as it tends towards 0, and that constraint violations are non-uniform throughout time e.g. as the cart swings back and forth, the pole, position and acceleration constraints are more frequently violated. 

Although the learner presented here ignores the constraints, we also include a multi-objective task which combines the task's reward function with a constraint violation penalty in Section \ref{sec:multiobj}.

\begin{table}
\small
\begin{center}
% \begin{minipage}[t]{0.45\linewidth}
\begin{tabular}{| l | l | }
 \multicolumn{2}{l}{\textbf{Cart-Pole} Variables: $x, \theta$} \\
 \hline
 \textbf{Type} & \textbf{Constraint} \\
 \hline
 \hline
 \texttt{slider\_pos} & 
  $x_l < x < x_r$ \\
 \hline
 \texttt{slider\_accel}
 &$ \ddot{x} < A_\text{max}$ \\
 \hline
 \texttt{balance\_velocity}
 & $ \left| \theta \right| > \theta_L \vee \dot{\theta} < \dot{\theta}_V $ \\
 \hline
\end{tabular}
\\
% \end{minipage}
% \begin{minipage}{0.45\linewidth}
\begin{tabular}[t]{| l | l |}
\multicolumn{2}{l}{} \\
\multicolumn{2}{l}{\textbf{Walker} Variables: $\bm{\theta}, \bm{u}, \bm{F}$} \\
\hline
 \textbf{Type} & \textbf{Constraint} \\
 \hline
 \hline
 \texttt{joint\_angle}
 & $\bm{\theta}_{L} < \bm{\theta} < \bm{\theta}_{U}$ \\
 \hline
 \texttt{joint\_velocity}
 & $ \max_i \left| \dot{\bm{\theta}_i} \right| < L_{\dot{\theta}} $ \\
 \hline 
 \texttt{dangerous\_fall}
 & $0 < (\bm{u}_z \cdot \bm{x})$ \\
 \hline
 \texttt{torso\_upright}
  &  $0 < \bm{u}_z$\\
 \hline
\end{tabular}
% \end{minipage}
\\
% \begin{minipage}{0.45\linewidth}
\begin{tabular}[t]{| l | l |}
\multicolumn{2}{l}{} \\
\multicolumn{2}{l}{\textbf{Quadruped} Variables: $\bm{\theta}, \bm{u}, \bm{F}$} \\
\hline
 \textbf{Type} & \textbf{Constraint} \\
 \hline
 \hline
  \texttt{joint\_angle}
 & $\theta_{L, i} < \bm{\theta}_i < \theta_{U,i}$ \\
 \hline
 \texttt{joint\_velocity}
 & $ \max_i \left| \dot{\bm{\theta}_i} \right| < L_{\dot{\theta}} $ \\
 \hline
 \texttt{upright}
 &  $0 < \bm{u}_z$\\
 \hline
\texttt{foot\_force}
 &$ \bm{F}_{\text{EE}} < F_\text{max} $ \\
\hline
\end{tabular}
% \end{minipage}
% \begin{minipage}{0.45\linewidth}
\\
\begin{tabular}[t]{| l | l |}
 \multicolumn{2}{l}{} \\
\multicolumn{2}{l}{\textbf{Humanoid} Variables: $\bm{\theta}, \bm{u}, \bm{F}$} \\
\hline
 \textbf{Type} & \textbf{Constraint} \\
 \hline
 \hline
 \texttt{joint\_angle\_constraint}
 & $\theta_{L, i} < \bm{\theta}_i < \theta_{U,i}$ \\
 \hline
 \texttt{joint\_velocity\_constraint}
 & $\max_i \left| \dot{\bm{\theta}_i} \right| < L_{\dot{\theta}}$ \\
 \hline
 \texttt{upright\_constraint} 
 & $0 < \bm{u}_z$ \\
 \hline
 \multirow{2}{*}{\texttt{dangerous\_fall\_constraint}}
 & $ \bm{F}_{head} < F_{\text{max},1}$ \\
 & $ \bm{F}_{torso} < F_{\text{max},2}$ \\
 \hline
 \texttt{foot\_force\_constraint} 
 & $ \bm{F}_{\text{Foot}} < F_\text{max,3} $ \\
 \hline

\end{tabular}
% \end{minipage}
\end{center}
\caption{Safety constraints for each domain.}
\label{tab:safe_envs_full}
\end{table}

% Previous work in the context of RL with constraints \cite{galdalal,AchiamHTA17,Tessler2018, bohez2019value} has been largely focused on variants of the Constrained-MDP (CMDP) formulation; that is $\max_{\pi \in \Pi} R(\pi) \mbox{ subject to } C^k(\pi) \leq V_k, k  = 1, \ldots,K$ \cite{altman1999constrained}. There has also been work on budgeted MDPs where the constraint level $V_k$ is unknown \citep{bmdp,Carrara2018AFA}. .

\FloatBarrier
\subsection{Challenge 5: Partial Observability and Non-Stationarity}
\label{sec:partial_obs}
\paragraph{Motivation \& Related Work} Almost all real systems where we would want to deploy RL are partially observable. For example, on a physical system, we likely do not have observations of the wear and tear on motors or joints, or the amount of buildup in pipes or vents. We have no observations on the quality of the sensors and whether they are malfunctioning. On systems that interact with users such as recommender systems, we have no observations of the mental state of the users. Often, these partial observations appear as noise (e.g., sensor wear and tear or uncalibrated/broken sensors), non-stationarity (e.g. as a pump's efficiency degrades) or as stochasticity (e.g. as each robot being operated behaves differently).

\paragraph{Partial observability.} 
Partially observable problems are typically formulated as a partially observable Markov Decision Process (POMDP)~\citep{cassandra1998survey}. The key difference from the MDP formulation is that the agent's observation $x \in X$ is now separate from the state, with an observation function $O(x\mid s)$ giving the probability of observing $x$ given the environment state $s$. There are a couple common approaches to handling partial observability in the literature. One is to incorporate history into the observation of the agent: DQN~\citep{mnih2015human} stacks four Atari frames together as the agent's observation to account for partial observability. Alternatively, an approach is to use recurrent networks within the agent, enabling them to track and recover hidden state. \cite{HausknechtS15} apply such an approach to DQN, and show that the recurrent version can perform equally well in Atari games when only given a single frame as input.
\cite{nagabandi}  propose an approach modeling the system as non-stationary with a time-varying reward function, and use meta-learning to find policies that will adapt to this non-stationarity. Much of the recent work on transferring learned policies from simulation to the real system also focuses on this area, as the underlying differences between the systems are not observable~\citep{andrychowicz2018learning,peng2018sim}. 

\begin{figure}[!htb]
  \centering
  \begin{subfigure}{.95\textwidth}
  \centering    
  \includegraphics[width=\textwidth]{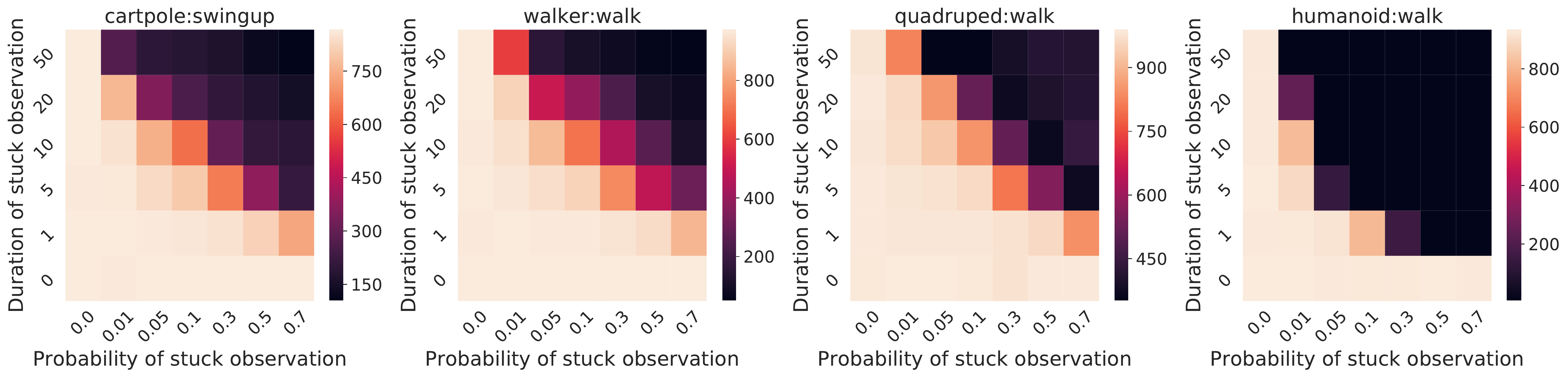}
  \caption{D4PG}\label{fig:d4pg_stuck}  
  \end{subfigure}
  \begin{subfigure}{.95\textwidth}
  \centering    
  \includegraphics[width=\textwidth]{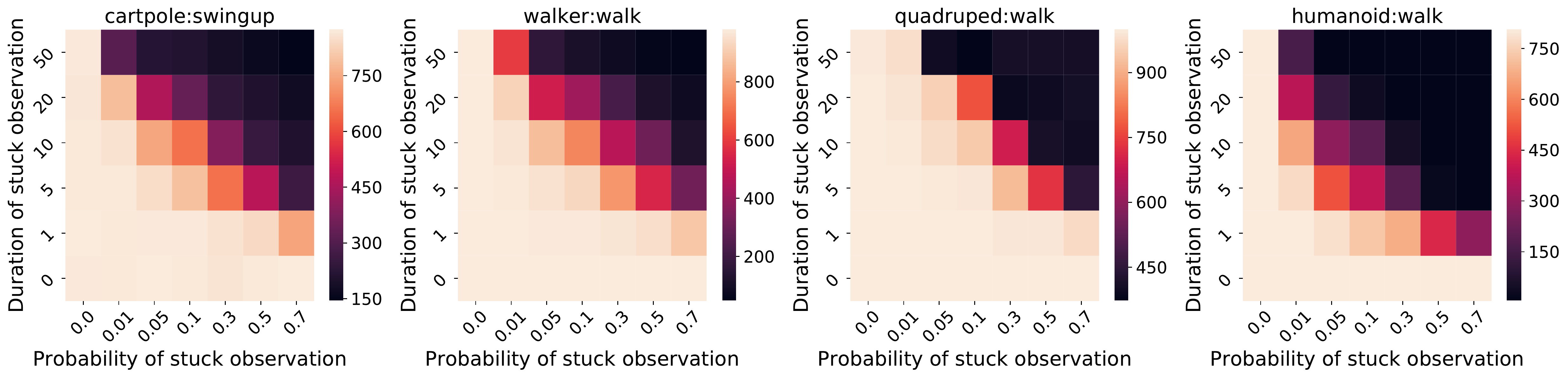}
  \caption{DMPO}\label{fig:dmpo_stuck}    
  \end{subfigure}    
  \caption{Average performance and standard deviation on the four tasks under the stuck sensors condition. Both the probability of a sensor becoming stuck and the number of steps it is stuck at the last value for are varied.}
\end{figure}

\begin{figure}[!htb]
  \centering
  \begin{subfigure}{.95\textwidth}
  \centering  
  \includegraphics[width=\textwidth]{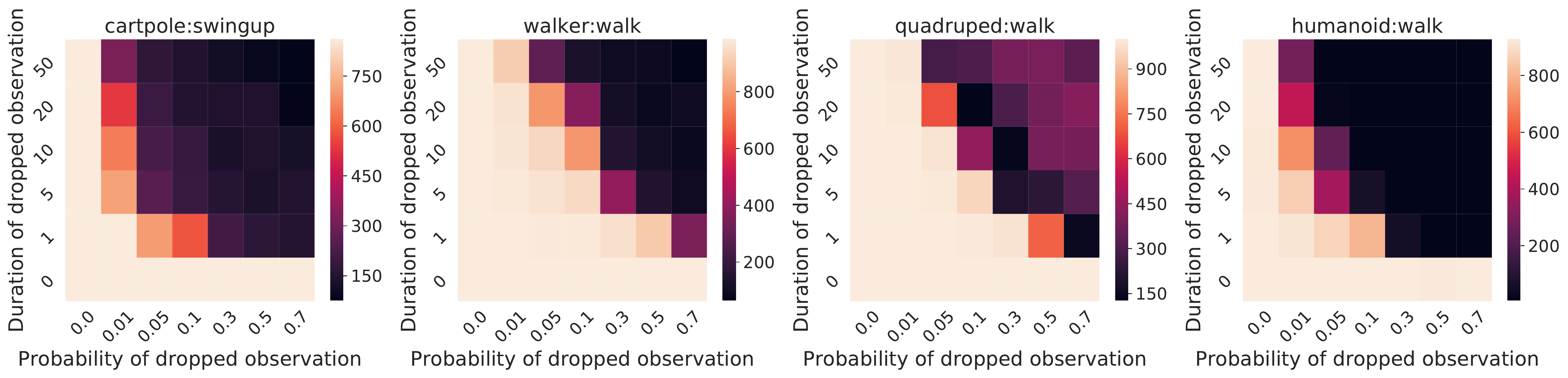}
  \caption{D4PG}\label{fig:d4pg_dropped}
  \end{subfigure}  
  \begin{subfigure}{.95\textwidth}
  \centering  
  \includegraphics[width=\textwidth]{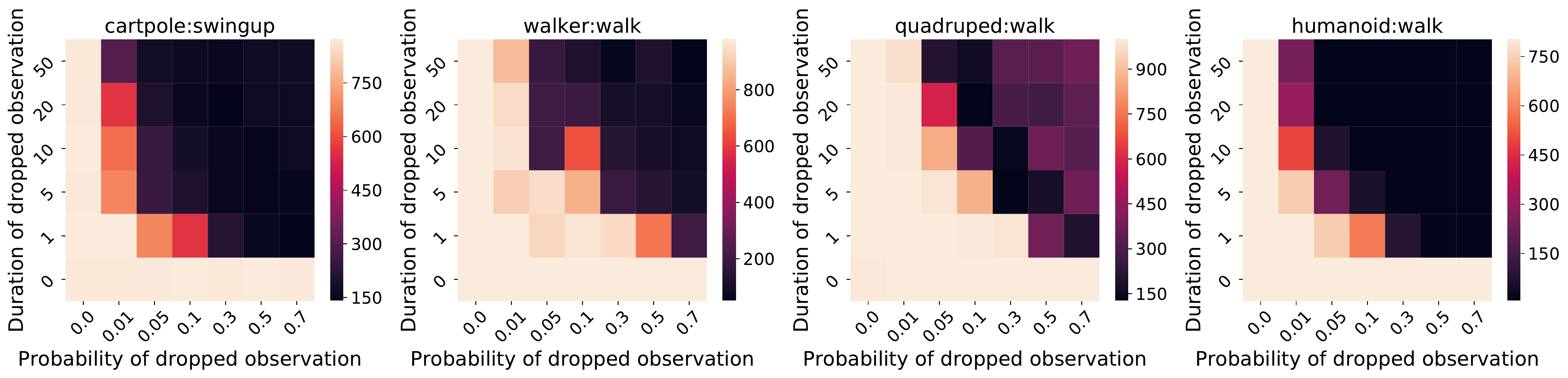}
  \caption{DMPO}\label{fig:dmpo_dropped}
  \end{subfigure}  
  \caption{Average performance on the four tasks under the dropped sensors condition. Both the probability of a sensor being dropped and the number of steps it is dropped for are varied.}
\end{figure}

\textit{Experimental Setup \& Results} Many real-world sensor issues can be viewed as a partial observability challenge (unobserved properties describing the functioning of the sensor) that could be helped by recurrent models or other approaches for partial observability. A common issue we see in real-world settings is malfunctioning sensors. On any real task, we can assume that the sensors are noisy, which we reproduce by adding increasing levels of Gaussian noise to the actions and observations.  Results of these perturbations can be observed in Figures~\ref{fig:d4pg_noise} and \ref{fig:dmpo_noise} (left and middle figures respectively) for D4PG and DMPO.  We frequently also see sensors that either get stuck at a certain value for a period of time or drop out entirely, with some default value being sent to the agent. We simulate both of these scenarios by setting both a probability of a sensor being stuck or dropped and varying the length of the malfunction being. Results for these perturbations are presented in Figures~\ref{fig:d4pg_stuck}, \ref{fig:dmpo_stuck} and Figures~\ref{fig:d4pg_dropped}, \ref{fig:dmpo_dropped} for stuck and dropped sensors.  We see from the figures that both dropped and stuck sensors have a significant effect on degrading the final performance. 

\paragraph{Non-stationarity.} 
Real world systems are often stochastic and noisy compared to most simulated environments. In addition, sensor and action noise as well as action delays add to the perturbations an agent may experience in the real-world setting. There are a number of RL approaches that have been utilized to ensure that an agent is robust to different subsets of these factors. We will focus on Robust MDPs, domain randomization and system identification as frameworks for reasoning about noisy, non-stationary systems.

A Robust MDP is defined by a tuple $\langle \S, \A, \mathcal{P}, r, \gamma \rangle$ where $S,A,r$ and $\gamma$ are as previously defined; $\mathcal{P}$ is a set of transition matrices referred to as the uncertainty set \citep{iyengar2005robust}. The objective that we optimize is the worst-case value function defined as:
\begin{equation}
\label{eq:robustness}
J(\pi)=\inf_{p \in \mathcal{P}}\mathbb{E}^p \biggl[\sum_{t=0}^\infty \gamma^t r_t \vert \mathcal{P}, \pi \biggr].
\end{equation}
At each step, nature chooses a transition function that the agent transitions with so as to minimize the long term value. The agent learns a policy that maximizes this worst case value function. Recently, a number of works have surfaced that have shown this formulation to yield robust policies that are agnostic to a range of perturbations in the environment \citep{tamar2014scaling,mankowitz2018learning,shashua2017deep,derman2018soft,Derman2019,Mankowitz2019b}. The solutions do tend to be overly conservative but some work has been done to yield less conservative, `soft-robust' solutions \citep{derman2018soft}.

In addition to the robust MDP formalism, the practitioner may be interested in both robustness due to domain randomization and system identification. Domain randomization~\citep{peng2018sim} involves explicitly training an agent on various perturbations of the environment and averaging these learning errors together during training. System identification involves training a policy that, once on a new system, can determine  the characteristics of the environment it is operating in and modify its policy accordingly \citep{finn2017model, nagabandi}.

\begin{figure}[!h]
    \centering
    \begin{minipage}{.45\textwidth}
        \centering
        \includegraphics[width=\linewidth ]{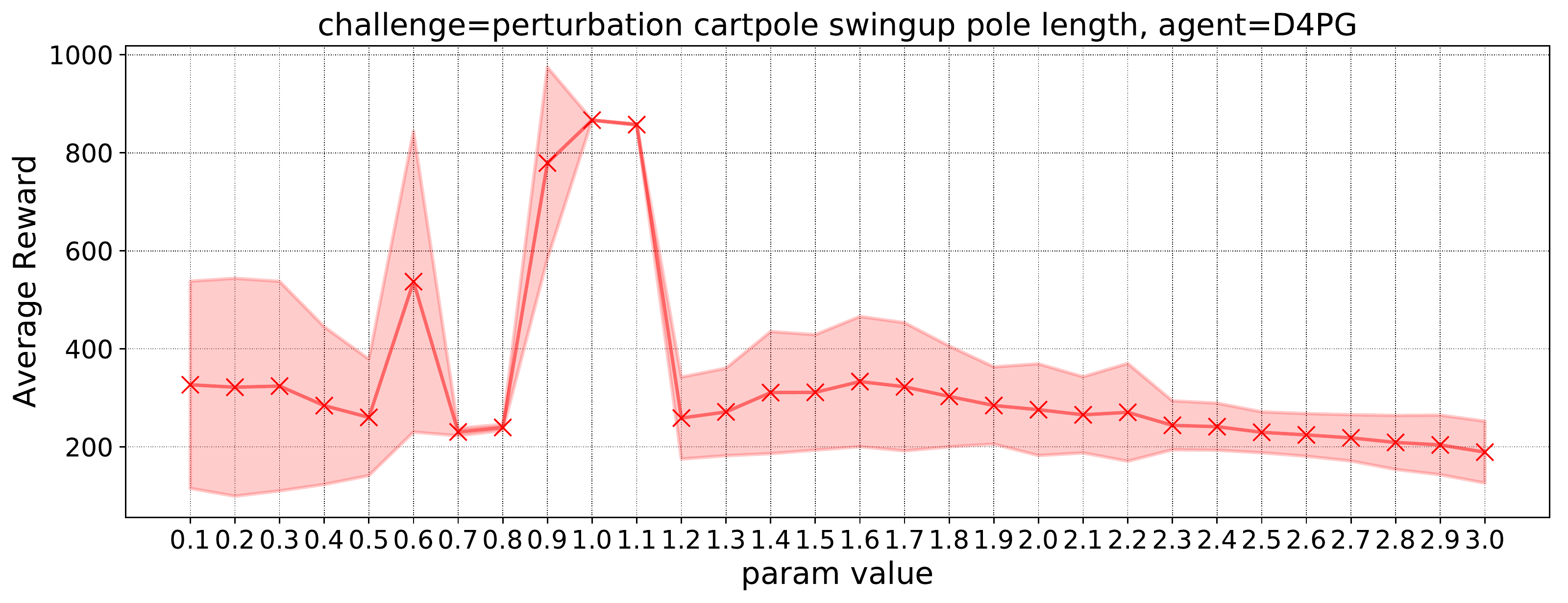}
    \end{minipage}
    \begin{minipage}{0.45\textwidth}
        \centering
        \includegraphics[width=\linewidth ]{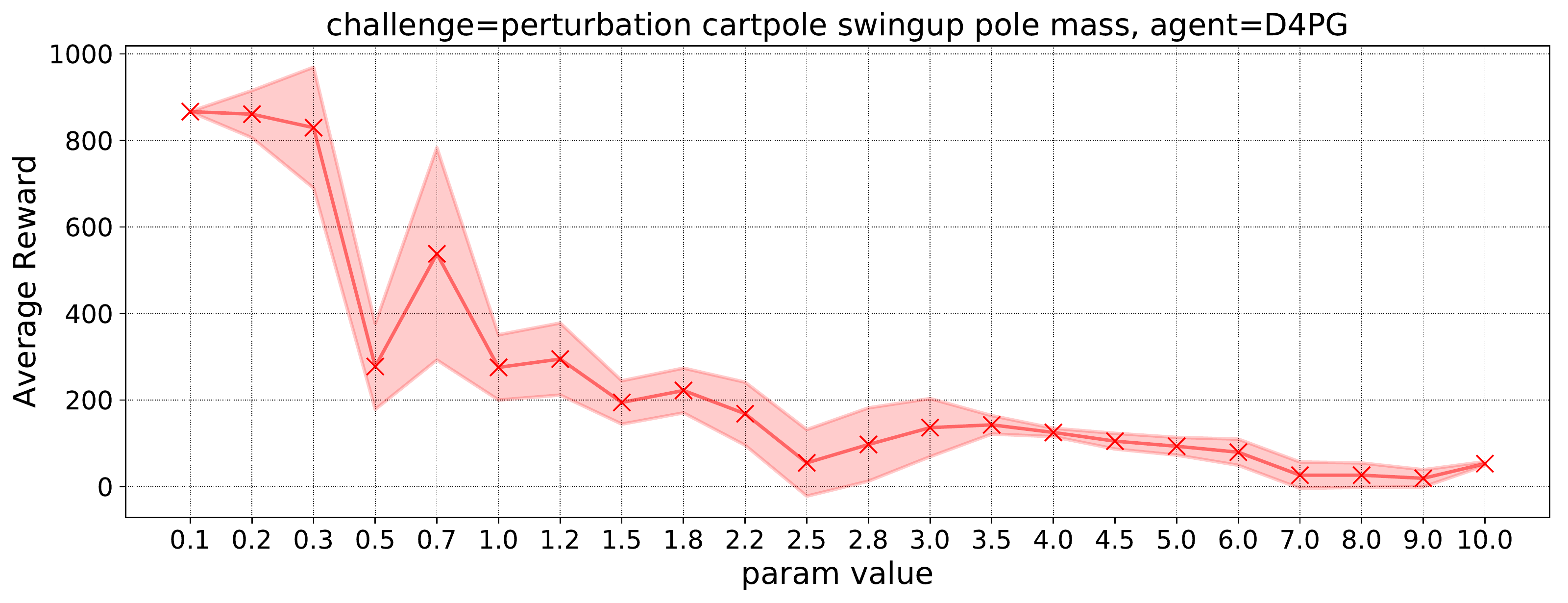}
    \end{minipage}
    \begin{minipage}{0.45\textwidth}
        \centering
        \includegraphics[width=\linewidth ]{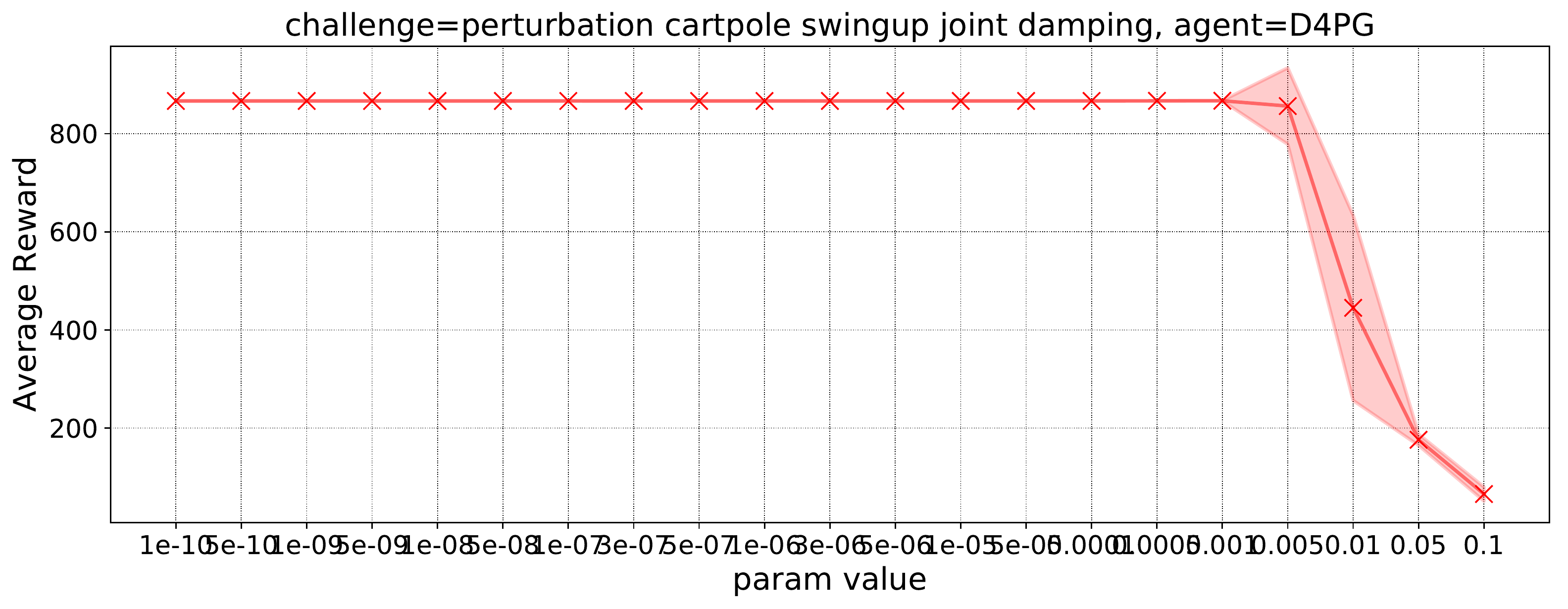}
    \end{minipage}
    \begin{minipage}{0.45\textwidth}
        \centering
        \includegraphics[width=\linewidth ]{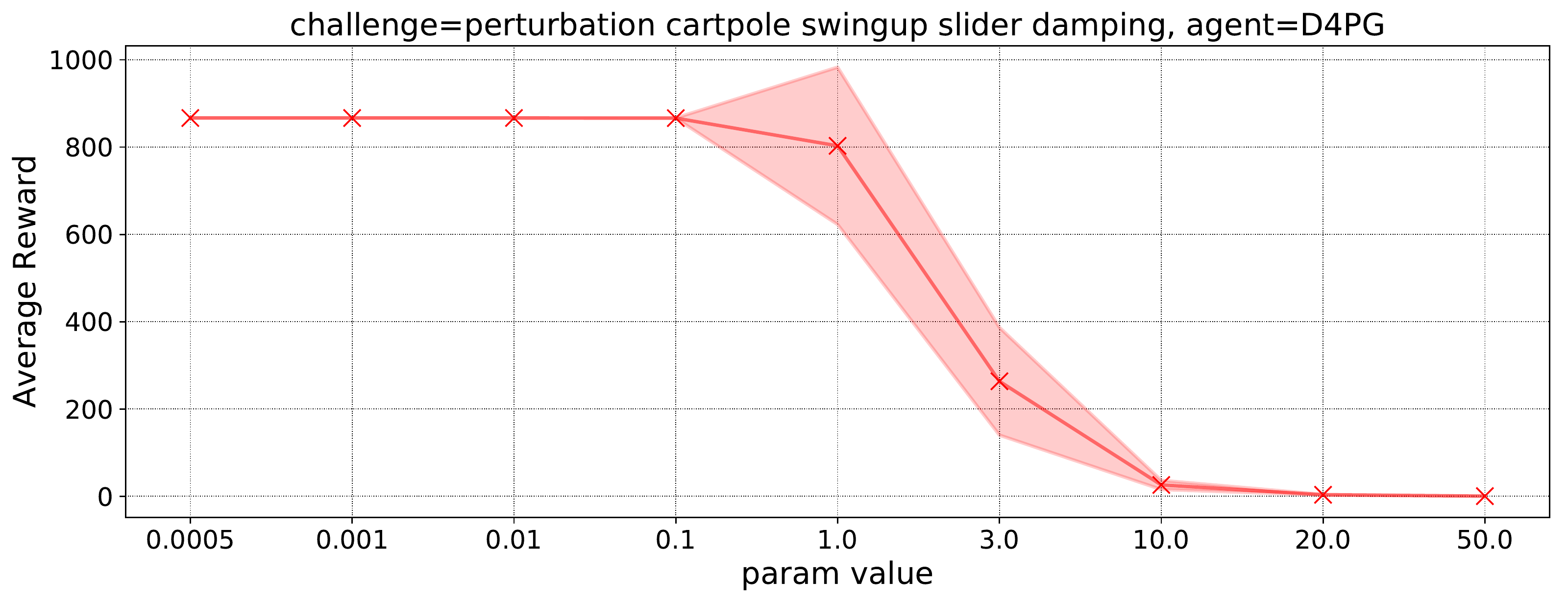}
    \end{minipage}
    \begin{minipage}{.45\textwidth}
        \centering
        \includegraphics[width=\linewidth ]{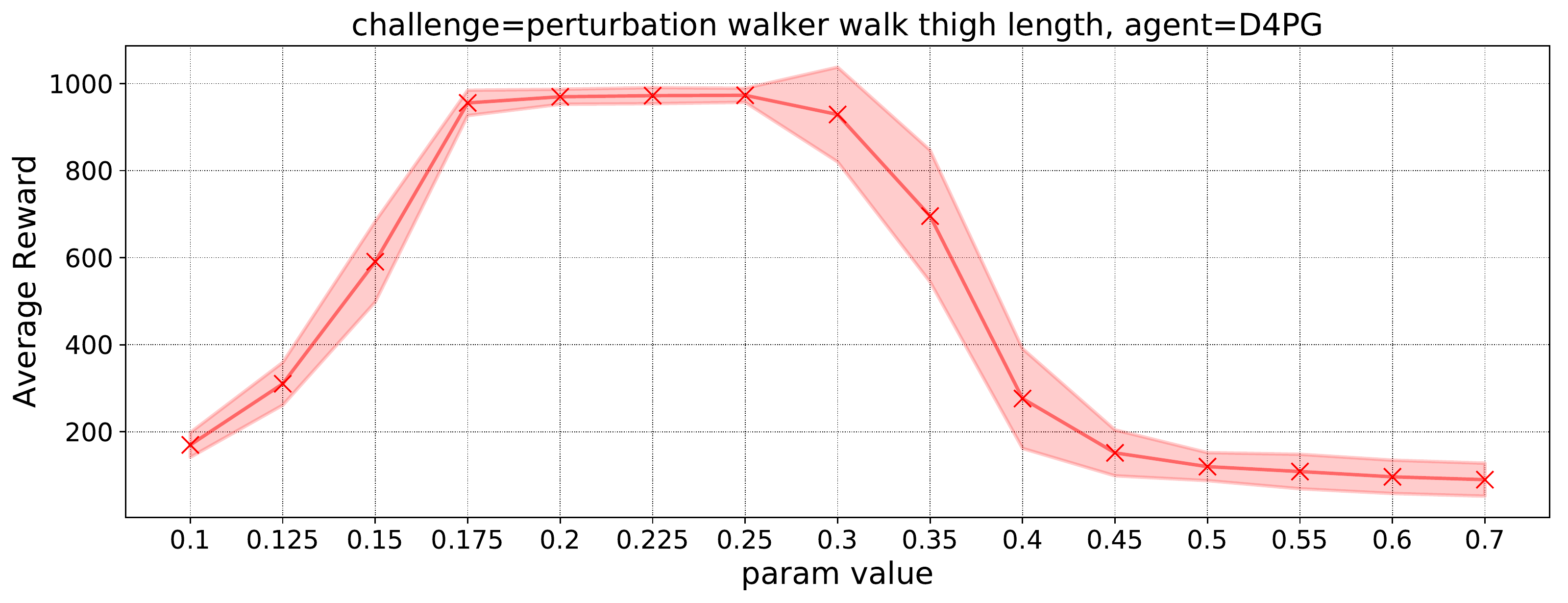}
    \end{minipage}%
    \begin{minipage}{0.45\textwidth}
        \centering
        \includegraphics[width=\linewidth ]{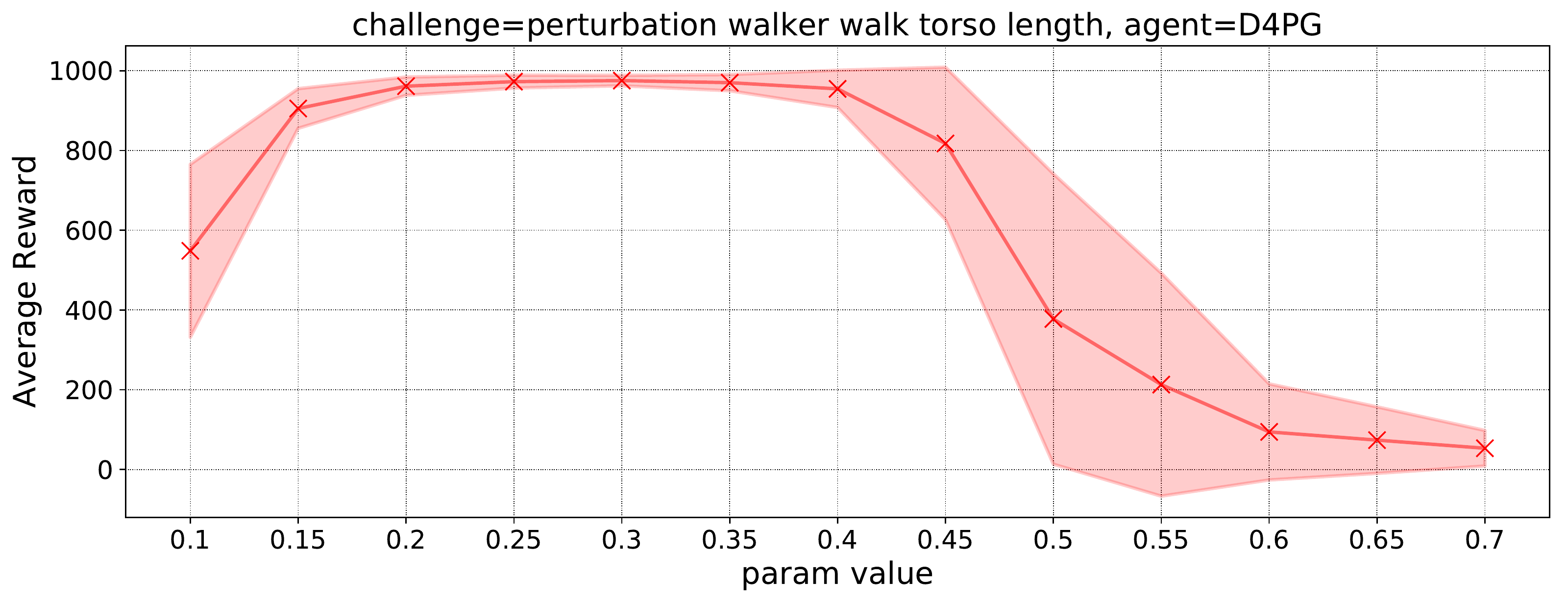}
    \end{minipage}
    \begin{minipage}{0.45\textwidth}
        \centering
        \includegraphics[width=\linewidth ]{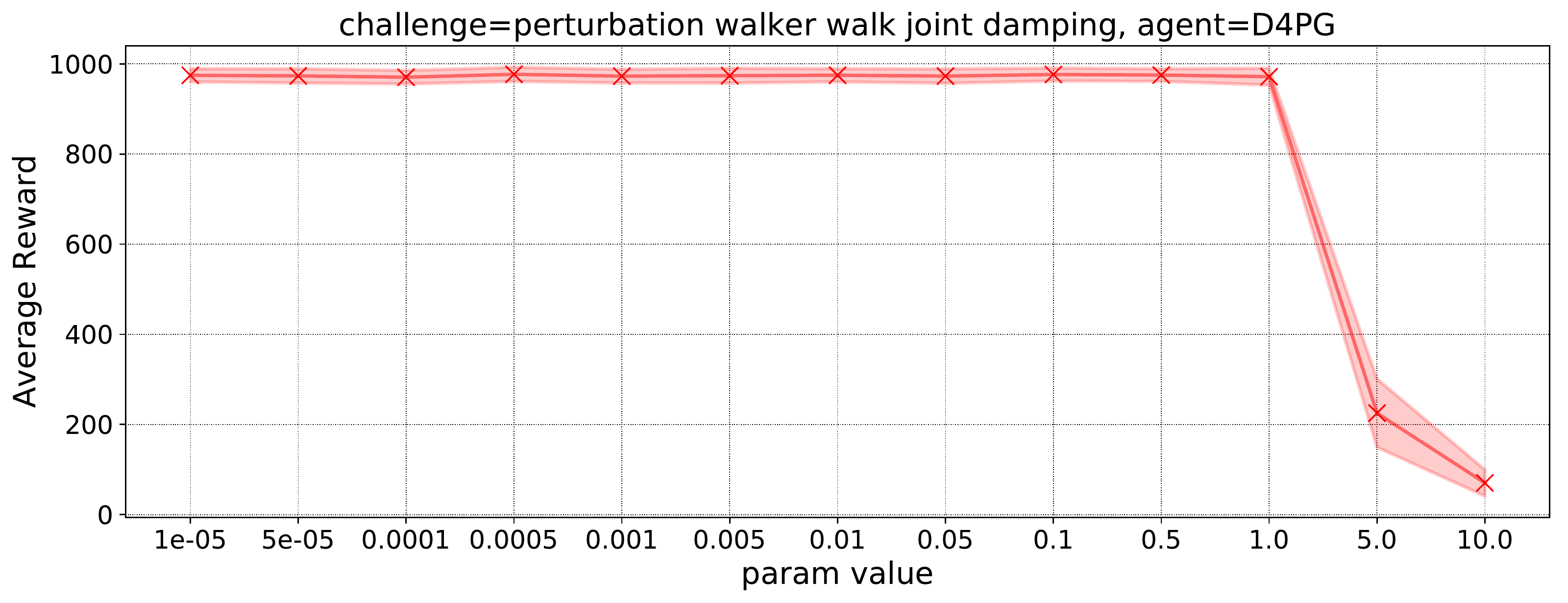}
    \end{minipage}
    \begin{minipage}{0.45\textwidth}
        \centering
        \includegraphics[width=\linewidth ]{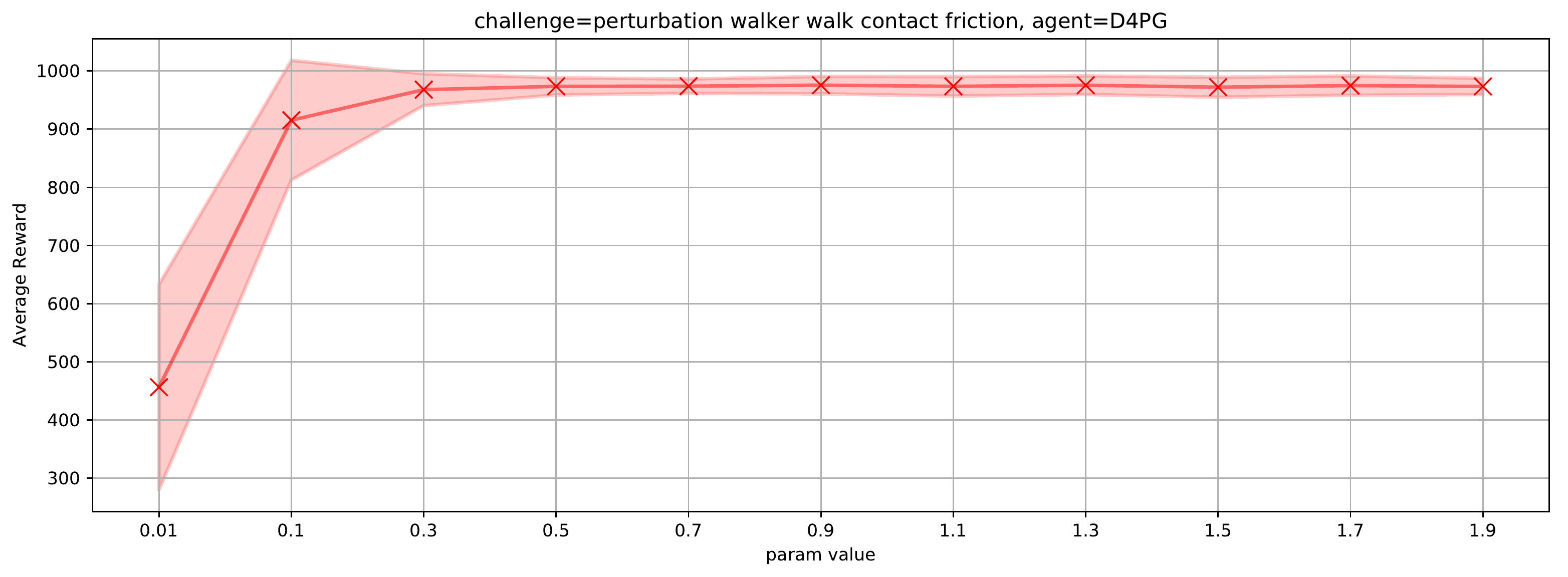}
    \end{minipage}    
    \begin{minipage}{0.45\textwidth}
        \centering
        \includegraphics[width=\linewidth ]{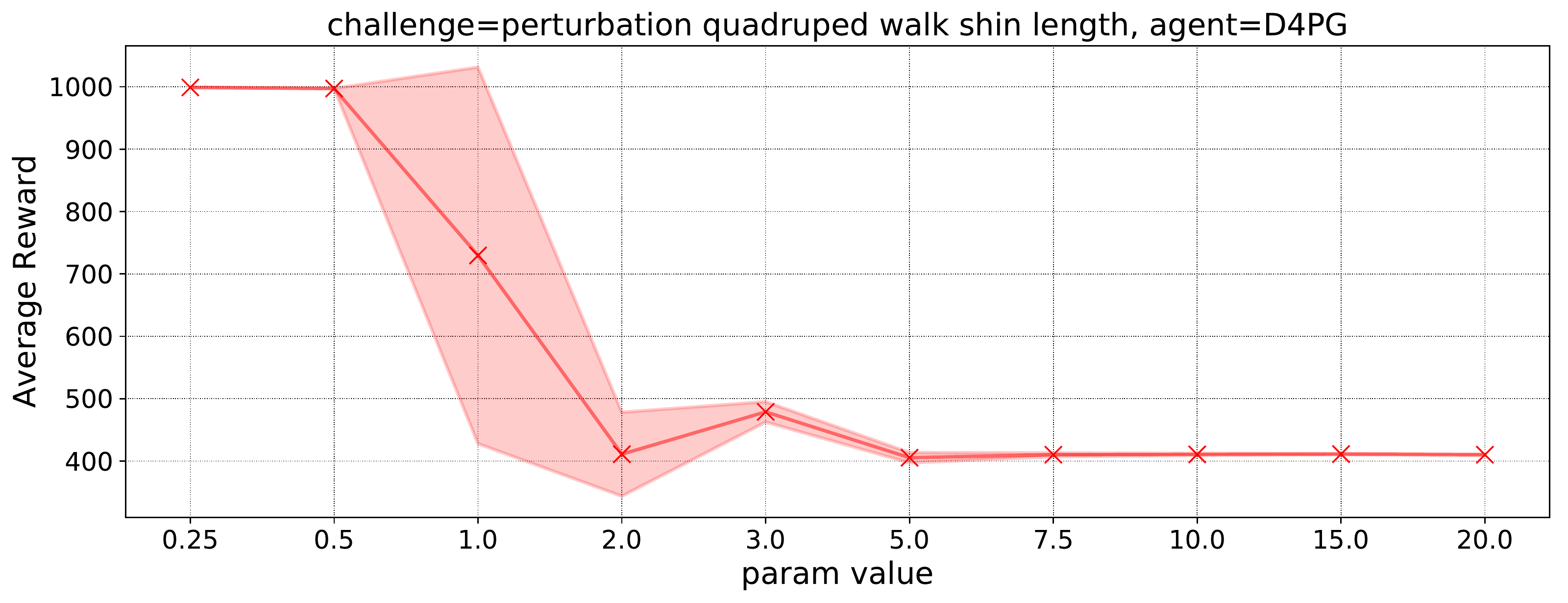}
    \end{minipage}
    \begin{minipage}{0.45\textwidth}
        \centering
        \includegraphics[width=\linewidth ]{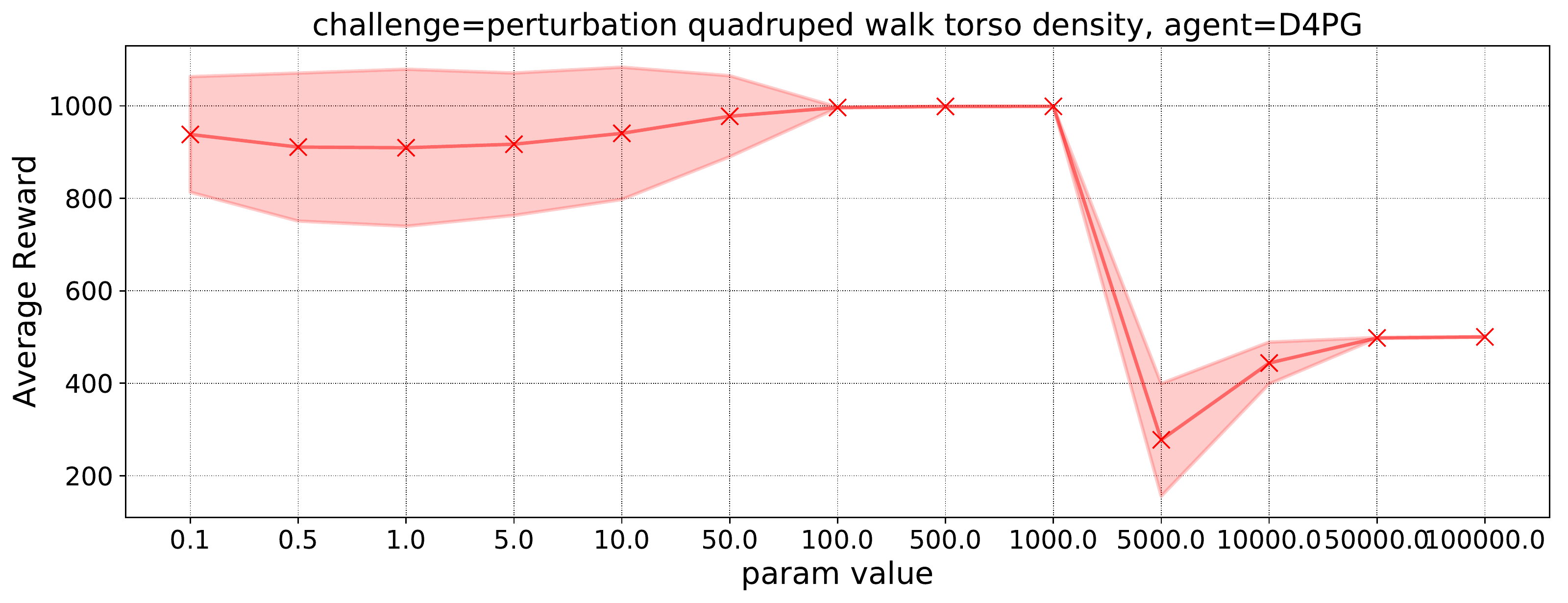}
    \end{minipage}
    \begin{minipage}{0.45\textwidth}
        \centering
        \includegraphics[width=\linewidth ]{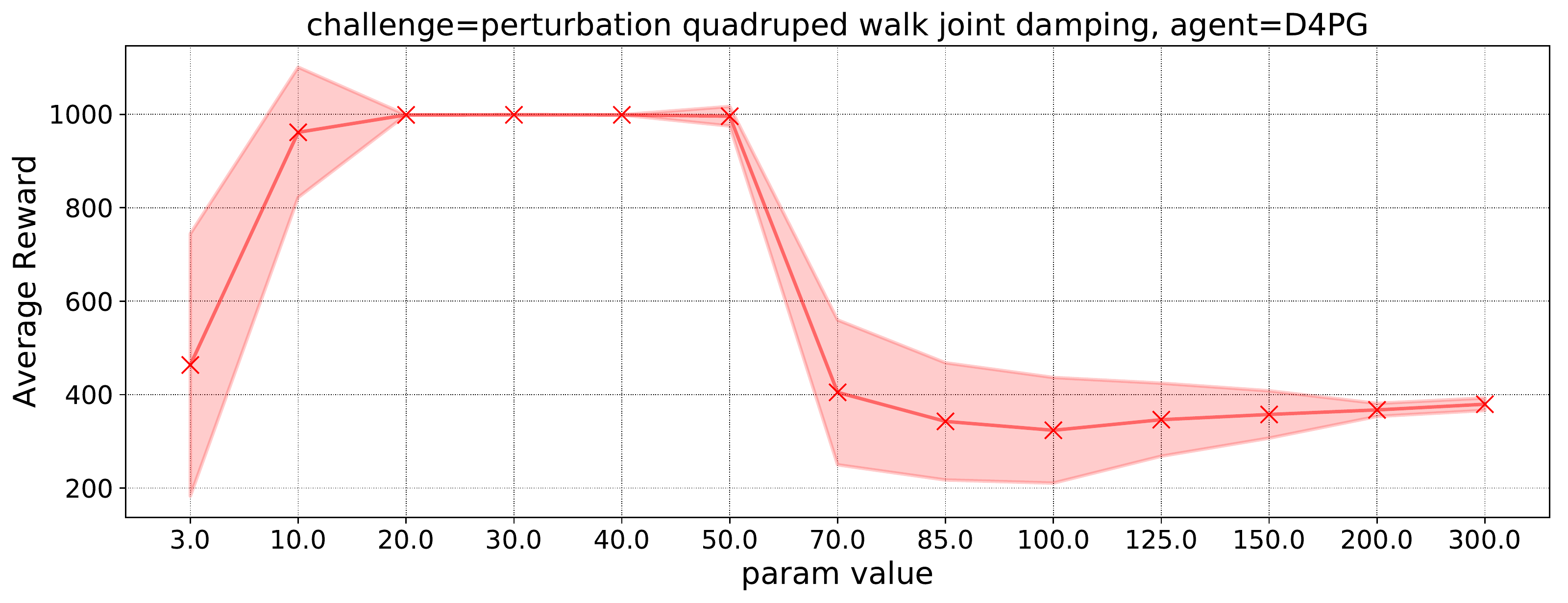}
    \end{minipage}
    \begin{minipage}{0.45\textwidth}
        \centering
        \includegraphics[width=\linewidth ]{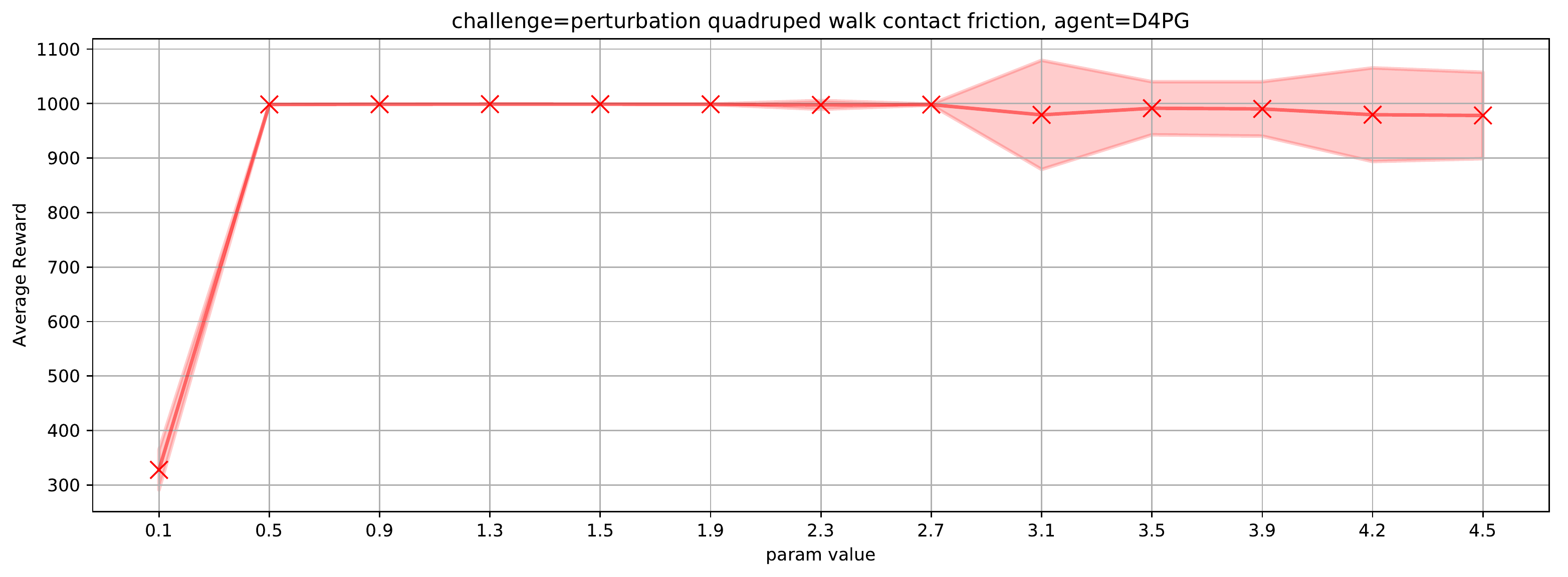}
    \end{minipage}    
    \begin{minipage}{0.45\textwidth}
        \centering
        \includegraphics[width=\linewidth ]{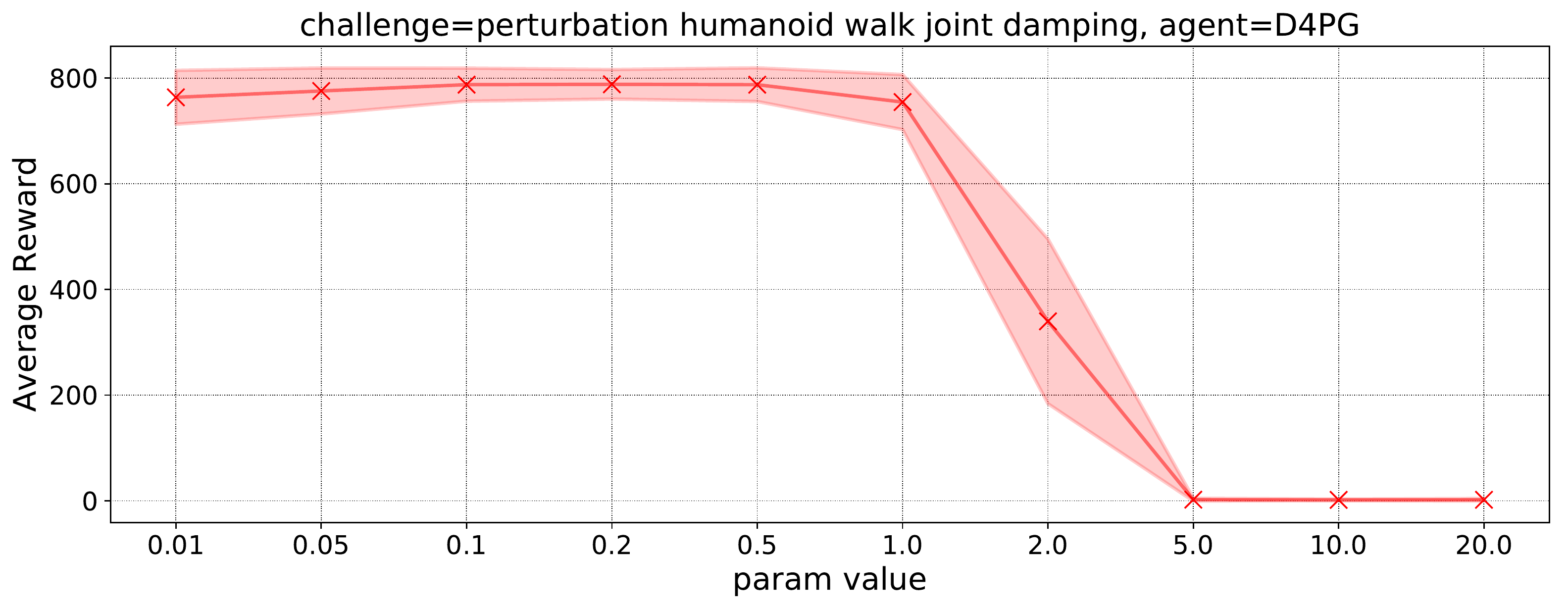}
    \end{minipage}
    \begin{minipage}{0.45\textwidth}
        \centering
        \includegraphics[width=\linewidth ]{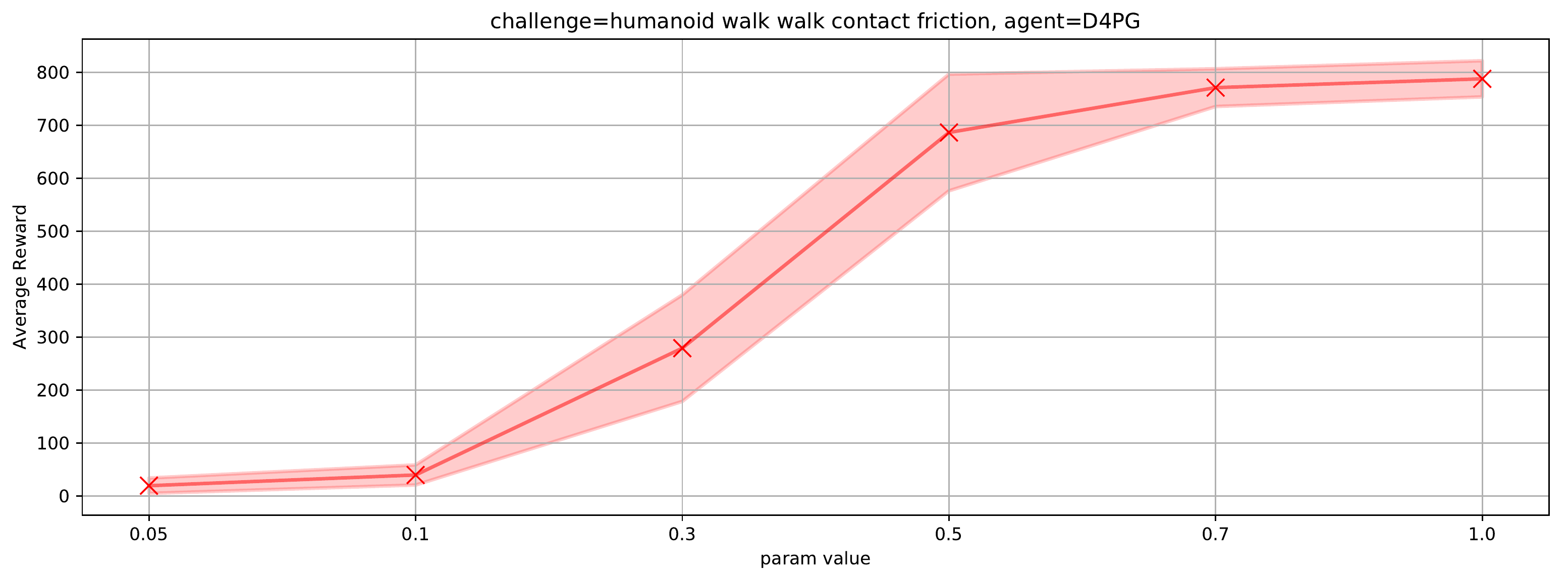}
    \end{minipage}    
    \begin{minipage}{0.45\textwidth}
        \centering
        \includegraphics[width=\linewidth ]{figures/D4PG/non-stationarity/converged_policy102_DISK.pdf}
    \end{minipage}    
    \caption{Perturbation effects on a converged D4PG policy due to varying specific environment parameters.}
    \label{fig:nonstationary_disk}    
\end{figure}

\textit{Experimental Setup \& Results} We perform a number of different experiments to determine the effects of non-stationarity. We first want to determine whether perturbations to the environment can have an effect on a converged policy that is trained without any challenges added to the environment. For each of the domains, we perturb each of the supported parameters shown in Table \ref{tab:perturb_env_full}. The effect of the perturbations on the converged D4PG policy for each domain and supported parameter can be seen in Figure \ref{fig:nonstationary_disk}. It is clear that varying the perturbations does indeed have an effect on the performance of the converged policy; in many instances this causes the converged policy to completely fail. This is consistent with the results in \cite{Mankowitz2019b}. This hyperparameter sweep also helps determine which parameter settings are more likely to have an effect on the learning capabilities of the agent during training.

The second set of experiments therefore aim to determine the consequences of incorporating non-stationarity effects during training. Every episode, new environment parameters are sampled between a $[perturb_{min}, perturb_{max}]$ where $perturb_{min}$ and $perturb_{max}$ indicate the minimum and maximum perturbation values of a particular parameter that we vary. For example, in \cartpoleswingup, the perturbation parameter is pole length and $perturb_{min}=0.5$, $perturb_{max}=3.0$ and the variance used for sampling is $perturb_{std}=0.05$.

Based on the previous set of experiments, for each task, we select domain parameters that we expect may change the optimal policy. We perform four hyperparameter training sweeps on the domain parameters for each domain \& each algorithm (D4PG and DMPO). These sweeps are in increasing orders of difficulty and have thus been named $\texttt{diff}_1$, $\texttt{diff}_2$, $\texttt{diff}_3$, $\texttt{diff}_4$ and are shown in Table \ref{tab:perturb_parameters}. We perturb the environment in two different ways: uniform and cyclic perturbations. For uniform perturbations, we sample each episode from a uniform distribution  and for the cyclic perturbations, a random positive change was sampled from a normal distribution, and the values were reset to the lower limit once the upper limit had been reached. Additional sampling methods and perturbation parameters are supported in the \suite and can also be seen in Table \ref{tab:perturb_env_full}. Cycle sampling simulates scenarios of equipment degrading over time until being replaced or fixed and returning to peak performance. The slow consistent changes over episodes also enables for the possibility of an algorithm adapting to the changes over time. 

\begin{figure}[!h]
    \centering
    \begin{subfigure}{\textwidth}
    \centering
        \includegraphics[width=0.48\textwidth]{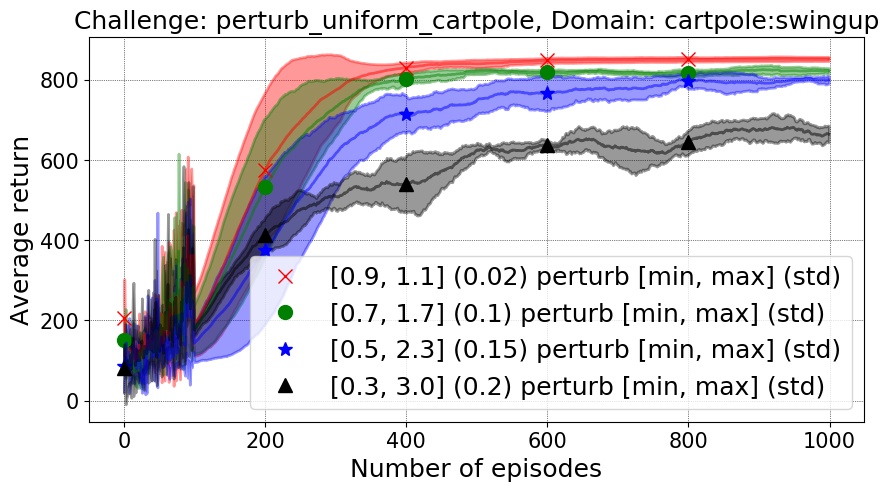}
        \includegraphics[width=0.48\textwidth]{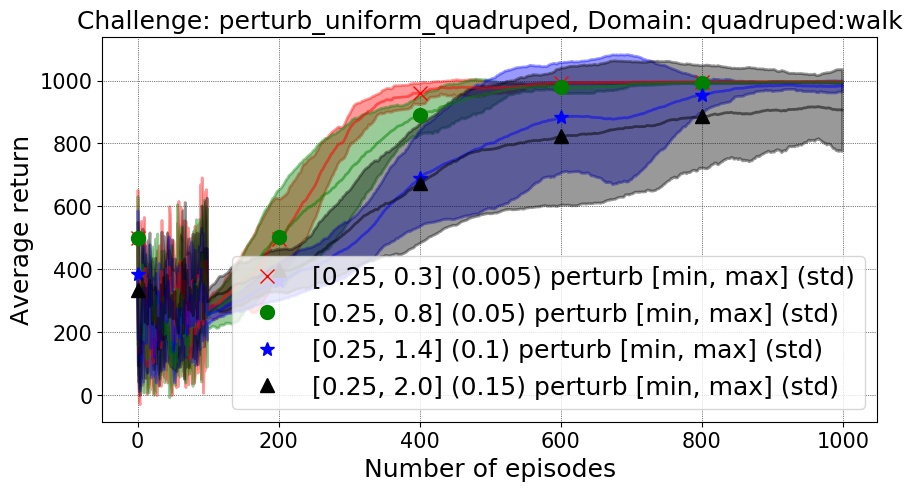}
        \includegraphics[width=0.48\textwidth]{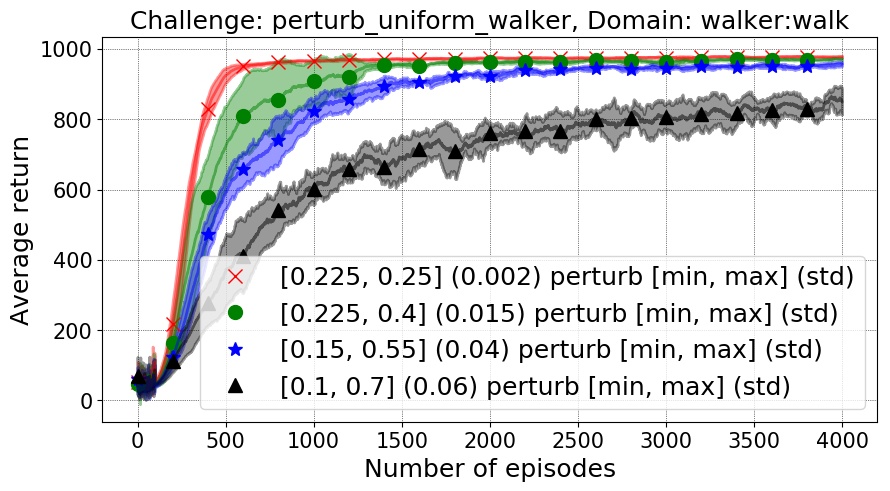}
        \includegraphics[width=0.48\textwidth]{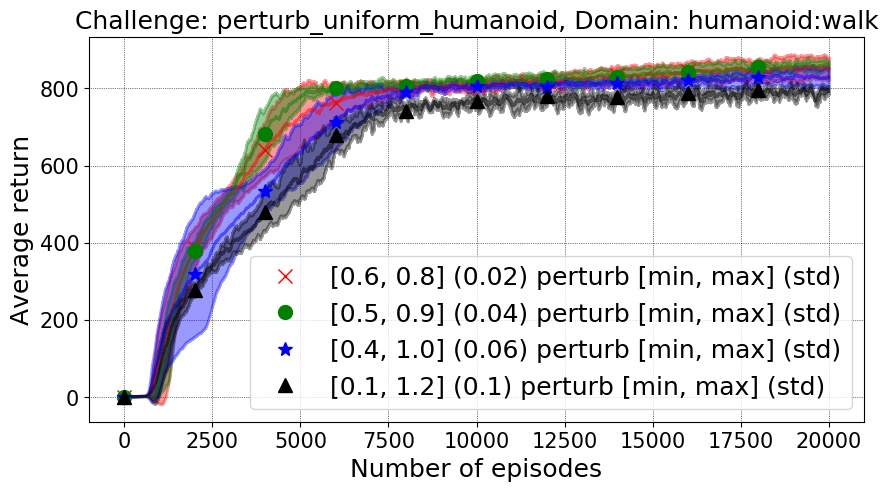}
    \caption{D4PG}
    \end{subfigure}    
    \begin{subfigure}{\textwidth}
    \centering
        \includegraphics[width=0.48\textwidth]{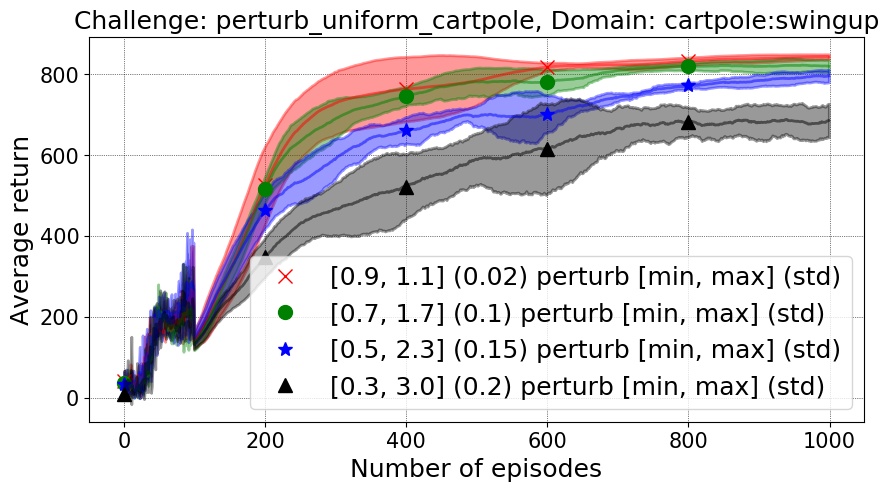}
        \includegraphics[width=0.48\textwidth]{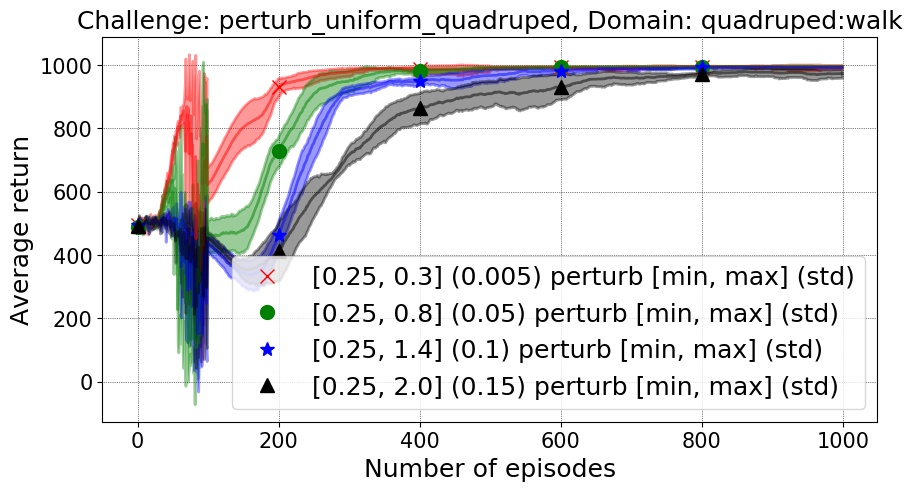}
        \includegraphics[width=0.48\textwidth]{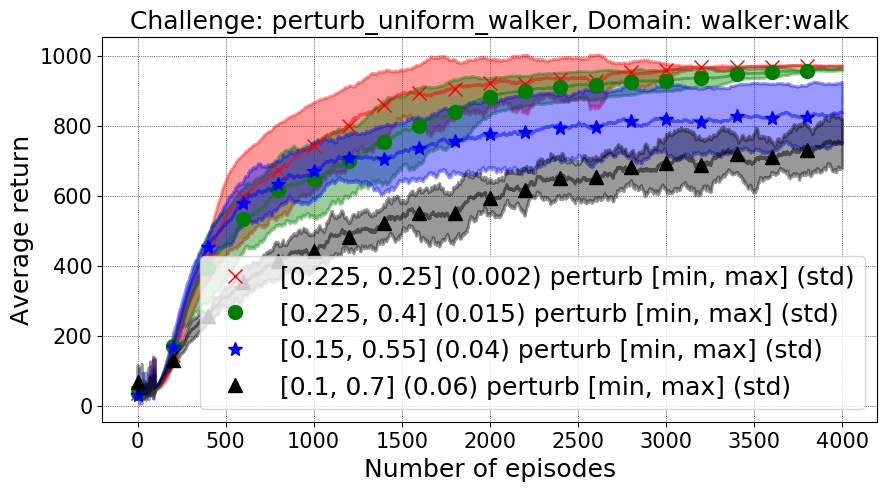}
        \includegraphics[width=0.48\textwidth]{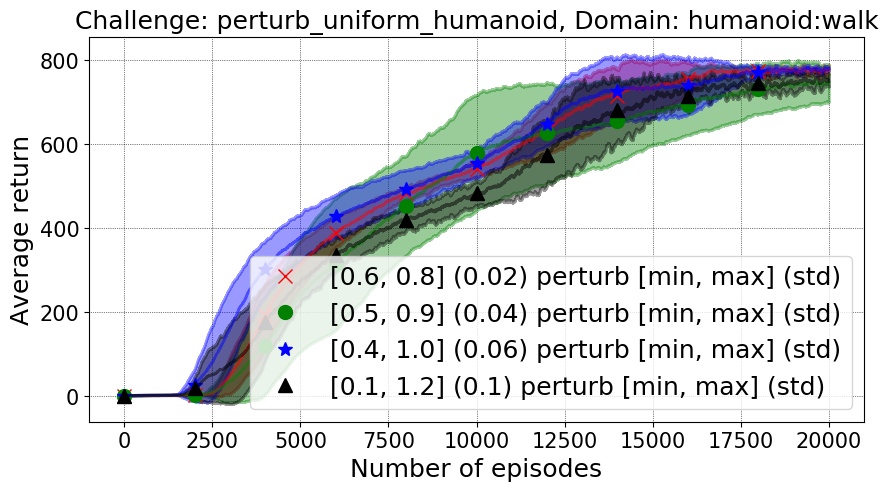}
    \caption{DMPO}            
    \end{subfigure}
    \caption{Uniform perturbations applied per episode for each of the four domains for D4PG and DMPO.}
    \label{fig:nonstationary_uniform}    
\end{figure}

\begin{figure}[!h]
    \centering
    \begin{subfigure}{\textwidth}
    \centering
        \includegraphics[width=0.48\textwidth]{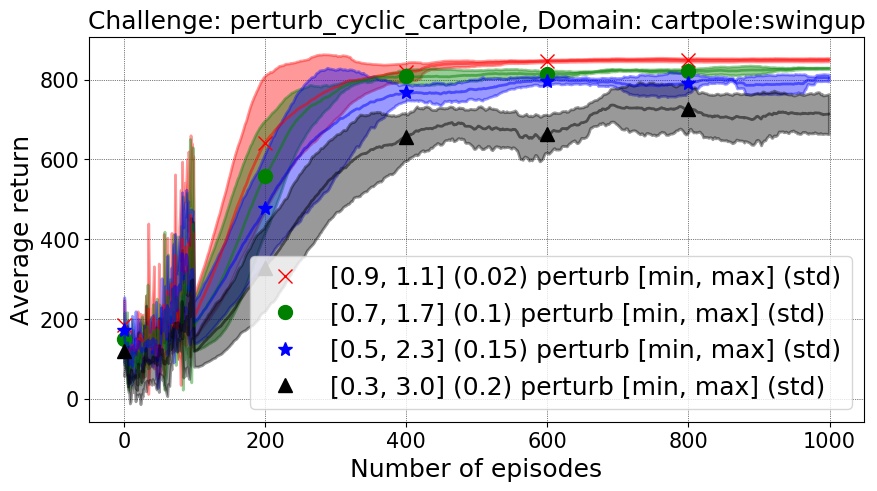}
        \includegraphics[width=0.48\textwidth]{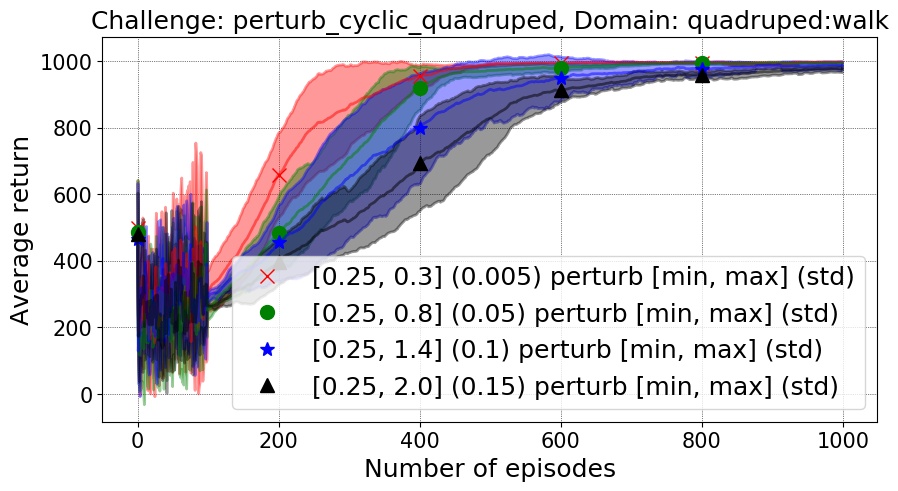}
        \includegraphics[width=0.48\textwidth]{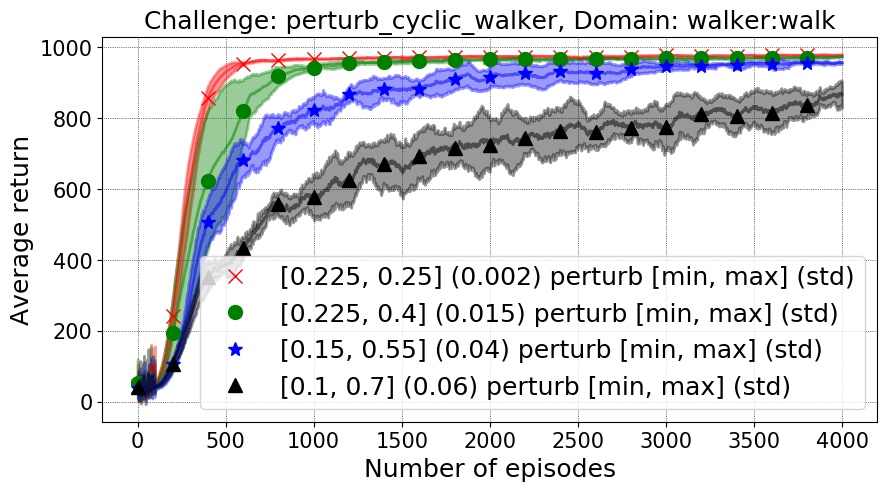}
        \includegraphics[width=0.48\textwidth]{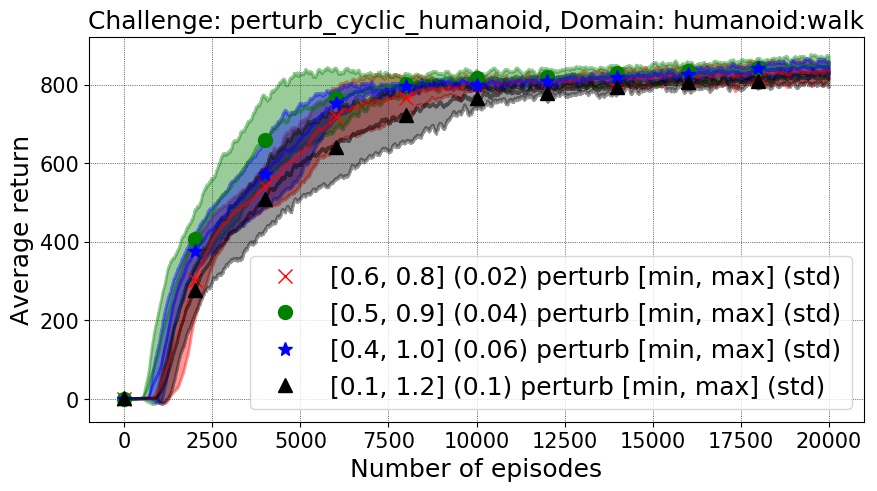}
    \caption{D4PG}
    \end{subfigure}    
    \begin{subfigure}{\textwidth}
    \centering
        \includegraphics[width=0.48\textwidth]{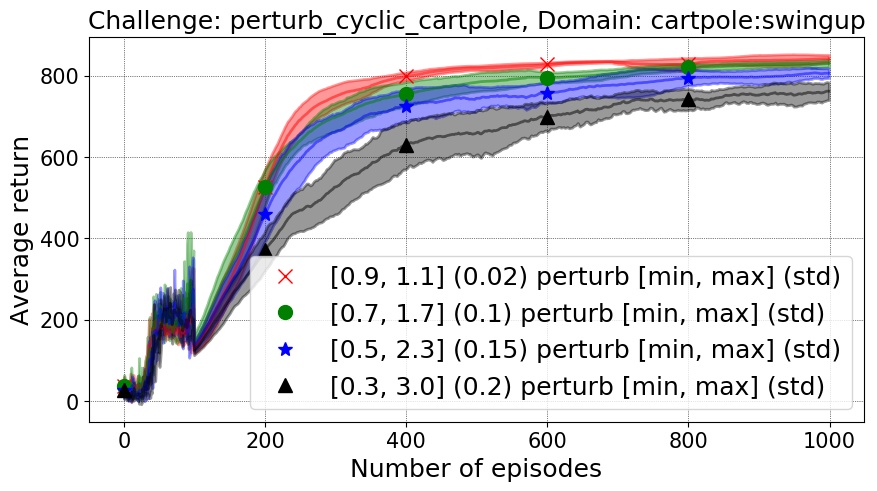}
        \includegraphics[width=0.48\textwidth]{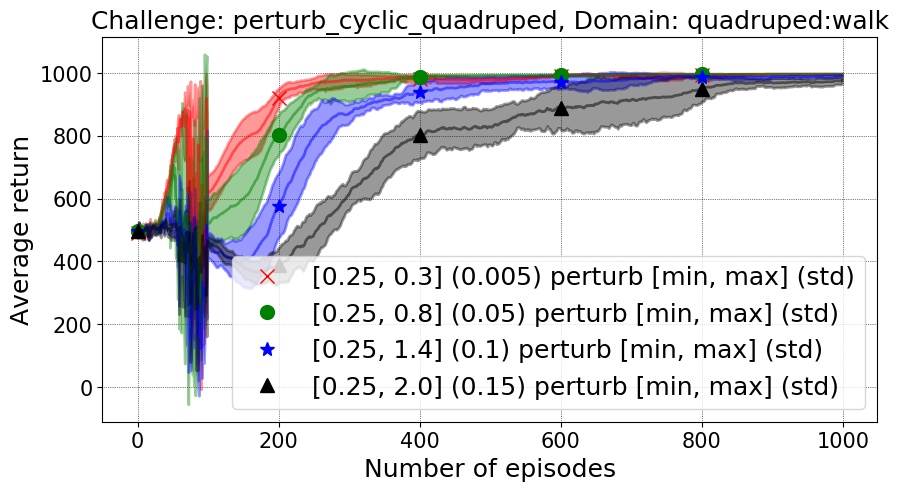}
        \includegraphics[width=0.48\textwidth]{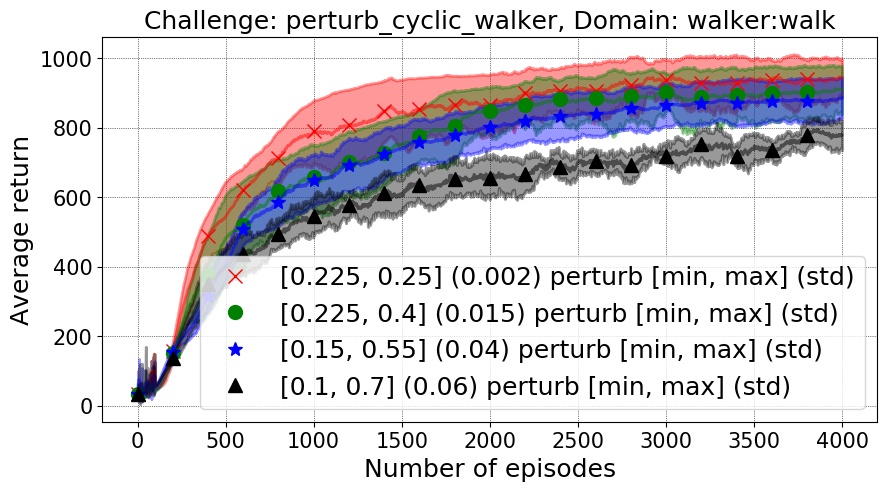}
        \includegraphics[width=0.48\textwidth]{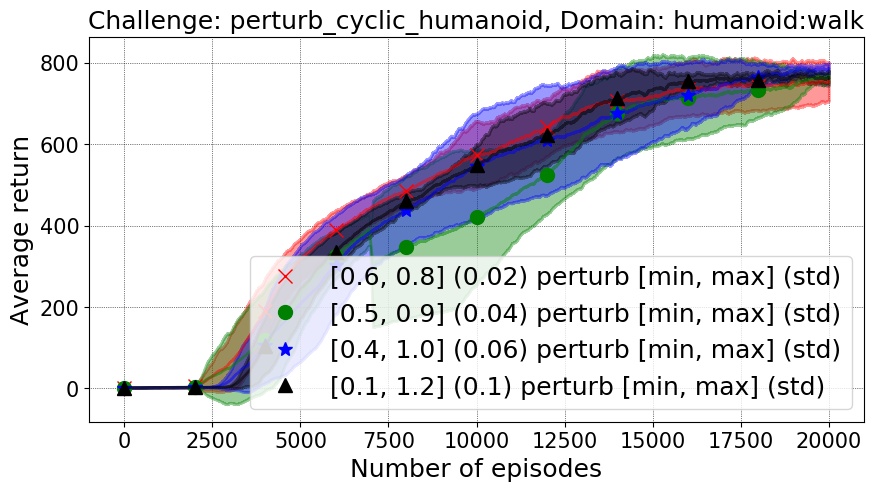}
    \caption{DMPO}            
    \end{subfigure}
    \caption{Cyclic perturbations applied per episode for each of the four domains for D4PG and DMPO.}
    \label{fig:nonstationary_cyclic}    
\end{figure}

Figures \ref{fig:nonstationary_uniform} and \ref{fig:nonstationary_cyclic} show the training performance for D4PG and DMPO when applying uniform and cyclic perturbations per episode respectively. As seen in the figures, increasing the range of the perturbation parameter has the effect of slowing down learning. This seems to be consistent across all of the domains we evaluated. 

\begin{table}[h]
\begin{center}
\begin{tabular}{|l|l|}
\hline
\textbf{Env.} & \textbf{Supported Parameters}\\
\hline
Cart-Pole & Pole length \\
          & Pole mass \\
          & Joint damping \\
          & Slider damping \\
\hline
Walker & Thigh length \\
      & Torso length \\
      & Joint damping \\
      & Contact friction \\
\hline
Quadruped & Shin length \\
      & Torso density \\
      & Joint damping \\
      & Contact friction \\
\hline
Humanoid &  Joint Damping \\
  & Contact Friction \\
  & Head Size \\
\hline
\end{tabular}
\end{center}
\caption{Supported perturbed parameters for each of the control tasks.}
\label{tab:perturb_env_full}
\vspace{-0.5in}
\end{table}

\begin{table}[h]
\begin{center}
\begin{tabular}{|l |l |l |l| l|}
\hline
\textbf{Env.} & \textbf{$Perturb_{min}$} & \textbf{$Perturb_{max}$} & \textbf{$Perturb_{std}$} & \textbf{Default Value} \\
\hline
Cart-Pole &  & & & \\
Parameter & pole\_length&  & &\\
$\texttt{diff}_1$         & 0.9& 1.1& 0.02 & 1.0 \\
$\texttt{diff}_2$       & 0.7& 1.7& 0.1& 1.0  \\
$\texttt{diff}_3$   & 0.5& 2.3 & 0.15 & 1.0  \\
$\texttt{diff}_4$   & 0.3& 3.0& 0.2&1.0  \\
\hline
Walker &  &  & &\\
Parameter &thigh\_length & & &\\
$\texttt{diff}_1$          & 0.225 & 0.25 & 0.002 & 0.225 \\
$\texttt{diff}_2$       & 0.225& 0.4& 0.015&  0.225\\
$\texttt{diff}_3$         & 0.15& 0.55& 0.04& 0.225  \\
$\texttt{diff}_4$     &0.1 & 0.7& 0.06& 0.225  \\
\hline
Quadruped &  &  & &\\
Parameter & shin\_length&  & &\\
$\texttt{diff}_1$          & 0.25& 0.3 & 0.005& 0.25 \\
$\texttt{diff}_2$        & 0.25& 0.8& 0.05& 0.25  \\
$\texttt{diff}_3$          & 0.25& 1.4& 0.1& 0.25 \\
$\texttt{diff}_4$     & 0.25& 2.0& 0.15& 0.25  \\
\hline
Humanoid &  &  & &\\
Parameter & join\_damping&  & &\\
$\texttt{diff}_1$          &0.6 & 0.8 & 0.02 & 0.1 \\
$\texttt{diff}_2$        & 0.5& 0.9& 0.04& 0.1  \\
$\texttt{diff}_3$          & 0.4& 1.0& 0.06& 0.1 \\
$\texttt{diff}_4$     & 0.1& 1.2& 0.1&0.1  \\
\hline
\end{tabular}
\end{center}
\caption{Perturbed parameters chosen for each control task, with varying levels of difficulty}
\label{tab:perturb_parameters}
\end{table}

% For our choice of parameter perturbations that we evaluated our agents on, we found that \humanoidwalk was most affected by non-stationary perturbations as seen in Appendix~\ref{app:mpo_results}, Figure~\ref{app:perturbations_mpo}. This is due to the high-dimensionality and relative instability of the task. Thus, even small perturbations to the task yield poor performance. In the other domains, we found that our hyperparameter settings were not significant enough to yield poor performance. These challenges can be made significantly more difficult by (1) increasing the size of the perturbations (without violating the physics of the domain), or (2) choosing a different perturbation to apply to the system. RL practitioners can specify the type of perturbation to apply as well as the magnitude range of the perturbation as seen in Appendix~\ref{app:codebase}. 

\clearpage
\subsection{Challenge 6: Multi-Objective Reward Functions}
\label{sec:multiobj}
% \begin{figure}[b]
%   \centering
%   \includegraphics[width=\textwidth]{figures/MPO/cartpole_multiobj.pdf}
%   \vspace{-0.5cm}
%   \caption{Left: Evolution of the task's base reward as  \texttt{multiobj\_coeff} is varied. Right: Average number of each violation in an episode upone convergence as \texttt{multiobj\_coeff} is varied.}
%   \label{fig:multiobj_cp}
% \end{figure}

\paragraph{Motivation \& Related Work} RL frames policy learning through the lens of optimizing a global reward function, yet most systems have multi-dimensional costs to be minimized. In many cases, system or product owners do not have a clear picture of what they want to optimize. When an agent is trained to optimize one metric, other metrics are often discovered that also need to be maintained or improved. Thus, a lot of the work on deploying RL to real systems is spent figuring out how to trade off between different objectives. 

There are many ways of dealing with multi-objective rewards: \cite{roijers2013survey} provide an overview of various approaches.  Various methods exist that deal explicitly with the multi-objective nature of the learning problems, either by predicting a value function for each objective~\citep{vanseijen}, or by finding a policy that optimizes each sub-problem~\citep{Li2019}, or that fits each Pareto-dominating mixture of objectives~\citep{Moffaert2014}. \cite{Yang2019} learn a general policy that can behave optimally for any desired mixture of objectives.  Multiple trivial objectives have been also used for enriching the reward signal to simply improve learning of the base task~\citep{Jaderberg2016}. \cite{abdolmaleki2020distributional} uses an expectation maximization approach to learn multiple Q-functions per objective. 

In the specific case of dealing with balancing a task reward with negative outcomes, a possible approach is to use a Conditional Value at Risk (CVaR) objective~\citep{Tamar2015}, which looks at a given percentile of the reward distribution, rather than expected reward. \citeauthor{Tamar2015} show that by optimizing reward percentiles, the agent is able to improve upon its worst-case performance. Distributional DQN~\citep{dabney18a,BellemareDM17} explicitly models the distribution over returns, and it would be straight-forward to extend it to use a CVaR objective.

When rewards can't be functionally specified, there are a number of works devoted to recovering an underlying reward function from demonstrations, such as inverse reinforcement learning~\citep{russell1998learning, ng2000algorithms, abbeel2004apprenticeship, ross2011reduction}. \citeauthor{menell} examine how to infer the truly intended reward function from the given reward function and training MDPs, to ensure that the agent performs as intended in new scenarios.

Because the global reward function is generally a balance of multiple sub-goals (e.g., reducing both time-to-target and energy use), a proper evaluation should separate the individual components of the reward function to better understand the policy's trade-offs.   Looking at the Pareto boundaries provides some insights to the relative trade-offs between objectives, but doesn't scale well beyond 2-3 objectives.
We propose a simple multi-objective analysis of return.  If we consider that the global reward function is defined as a linear combination of sub-rewards, $r(s,a) = \sum_{j=1}^K \alpha_j r_j(s,a)$, then we can consider the vector of per-component rewards for evaluation:
\begin{equation}
    \label{eq:multiobj}
    \bm{J}^{multi}(\pi) = \left(\sum_{i=1}^{T_n} r_j(s_i,a_i)\right)_{1\leq j \leq K} \in \mathbb{R}^K.
\end{equation}

When dealing with multi-objective reward functions, it is important to track the different objectives individually when evaluating a policy.  This allows for a more clear understanding of the different trade-offs the policy is making and choose which compromises they consider best.

To evaluate the performance of the algorithm across the full distribution of scenarios (e.g. users, tasks, robots, objects,etc.), we suggest independently analyzing the performance of the algorithm on each cohort. This is also important for ensuring fairness of an algorithm when interacting with populations of users. Another approach is to analyze the CVaR return rather than expected returns, or to directly determine whether rare catastrophic rewards are minimized \citep{Tamar2015,tamar2015policy}. Another evaluation procedure is to observe behavioural changes when an agent needs to be risk-averse or risk-seeking such as in football~\citep{Mankowitz2016}.

\paragraph{Experimental Setup \& Results}
We illustrate the multi-objective challenge by looking at the effects of a multi-objective reward function that encourages both task success and the satisfaction of safety constraints specified in Section \ref{sec:constraints}.  We use a naive mixture reward:
\begin{equation}
\label{eq:mixture}
r_m = (1-\alpha) r_b + \alpha r_c,
\end{equation} where $r_b$ is the task's base reward, $r_c$ is the number of satisfied constraints during that timestep and $\alpha \in[0,1]$ is the multi-objective coefficient that balances between the objectives.

\begin{figure}[h]
% \begin{subfigure}[b]{width=\textwidth}
%  \centering
  \begin{subfigure}{\textwidth}
  \centering
  \includegraphics[width=0.49\textwidth]{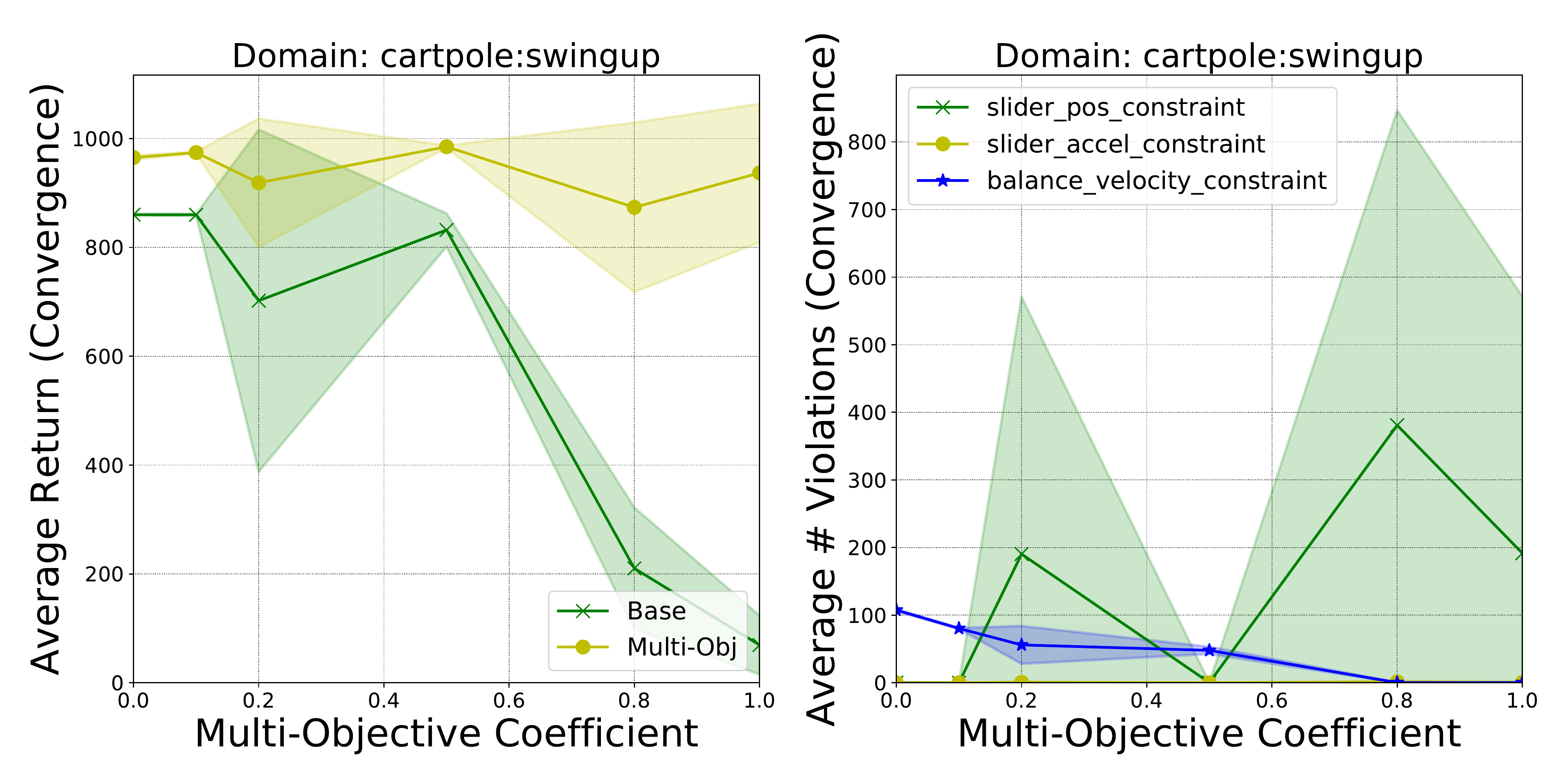}
  \includegraphics[width=0.49\textwidth]{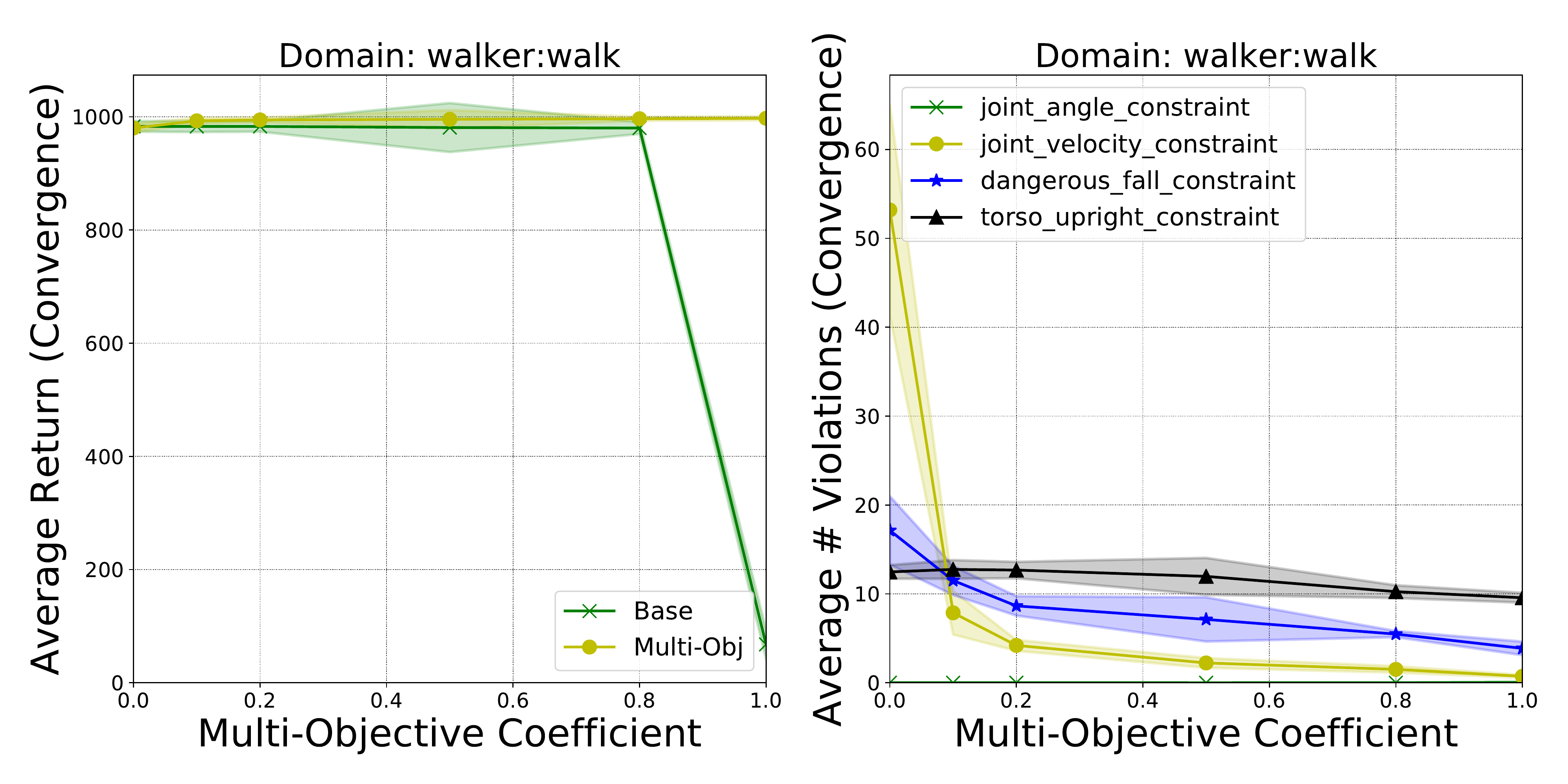}
  \includegraphics[width=0.49\textwidth]{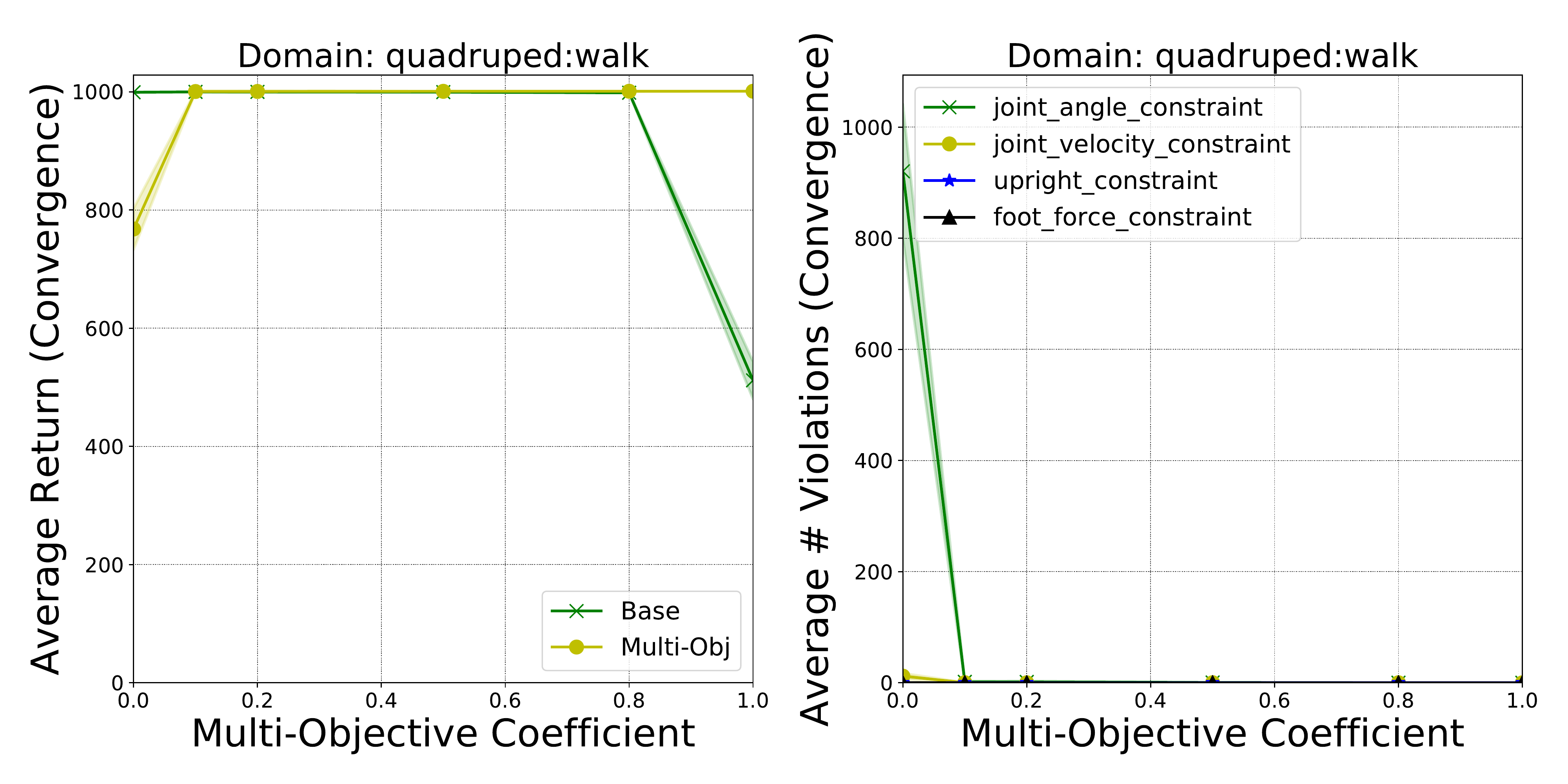}
  \includegraphics[width=0.49\textwidth]{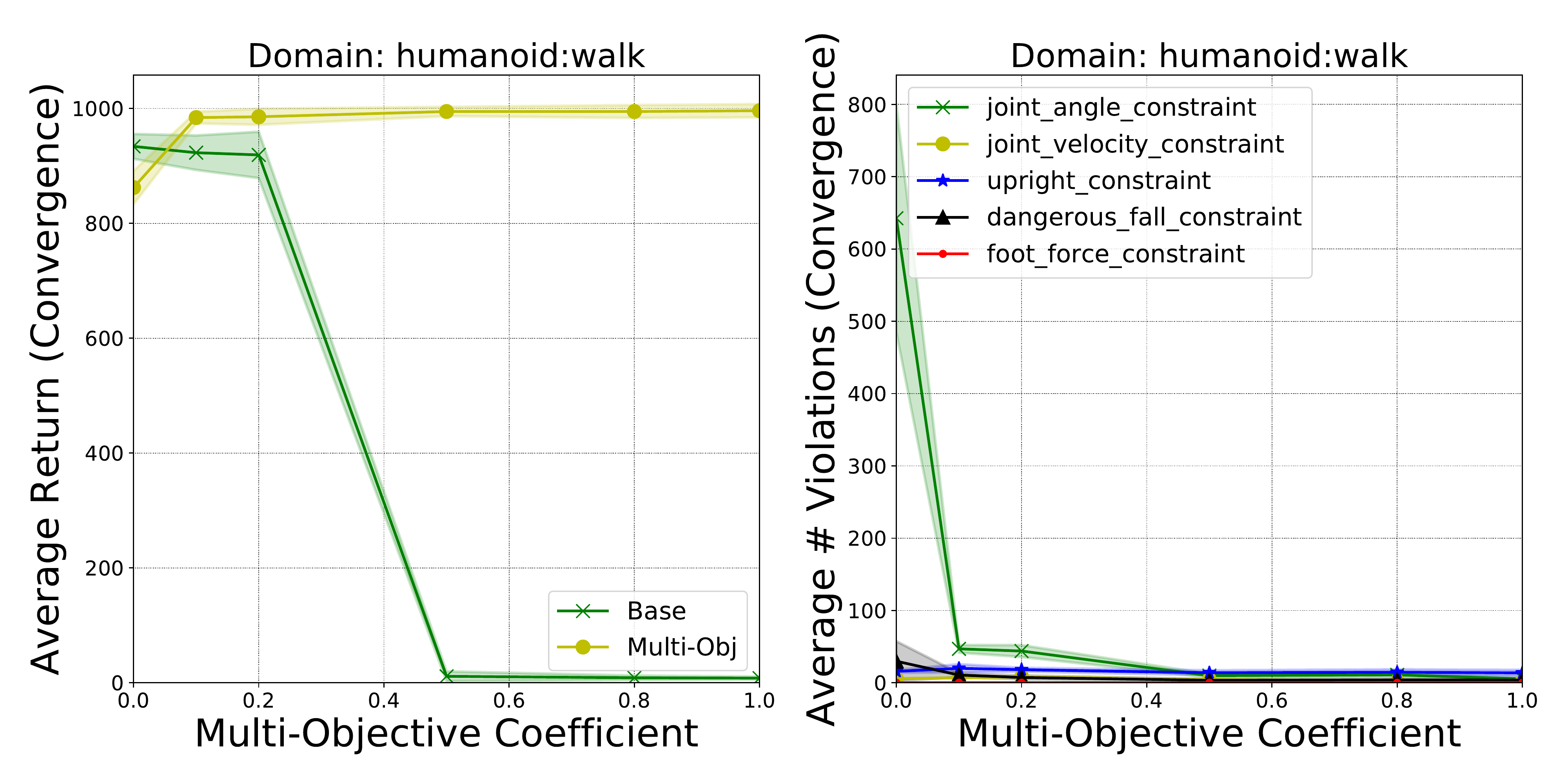}
  \caption{Performance vs. constraint satisfaction trade-offs as $\alpha$, the multiobjective coefficient, is varied for D4PG.}
\end{subfigure}

\begin{subfigure}{\textwidth}
 \centering
  \includegraphics[width=0.49\textwidth]{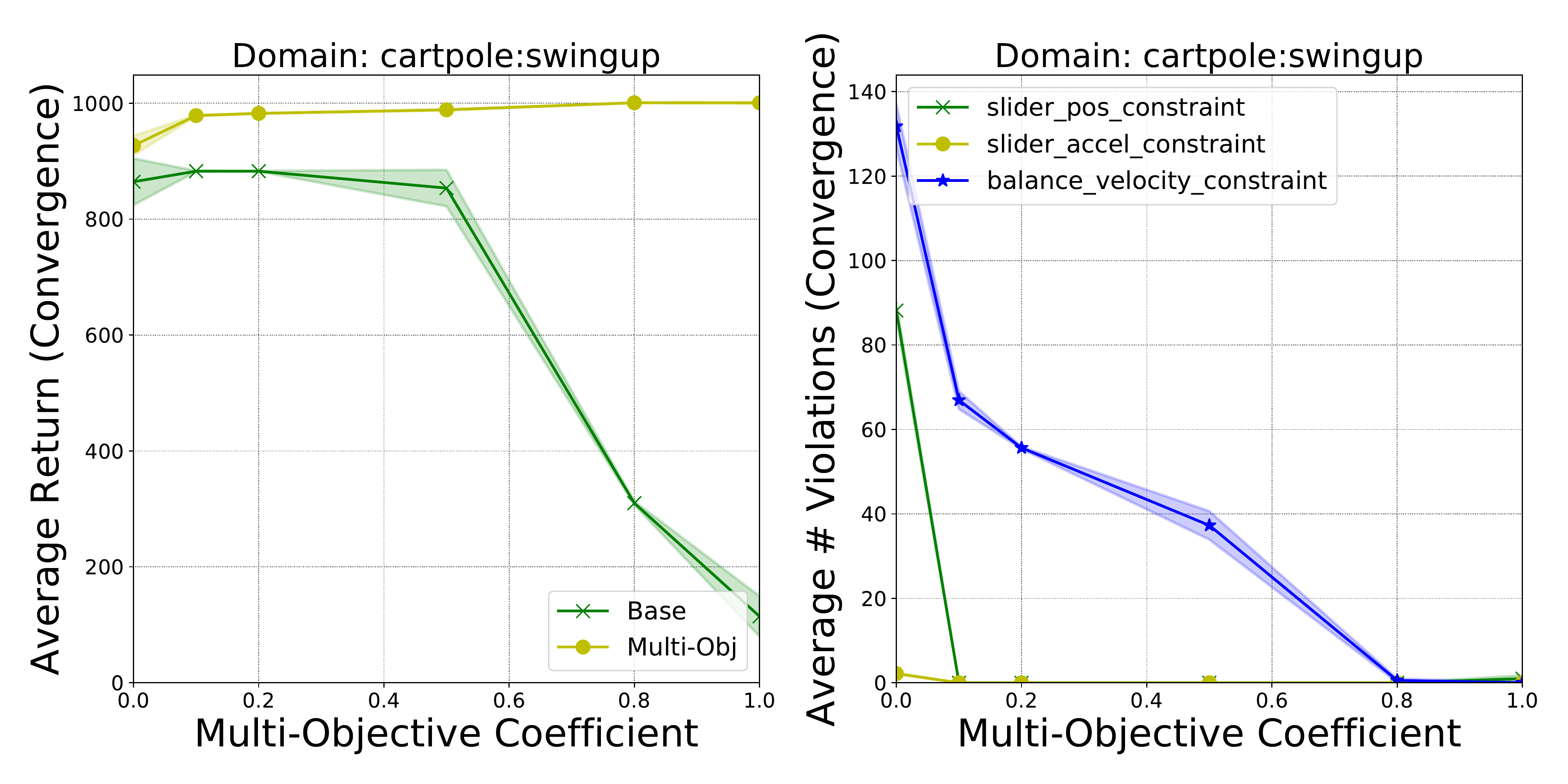}
  \includegraphics[width=0.49\textwidth]{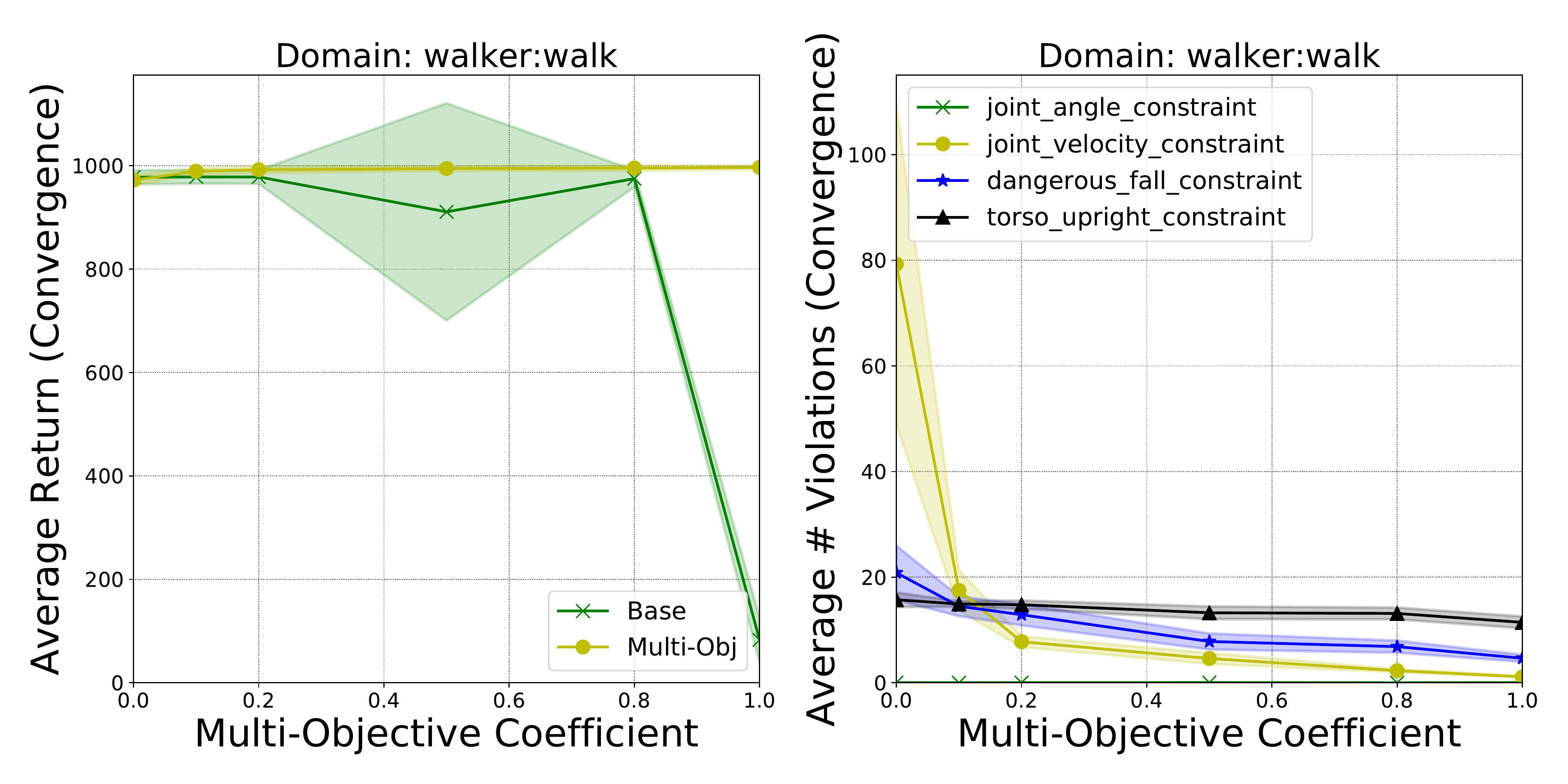}
  \includegraphics[width=0.49\textwidth]{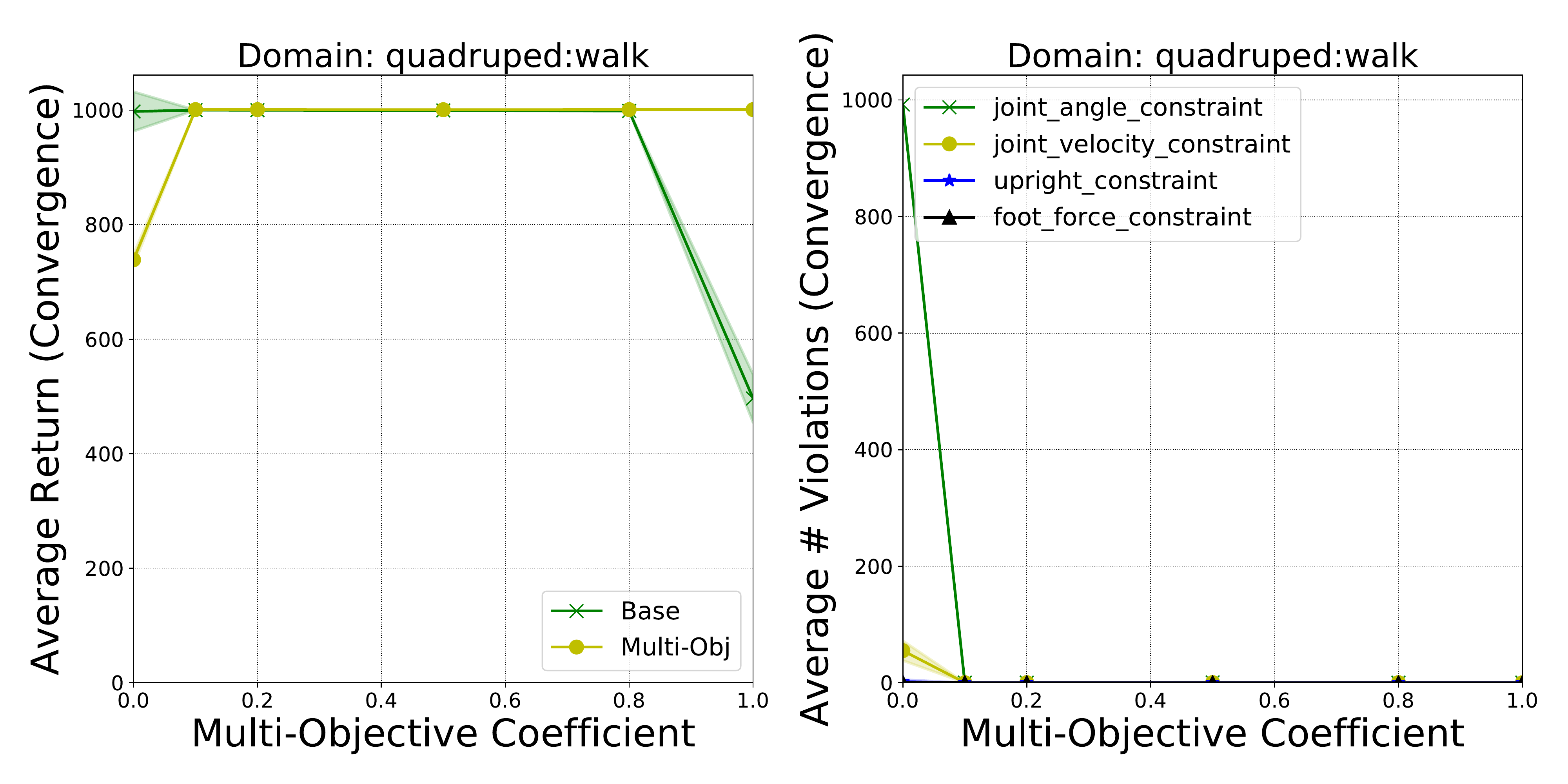}
  \includegraphics[width=0.49\textwidth]{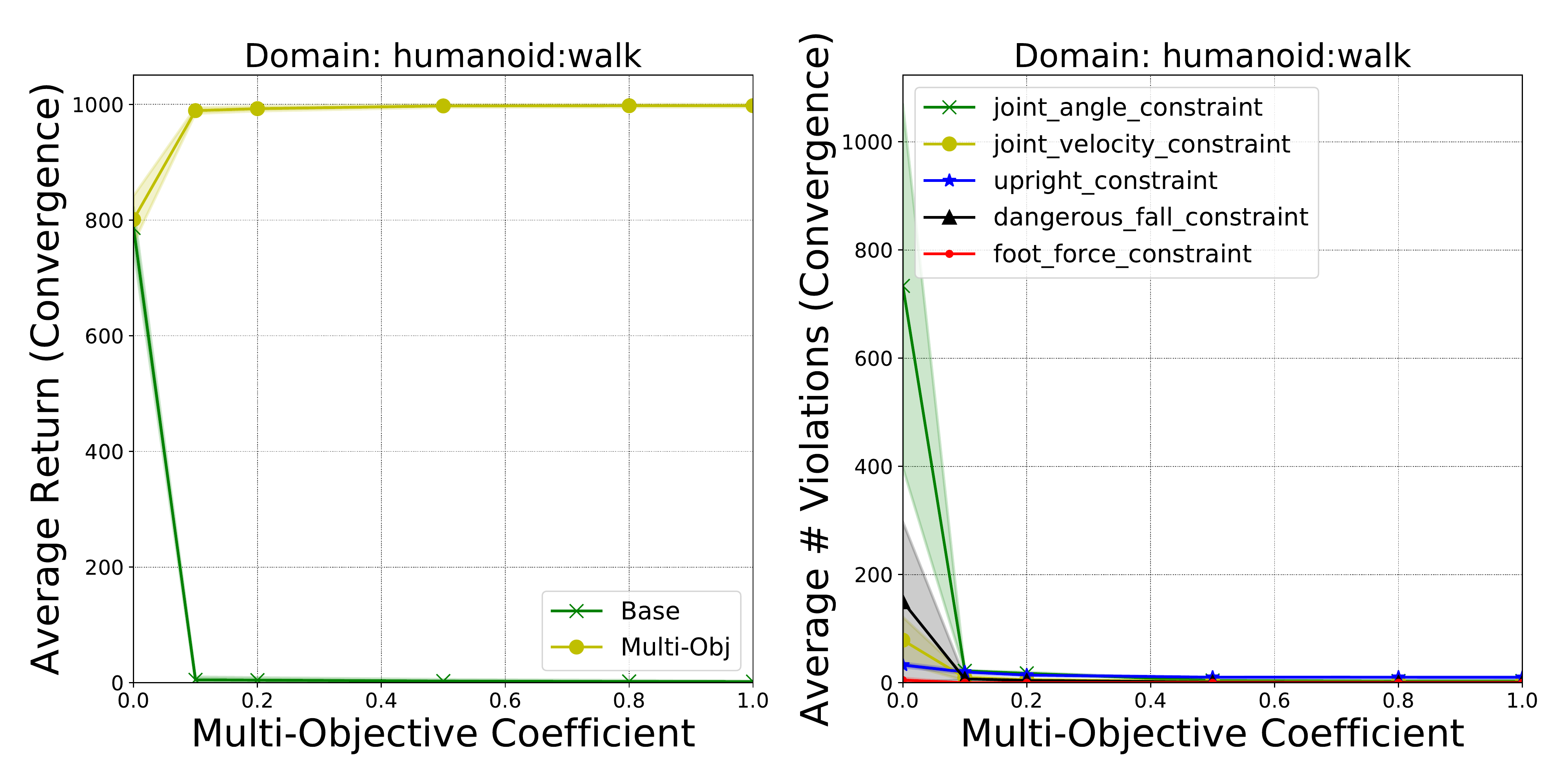}
    \caption{Performance vs. constraint satisfaction trade-offs as $\alpha$, the multiobjective coefficient, is varied for DMPO. }
\end{subfigure}
    \caption{Performance vs. constraint satisfaction trade-offs as $\alpha$, the multiobjective coefficient, is varied. The multi-objective coefficient is the reward-mixture coefficient that makes the agent's perceived reward lean more towards the original task reward or more towards the constraint satisfaction reward. For each task, the left plot shows the evolution of the tasks' original reward as the reward-mixture mixture coefficient is altered.  The right plot shows the average number of constraint violations upon convergence per episode for each individual constraint.}\label{fig:multiobj}
\end{figure}

The \suite allows multi-objective rewards to be defined, providing the multiple objectives either as observations to the agents, as modifications to the original task's reward, or both. We use the suite to model the multi-objective problem  by letting $\alpha$ correspond to the \texttt{multiobj\_coeff} in the \suite, and changing the task's reward to correspond to Equation \eqref{eq:mixture}.  For each task, we visualize both the per-element reward, as defined in Equation \eqref{eq:multiobj}, and the average number of each constraint's violations upon convergence.  Figure \ref{fig:multiobj} shows the varying effects of this multi-objective reward on each reward component, $r_b$ and $r_c$, as a function of $\texttt{multiobj\_coeff}$, where we adjust \texttt{safety\_coeff} to $0.5$ and vary \texttt{multiobj\_coeff}.  We can see the evolution in performance relative to $r_b$ and $r_c$ (left), as well as the resulting effects on constraint satisfaction (right) as \texttt{multiobj\_coeff} is varied. As $r_c$ becomes more important in the global reward, constraints are quickly taken into account.  However, over-emphasis on $r_c$ quickly degrades $r_b$ and therefore base task performance.  Although this is a naive way to deal with safety constraints, it illustrates the often contradictory goals that a real-world task might have, and the difficulty in satisfying all of them.  We also believe it provides an interesting framework to analyze how different algorithmic approaches better balance the need to satisfy constraints with the ability to maintain adequate system performance.

\clearpage
\subsection{Challenge 7: Real-time Inference Challenge}
\label{sec:real-time-inference}

%motivation for the challenge

\paragraph{Motivation \& Related Work}
To deploy RL to a production system, policy inference must be done in real-time at the control frequency of the system. This may be on the order of milliseconds for a recommender system responding to a user request~\citep{covington2016deep} or the control of a physical robot, and up to the order of minutes for building control systems~\citep{DM_Datacenter}. This constraint both limits us from running the task faster than real-time to generate massive amounts of data quickly~\citep{silver2016mastering,impala} and limits us from running slower than real-time to perform more computationally expensive approaches (e.g. some forms of model-based planning \citep{doya2002multiple, levine2019prediction, schrittwieser2019mastering}).

One approach is to take existing algorithms and validate their feasibility to run in real-time \citep{adam2011experience}. Another approach is to design algorithms with the explicit goal of running in real-time \citep{cai2017real, wang2015real}. Recently \cite{ramstedt2019real} presented a different view on real-time inference and proposed the Real-Time Markov Reward Process, in which the state evolves during an action selection. Anytime inference \citep{vlasselaer2015anytime,spirtes2001anytime} is a family of algorithms that can return a valid solution at any time they are being interrupted, and are expected to produce better performing solutions the longer they run. \cite{travnik2018reactive} propose a class of reactive SARSA RL algorithms that address the problem of asynchronous environments which occur in many real-world tasks. That is, the state is continuously changing while the agent is computing an action to take, or executing an action.

\paragraph{Experimental Setup \& Results}
The \suite offers two ways in which one can measure the effect of real-time inference: \textit{latency} and \textit{throughput}. Latency corresponds to the amount of time it takes an agent to output an action based on an observation. Even if the agent is replicated over multiple machines, allowing it to handle the frequency of the observations arriving from the system, it still may have latency issues due to the time it needs in order to output an action for a single observation.  To be able to see how a system will react in the face of latency, we use the action delay mechanism, where at time step $t$ the agent outputs an action $a_t$ based on $s_{t}$, but the system actually responds to $a_{t-n}$, where $n$ is the delay in time steps. Throughput correspond to the frequency of input observations the agent is able to process which depends on the amount of hardware or compute that is available for it as well as the complexity of the agent itself. We modeled the effects of throughput bottlenecks as action repetition: we denote the length of the action repetition by $k$, then at time step $k\cdot t$ the agent outputs an action $a_{k \cdot t}$ based on the observation $s_{k \cdot t}$, however, for the next $k-1$ time steps (i.e., time steps $k \cdot t + 1, k \cdot t + 2, .. (k+1) \cdot t - 1$), the agent repeats the same output $a_{k \cdot t}$. These two perturbations allow us to see how agents that have latency and throughput issues will affect their environment, and additionally can show us how well an agent can learn to plan accordingly to compensate for its computational shortcomings.

\begin{figure}[!htb]
    \centering
    \begin{subfigure}{.45\textwidth}
        \centering
  \includegraphics[width=\textwidth]{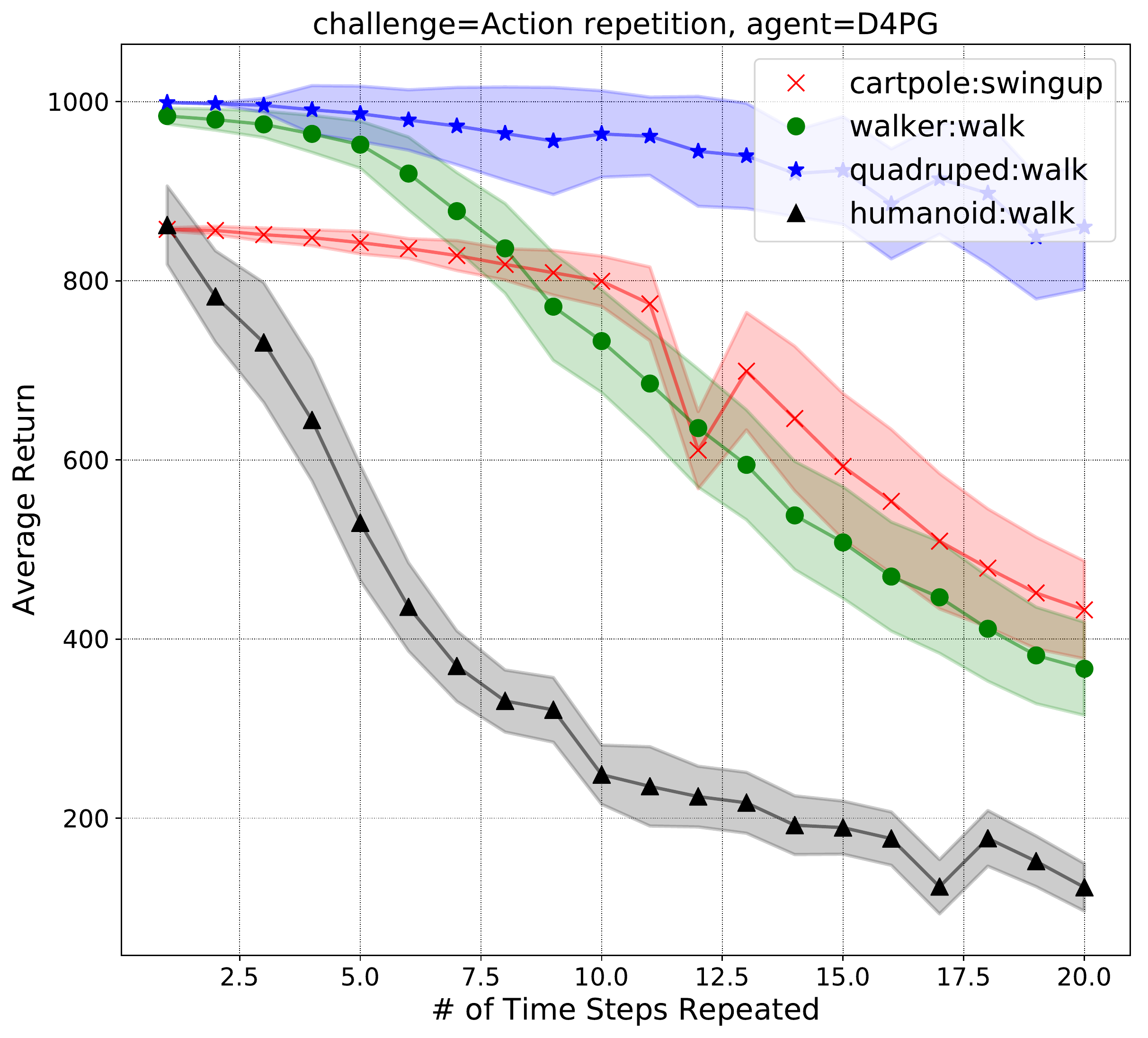}
  \caption{D4PG}\label{fig:action_repetition_d4pg}
\end{subfigure}
\begin{subfigure}{.45\textwidth}
        \centering
  \includegraphics[width=\textwidth]{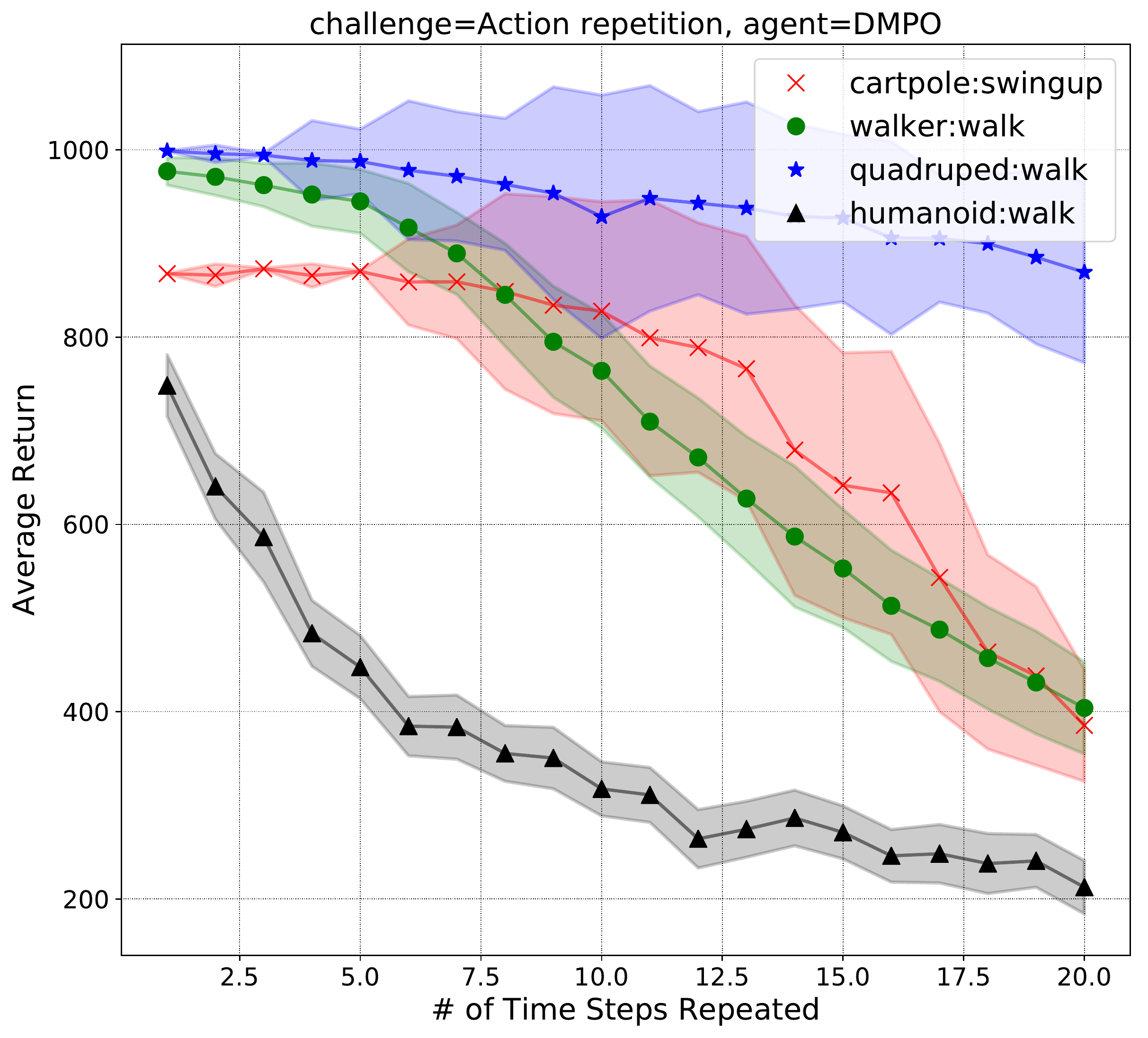}
  \caption{DMPO}\label{fig:action_repetition_dmpo}
  \end{subfigure}
  \caption{Average performance and standard deviation for D4PG (left) and DMPO (right) on the four tasks when repeating actions for a fixed number of steps.}
\end{figure}

%challenge results analysis
Figures~\ref{fig:d4pg_all_delays} and \ref{fig:dmpo_all_delays} show the performance of D4PG and DMPO, respectively, on the action delay challenge. For discussion on these results we refer the reader to Section~\ref{subsec:delays}. Figures~\ref{fig:action_repetition_d4pg} and \ref{fig:action_repetition_dmpo} shows the performance on the action repetition challenge for D4PG and DMPO, respectively. We note that generally, as expected, the performance of the agents deteriorates as the number of repeated actions increases. More interestingly though, we observe that albeit \texttt{quadruped} has larger state and action spaces than \texttt{cartpole} and \texttt{walker}, it still more robust to action repetition. We believe the reason for that lies in the inherit stability of the different tasks, where \texttt{humanoid} is the least stable, and \texttt{quadruped} is the most stable.

% \newpage
\subsection{Challenge 8: Offline Reinforcement Learning - Training from Offline Logs}

\paragraph{Motivation \& Related Work}
For many systems, learning from scratch through online interaction with the environment is too expensive or time-consuming. Therefore, it is important to design algorithms for learning good policies from offline logs of the system's behavior. In many cases these comes from an existing rule-based, heuristic or myopic policy that we are trying to replace with an RL approach. This setting is typically referred to as Offline Reinforcement Learning \footnote{Offline RL is also referred to as 'batch RL' in the literature.}. Offline and off-policy learning are closely related:

\textit{Off-policy learning} consists of a behaviour policy that generates the data and a target policy that learns from the data generated by the behaviour policy \citep{sutton2018reinforcement}. The behaviour policy continuously collects data for the agent in the environment (typically a simulator). An example of this is in deep RL where data is collected using past policies up to time $k$ during training $\pi_0, \pi_1 \cdots \pi_k$ and stored in a replay buffer. This data is then used to train the policy $\pi_{k+1}$ \citep{levine2020offline}. There are numerous examples of off-policy RL such as Q-learning \citep{sutton2018reinforcement}, Deep Q-Networks \citep{mnih2015human} as well as actor critic variants such as IMPALA \citep{espeholt2018impala}. 
\textit{Offline} RL, however, does not have the luxury of a behaviour policy that continuously interacts with the environment. In this setting, a dataset of trajectories is made available to the agent from a potentially unknown behaviour policy $\pi_B$. The dataset is collected once and is not altered during training \citep{levine2020offline}.  \\

% Reinforcement learning is already concerned with the task of learning from actions taken from a \textit{behavior} policy that is sub-optimal, or not aligned with the current policy being learned, in the context of off-policy learning \citep{sutton2018reinforcement, mnih2015human}.  However, off-policy learning nevertheless assumes that the behavior policy will eventually follow a similar state-action distribution as the policy being learnt.  In the context of off-line\footnote{Offline RL has been also called 'batch RL' in the litterature.} RL, the state-action distribution is fixed, and the learning algorithm will try to find an optimal policy that may never actually follow the given distribution (in cases where the static dataset is sub-optimal or for a different task), or follow it only once converged.

%An extension of this setup is the ``growing-batch'' setting, where a new policy is trained offline at each iteration, with the logs including new data from all the previous policies. 

Some of the early examples of offline RL include least squares temporal difference methods~\citep{bradtke1996linear,lagoudakis2003least} and fitted Q iteration~\citep{ernst2005tree, riedmiller2005neural}. 
More works such as \citet{agarwal2019striving}, \citet{fujimoto2019off}, or \citet{kumar2019stabilizing} have shown that naively applying well-known deep RL methods such as DQN~\citep{mnih2015human} in the offline setting can lead to poor performance. 
This has been attributed to a combination of poor generalization outside the training data's distribution as well as overly confident Q-function estimates when performing backups with a $\max$ operator. 
However, distributional deep RL approaches \citep{dabney18a,BellemareDM17,barthmaron2018d4pg} have been shown to produce better performance in the offline setting in both Atari \citep{agarwal2019striving} and robot manipulation \citep{cabi2019scaling}. 
There have also been a number of recent methods explicitly addressing the issues stemming from combining generalization outside the training data along with issues related to the $\max$ operator, which come in two main flavors. 
The first family of approaches constrain the action choice to the support of the training data~\citep{fujimoto2019off,kumar2019stabilizing,siegel2020keep,jaques2019way, wu2019behavior, wang2020critic}.
The second type of approaches start with behavior cloning \citep[BC;][]{pomerleau1989alvinn}, which trains a policy using the objective of predicting the action seen in the offline logs. 
Works such as \citet{wang2018exponentially}, \citet{chen2019bail}, or \citet{peng2019advantage} then use the advantage function to select the best actions in the dataset for training behavior cloning.  
Finally, model-based approaches also offer a solution to the offline setup, by training a model of the system dynamics offline and then exploiting it to solve the problem.  
Works such as MOPO~\citep{yu2020mopo} and MoREL~\citep{kidambi2020morel} leverage the learnt model to learn a model-free policy, and approaches such as MBOP~\citep{argenson2020model} leverage the model directly using an MPC-based planner. 

\paragraph{Experimental Setup \& Results}
The \suite version of the offline / batch RL challenge is to learn from data logs generated from sub-optimal policies running on the no-challenge setting, where all challenge effects are turned off, and the \textit{combined challenge} setting (see Section \ref{sec:combined_challenges}) where data logs are generated from an environment that includes effects from combining all the challenges (except for safety and multi-objective rewards). The policies were obtained by training three DMPO agents until convergence with different random weight initializations, and then taking snapshots corresponding to roughly $75\%$ of the converged performance. For the \textit{no challenge} setting, we generated three datasets of different sizes for each environment by combining the three snapshots, with the total dataset sizes (in numbers of episodes) provided in Table~\ref{tab:batch_rl_data}. Further, we repeated the procedure with the easy combination of the other challenges (see section \ref{sec:combined_challenges}). We chose to use the ``large data'' setting for the combined challenge to ensure the task is still solvable. The algorithms used for offline learning were an offline version of D4PG~\citep{barthmaron2018d4pg} that uses the data logs as a fixed experience replay buffer, as well as Critic Regularized Regression (CRR) \cite{wang2020critic}, which restricts the learned model to mimic the behavior policy when it has a positive advantage.
% \todd{should explain ABM more}
% \todo{Need to figure out how to cite CRR paper. Also Ziyu said the hyperparameters we use (which are the default values for the g3 code) were tuned on the control suite for CRR, and the same hyperparameters were used for D4PG as well, which doesn't seem very fair. How should we phrase this / should we mention it?}
% \todo{We don't expect D4PG to perform well on any of the domains, so I dont think it matters}
% \dan{I spoke to Ziyu and he said we can describe it as a modification of the Keep doing what worked paper and then cite it properly in the camera ready version once their paper is available on arxiv}

% \dan{Also, we should also have combined challenges, so a description here would be needed too}

\begin{figure}[!htb]
    \centering
    \begin{minipage}{.45\textwidth}
        \centering
        \includegraphics[width=\linewidth]{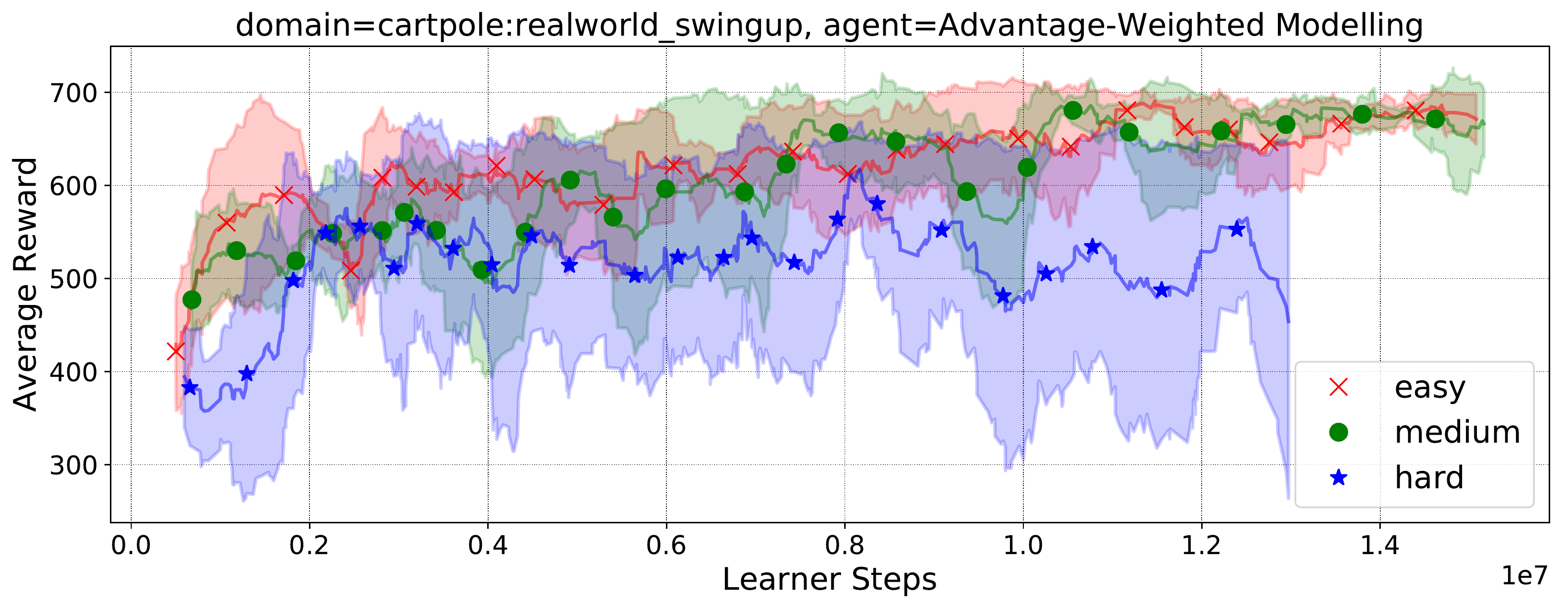}
    \end{minipage}%
    \begin{minipage}{0.45\textwidth}
        \centering
        \includegraphics[width=\linewidth]{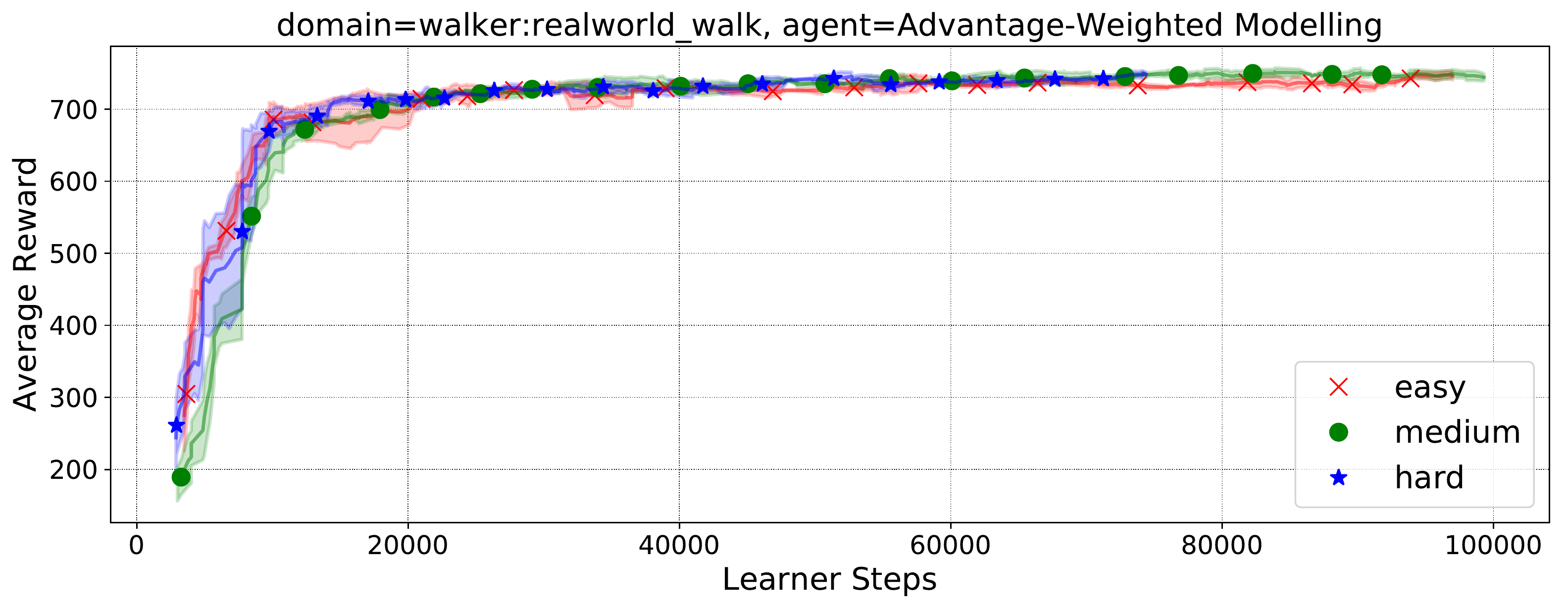}
    \end{minipage}
    \begin{minipage}{0.45\textwidth}
        \centering
        \includegraphics[width=\linewidth]{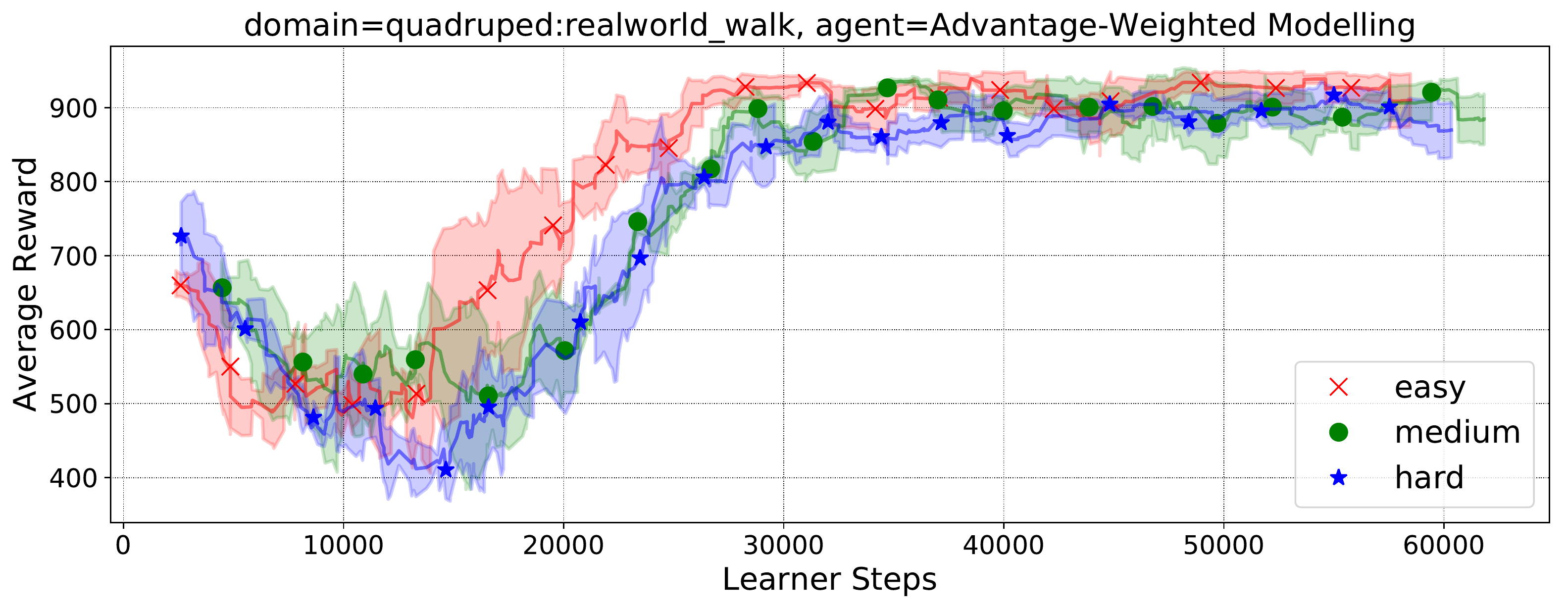}
    \end{minipage}
    \begin{minipage}{0.45\textwidth}
        \centering
        \includegraphics[width=\linewidth]{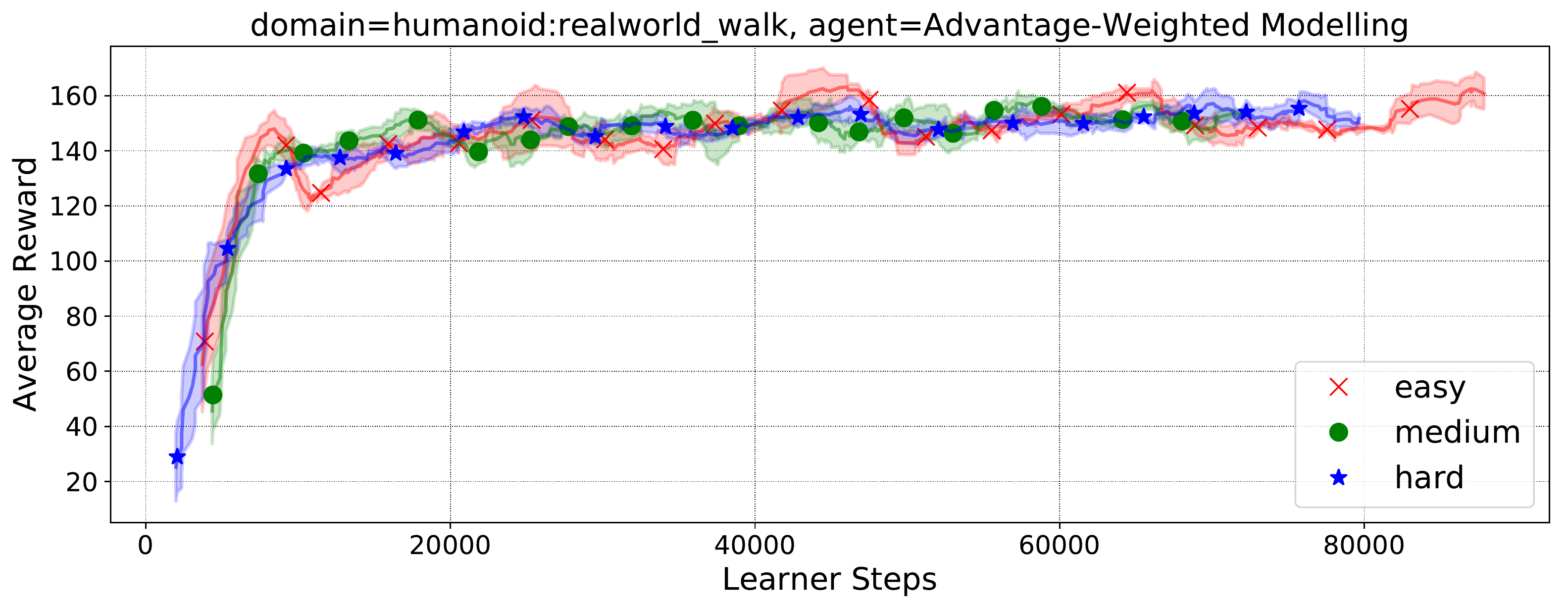}
    \end{minipage}
    \caption{Learning from offline data on small, medium and large datasets in the no challenge setting using CRR. For the cartpole domain, the X-axis is extended to show a clearer learning curve.}
    \label{fig:offline_learning}    
\end{figure}

The performance of the ABM algorithm trained on the small, medium and large batch datasets can be found in Figure \ref{fig:offline_learning} (learning curves) for each of the domains. D4PG was also trained on each of the tasks, but failed to learn in each case and therefore the results have been omitted. 
As seen in the figures, the agent fails to learn properly in the \humanoidwalk and \cartpoleswingup domain, but manages to reach a decent level of performance in \walkerwalk and \quadrupedwalk. In addition, the size of the dataset does not seem to have a significant effect on performance. This may indicate that the dataset sizes are still too large to handicap an agent's learning capabilities for a state-of-the-art offline RL agent, while being too difficult to solve for D4PG.

\begin{figure}%[!htb]
    \centering
    \begin{subfigure}{.75\textwidth}
        \centering
  \includegraphics[width=\textwidth]{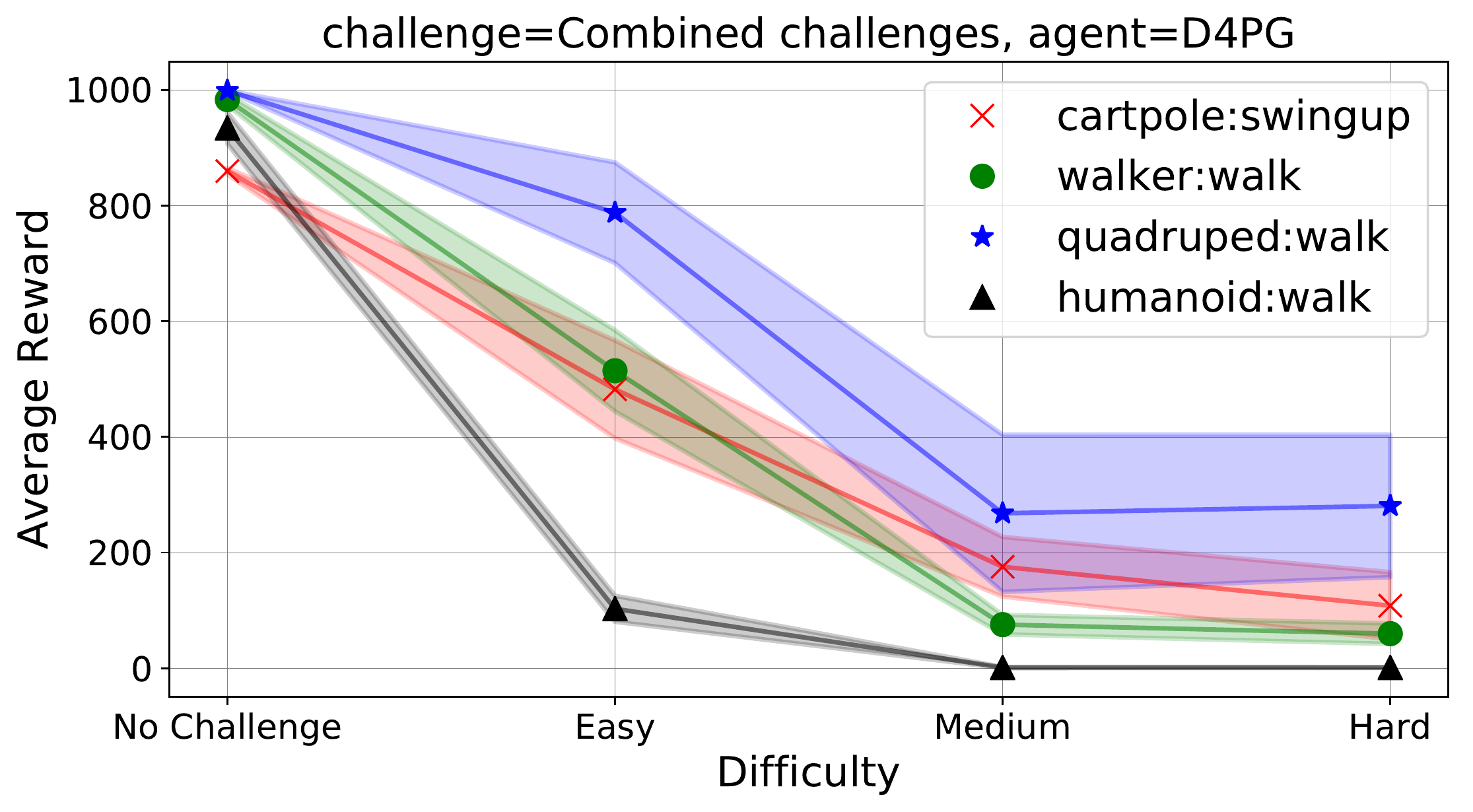}
  \caption{D4PG}\label{fig:d4pg_all_challenges_plot}
\end{subfigure}
\begin{subfigure}{.75\textwidth}
        \centering
  \includegraphics[width=\textwidth]{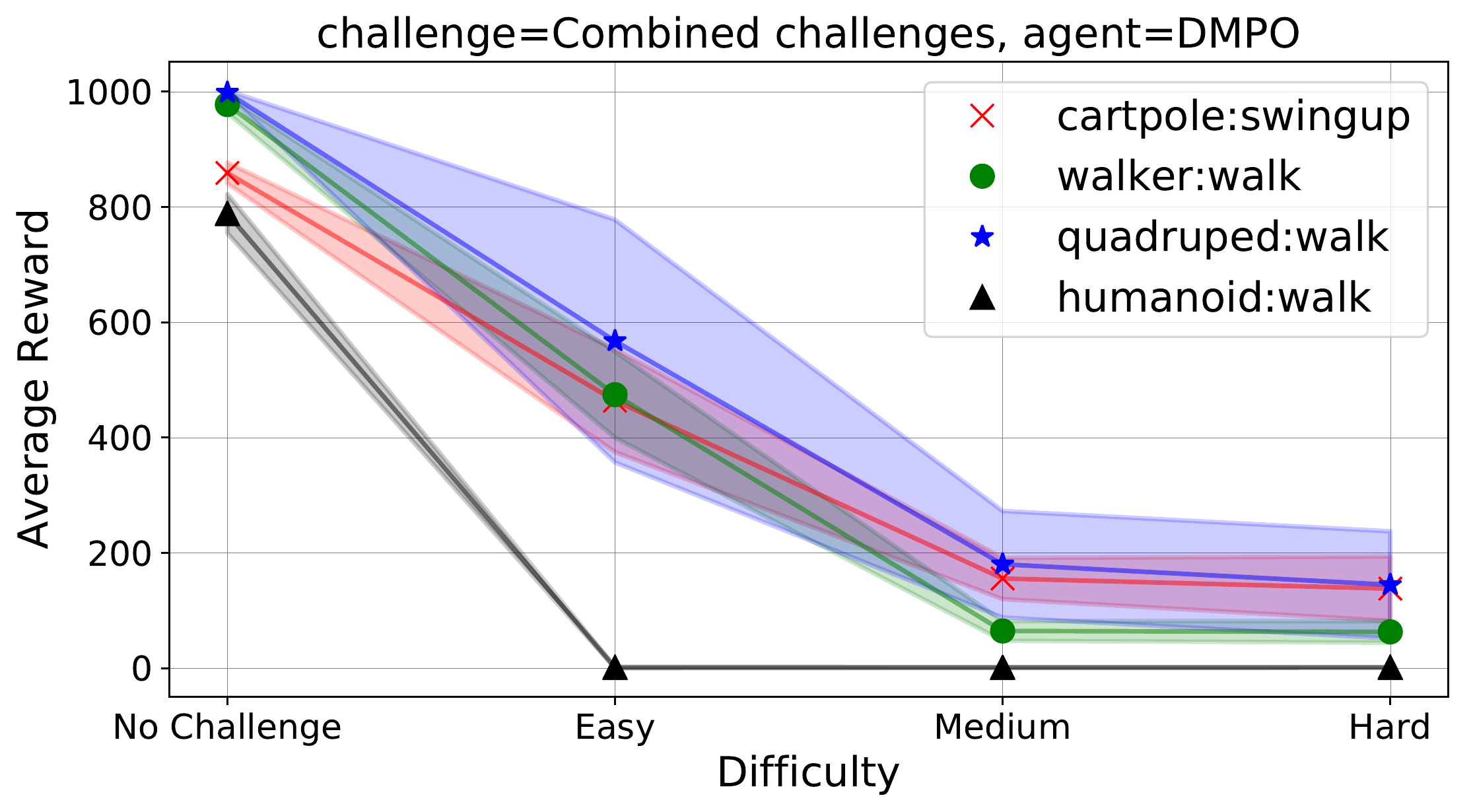}
  \caption{DMPO}\label{fig:dmpo_all_challenges_plot}
  \end{subfigure}
  \caption{D4PG and DMPO performance when incorporating all challenges into the system.}
\end{figure}

For the `Easy' combined challenge offline task, we used DMPO behaviour policies trained on each task. The \humanoidwalk DMPO behaviour policy was too poor to generate reasonable data (see Figure \ref{fig:dmpo_all_challenges_plot}) and we therefore focused on \cartpoleswingup, \walkerwalk and \quadrupedwalk for this task. This also motivates why we need to make progress on the combined challenges \textit{online} task (see Section \ref{sec:combined_challenges}) so that we can generate reasonable behaviour policies to generate the datasets for batch RL algorithms to train on. 

\begin{figure}[!htb]
    \centering
    \begin{minipage}{.65\textwidth}
        \centering
        \includegraphics[width=\linewidth]{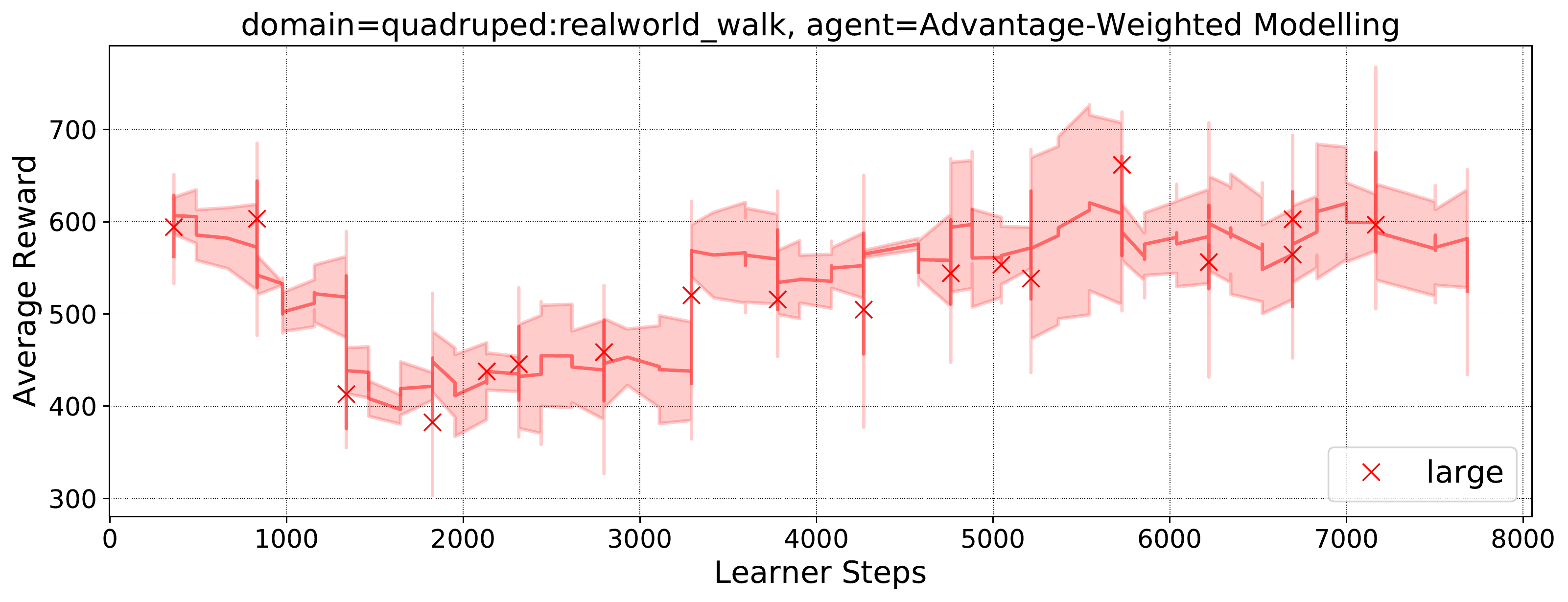}
    \end{minipage}%
    \caption{Learning from offline data on large datasets in the easy combined challenge setting using CRR on quadruped.}
    \label{fig:offline_learning_combined}    
\end{figure}

We subsequently trained CRR and D4PG (offline version) on the data generated from the behaviour policies. The agents failed to achieve any reasonable level of performance on cartpole and walker, and have thus been omitted. 
The learning curves of CRR trained on quadruped on the combined easy challenge can be found in Figure \ref{fig:offline_learning_combined}. Although the performance is still sub-optimal, it is encouraging to see that the batch agents can learn something reasonable. The D4PG offline agent failed to learn in each case and the results have therefore been omitted.

% Both cartpole and walker converged quickly, but to an inferior 300 cumulative reward per episode and thus are omitted. Humanoid is not included because DMPO cannot learn the combined easy challenge online (Figure \ref{fig:dmpo_all_challenges_plot}) and thus cannot serve as a meaningful behavior policy. 

\begin{table}[H]
\centering
\begin{tabular}{lllll}
\toprule
{} & cartpole:swingup &     walker:walk &  quadruped:walk & humanoid:walk \\
\midrule
Small Dataset  &   100  &  1000  &  100  &  4000  \\
Medium Dataset   &   200  &  2000  &  200  &  8000  \\
Large Dataset  &   500  &  5000  &  500  &  20000  \\

\bottomrule
\end{tabular}
\caption{Amount of data (number of episodes) used for different versions of the offline RL challenge. When we added the combined version of the other challenges as well, we used the "most data" version in order to keep the task solvable. We chose these numbers to be approximately four times the number of epsiodes that it takes for each agent to converge in the online setting.}
\label{tab:batch_rl_data}
\end{table}

\begin{figure}[!htb]
    \centering
  \includegraphics[width=0.6\textwidth]{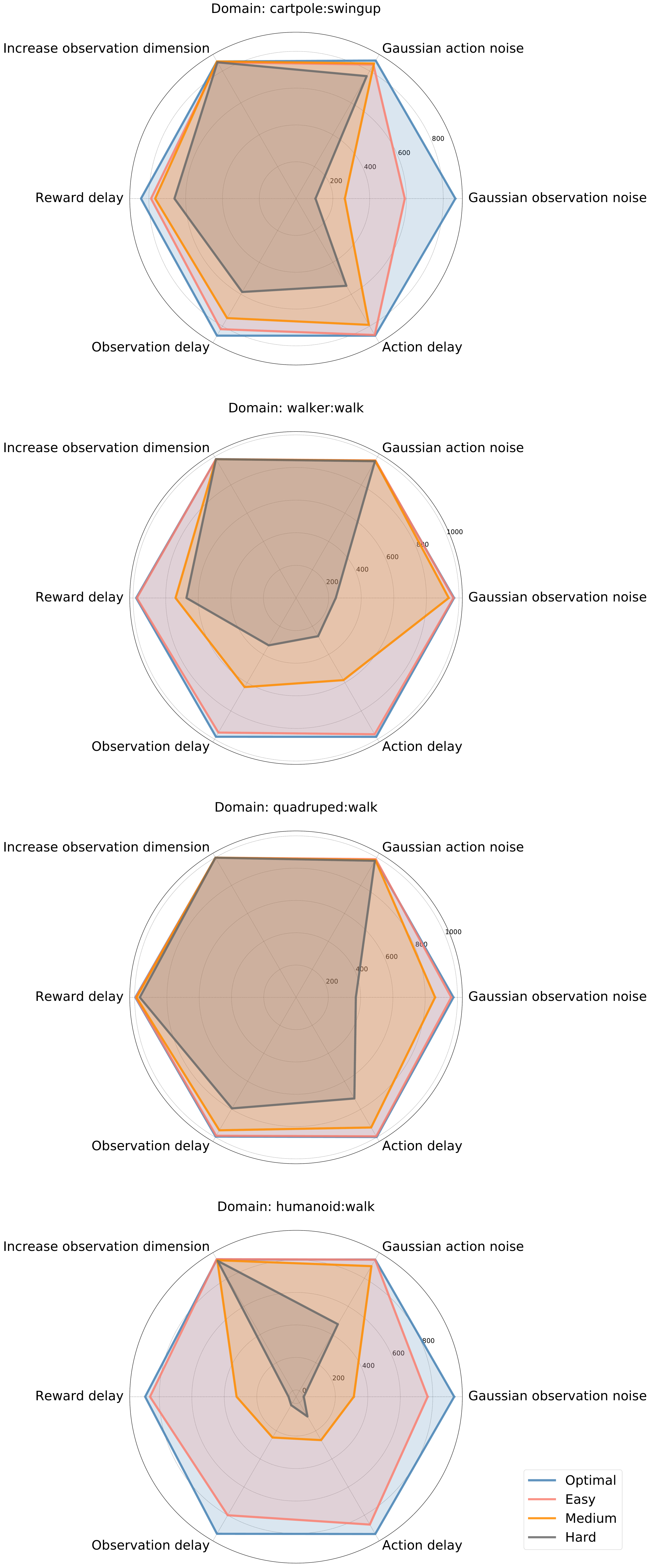}
  \caption{D4PG radar plot for the domains of \cartpoleswingup, \walkerwalk, \quadrupedwalk and \humanoidwalk respectively. The performance is measured individually for each challenge. There are four overlapping plots on each radar plot, namely optimal performance (blue) as well as Diff1 (red), Diff2 (Orange) and Diff3 (black) which corresponds to the second, third and fourth parameters (in ascending order of difficulty) for each challenge from Table \ref{app1:hyperparameters_sweeps}.}
  \label{fig:d4pg_r}
\end{figure}

\subsection{Summarizing the overall performance of an agent}

If a research or practitioner is testing out the capabilities of an agent, it would be useful to be able to summarize the performance of an agent across each challenge dimension. One such approach is to do a radar plot with respect to the various challenges. We provide an example radar plot of D4PG agent's performance on a subset of the challenges (for visualization purposes) in Figure \ref{fig:d4pg_r} for the domains of \cartpoleswingup, \walkerwalk, \quadrupedwalk and \humanoidwalk respectively. The performance is measured individually for each challenge. There are four overlapping plots on each radar plot, namely optimal performance (blue) as well as Diff1 (red), Diff2 (Orange) and Diff3 (black) which corresponds to the second, third and fourth parameters (in ascending order of difficulty) for each challenge from Table \ref{app1:hyperparameters_sweeps}. 

As you can see in the figure, D4PG struggles with the hard setting along each of the challenge dimensions, other than increased observation dimension. In addition it appears to be less sensitive to reward delay and adding Gaussian action noise on all domains except for humanoid. This kind of summary will immediately identify the weak points of an algorithm. We will make this plotting code available in the real-world RL suite open-source codebase.

\newpage
\subsection{Combining the Challenges: RWRL Benchmark}
\label{sec:combined_challenges} 

While each of these challenges present difficulties independently, many real world domains possess all of the challenges together.  To demonstrate the difficulty of learning to control a system with multiple dimensions of real-world difficulty, we combine multiple challenges described above into a set of benchmark tasks to evaluate real-world learning algorithms. Our combined challenges include parameter perturbations, additional state dimensions, observation delays, action delays, reward delays, action repetition, observation \& action noise, and stuck \& dropped sensors.  Even taking the relatively easy versions of each challenge (where the algorithm still reached close to the optimal performance individually) and combining them together creates a surprisingly difficult task. Performance on these challenges can be seen in  Table~\ref{tab:d4pg_all_challenges} for D4PG and Table~\ref{tab:dmpo_all_challenges} for DMPO, and Figures~\ref{fig:d4pg_all_challenges_plot} and \ref{fig:dmpo_all_challenges_plot} respectively.  We can see that both learners' performance drops drastically, even when applying the smallest perturbations of each challenge.

Due to both the application interest in these combined challenges, as well as their clear difficulty, we believe them to be good benchmark tasks for researchers looking to create RL algorithms for real-world systems.   We provide the parameters for each challenge in Table \ref{tab:challengehyperparams} (taken from the individual hyperparameters sweeps, see Table~\ref{app1:hyperparameters_sweeps} in the Appendix). The \suite can load the challenges directly, making it easy to replicate these benchmark environments in any experimental setup.  Although the baseline performance we provide is with a naive learner that is not designed to answer these challenges, we believe it provides a good starting point for comparison and look forward to followup work that provides more performant algorithms on these reference challenges.

% \begin{figure}[h]
%     \centering
%     \begin{minipage}{.45\textwidth}
%         \centering
%   \includegraphics[width=\textwidth]{figures/D4PG/all_challenges_D4PG.pdf}
% \end{minipage}
%     \begin{minipage}{.45\textwidth}
%         \centering
%   \includegraphics[width=\textwidth]{figures/DMPO/all_challenges_DMPO.pdf}
%   \end{minipage}
%   \caption{D4PG (left) and DMPO (right) performance when incorporating all challenges into the system.}
%   \label{fig:d4pg_dmpo_all_challenges_plot}
% \end{figure}

\begin{table}
\small
\centering
\begin{tabular}{|l|llllll}
\hline

\textbf{Experiment}                & \multicolumn{2}{l|}{\textbf{Challenge 1}}    & \multicolumn{2}{l|}{\textbf{Challenge 2}}       & \multicolumn{2}{l|}{\textbf{Challenge 3}}        \\
               & \multicolumn{2}{l|}{\textbf{(easy)}}    & \multicolumn{2}{l|}{\textbf{(Medium)}}       & \multicolumn{2}{l|}{\textbf{(Hard)}}        \\
\hline

\textbf{System Delays}             & \multicolumn{2}{l|}{\textit{Time Steps}}                 & \multicolumn{2}{l|}{\textit{Time Steps}}         & \multicolumn{2}{l|}{\textit{Time Steps}}                \\
Action                       & \multicolumn{2}{l|}{3}    & \multicolumn{2}{l|}{6}     & \multicolumn{2}{l|}{9}     \\ 
Observation                  & \multicolumn{2}{l|}{3}   & \multicolumn{2}{l|}{6}                             & \multicolumn{2}{l|}{9}    \\ 
Rewards                      & \multicolumn{2}{l|}{10}                      & \multicolumn{2}{l|}{20}   & \multicolumn{2}{l|}{40} \\ \hline
\textbf{Action Repetition}             & \multicolumn{2}{l|}{1}                 & \multicolumn{2}{l|}{2}         & \multicolumn{2}{l|}{3}                \\ 
\hline
\textbf{Gaussian Noise}                     & \multicolumn{2}{l|}{\textit{Std. Deviation}}       & \multicolumn{2}{l|}{\textit{Std. Deviation}}   & \multicolumn{2}{l|}{\textit{Std. Deviation}}     \\ 
Action              & \multicolumn{2}{l|}{0.1}     & \multicolumn{2}{l|}{0.3} & \multicolumn{2}{l|}{1.0}         \\
Observation              & \multicolumn{2}{l|}{0.1}     & \multicolumn{2}{l|}{0.3} & \multicolumn{2}{l|}{1.0}         \\ \hline
\textbf{Stuck/}                  & \multicolumn{1}{l|}{\textit{Prob.}}                    & \multicolumn{1}{l|}{\textit{Time}}       & \multicolumn{1}{l|}{\textit{Prob.}}                    & \multicolumn{1}{l|}{\textit{Time}}       & \multicolumn{1}{l|}{\textit{Prob.}}                     & \multicolumn{1}{l|}{\textit{Time}}      \\
\textbf{Dropped Noise}                  &   \multicolumn{1}{l|}{}                 &  \multicolumn{1}{l|}{\textit{steps}}    &    \multicolumn{1}{l|}{}                & \multicolumn{1}{l|}{\textit{steps}}       &     \multicolumn{1}{l|}{}                 &    \multicolumn{1}{l|}{\textit{steps}}   \\
Stuck Sensor                 & \multicolumn{1}{l|}{0.01}                    & \multicolumn{1}{l|}{1}       & \multicolumn{1}{l|}{0.05}                    & \multicolumn{1}{l|}{5}       & \multicolumn{1}{l|}{0.1}                     & \multicolumn{1}{l|}{10}      \\
Dropped Sensor               & \multicolumn{1}{l|}{0.01}                    & \multicolumn{1}{l|}{1}       & \multicolumn{1}{l|}{0.05}                    & \multicolumn{1}{l|}{5}       & \multicolumn{1}{l|}{0.1}                     & \multicolumn{1}{l|}{10}      \\ \hline
\textbf{Perturbation}                  & \multicolumn{1}{l|}{\textit{[Min,Max]}}                    & \multicolumn{1}{l|}{\textit{Std.}}       & \multicolumn{1}{l|}{\textit{[Min,Max]}}                    & \multicolumn{1}{l|}{\textit{Std.}}       & \multicolumn{1}{l|}{\textit{[Min,Max]}}                     & \multicolumn{1}{l|}{\textit{Std.}}      \\ 
\textbf{Cartpole}                  & \multicolumn{1}{l|}{}                   & \multicolumn{1}{l|}{}     &      \multicolumn{1}{l|}{}               &  \multicolumn{1}{l|}{}     &                \multicolumn{1}{l|}{}     &  \multicolumn{1}{l|}{}    \\ 
             & \multicolumn{1}{l|}{[0.9,1.1]}                       & \multicolumn{1}{l|}{0.02} & \multicolumn{1}{l|}{[0.7,1.7]}                       & \multicolumn{1}{l|}{0.1} & \multicolumn{1}{l|}{[0.5,2.3]}                       & \multicolumn{1}{l|}{0.15} \\ \hline
\textbf{Perturbation}                  & \multicolumn{1}{l|}{\textit{[Min,Max]}}                    & \multicolumn{1}{l|}{\textit{Std.}}       & \multicolumn{1}{l|}{\textit{[Min,Max]}}                    & \multicolumn{1}{l|}{\textit{Std.}}       & \multicolumn{1}{l|}{\textit{[Min,Max]}}                     & \multicolumn{1}{l|}{\textit{Std.}}      \\ 
\textbf{Quadruped}                  & \multicolumn{1}{l|}{}                   & \multicolumn{1}{l|}{}     &      \multicolumn{1}{l|}{}               &  \multicolumn{1}{l|}{}     &                \multicolumn{1}{l|}{}     &  \multicolumn{1}{l|}{}    \\ 
             & \multicolumn{1}{l|}{[0.25,0.3]}                       & \multicolumn{1}{l|}{0.005} & \multicolumn{1}{l|}{[0.25,0.8]}                       & \multicolumn{1}{l|}{0.05} & \multicolumn{1}{l|}{[0.25,1.4]}                       & \multicolumn{1}{l|}{0.1} \\ \hline
\textbf{Perturbation}                  & \multicolumn{1}{l|}{\textit{[Min,Max]}}                    & \multicolumn{1}{l|}{\textit{Std.}}       & \multicolumn{1}{l|}{\textit{[Min,Max]}}                    & \multicolumn{1}{l|}{\textit{Std.}}       & \multicolumn{1}{l|}{\textit{[Min,Max]}}                     & \multicolumn{1}{l|}{\textit{Std.}}      \\ 
\textbf{Walker}                  & \multicolumn{1}{l|}{}                   & \multicolumn{1}{l|}{}     &      \multicolumn{1}{l|}{}               &  \multicolumn{1}{l|}{}     &                \multicolumn{1}{l|}{}     &  \multicolumn{1}{l|}{}    \\ 
             & \multicolumn{1}{l|}{[0.225,0.25]}                       & \multicolumn{1}{l|}{0.002} & \multicolumn{1}{l|}{[0.225,0.4]}                       & \multicolumn{1}{l|}{0.015} & \multicolumn{1}{l|}{[0.15,0.55]]}                       & \multicolumn{1}{l|}{0.04} \\ \hline
\textbf{Perturbation}                  & \multicolumn{1}{l|}{\textit{[Min,Max]}}                    & \multicolumn{1}{l|}{\textit{Std.}}       & \multicolumn{1}{l|}{\textit{[Min,Max]}}                    & \multicolumn{1}{l|}{\textit{Std.}}       & \multicolumn{1}{l|}{\textit{[Min,Max]}}                     & \multicolumn{1}{l|}{\textit{Std.}}      \\ 
\textbf{Humanoid}                  & \multicolumn{1}{l|}{}                   & \multicolumn{1}{l|}{}     &      \multicolumn{1}{l|}{}               &  \multicolumn{1}{l|}{}     &                \multicolumn{1}{l|}{}     &  \multicolumn{1}{l|}{}    \\ 
             & \multicolumn{1}{l|}{[0.6,0.8]}                       & \multicolumn{1}{l|}{0.02} & \multicolumn{1}{l|}{[0.5,0.9]}                       & \multicolumn{1}{l|}{0.04} & \multicolumn{1}{l|}{[0.4, 1.0]}                       & \multicolumn{1}{l|}{0.06} \\ \hline
\textbf{High}                     & \multicolumn{2}{l|}{\textit{State Dimension}}       & \multicolumn{2}{l|}{\textit{State Dimension}}   & \multicolumn{2}{l|}{\textit{State Dimension}}     \\ 
\textbf{Dimensionality} & \textit{Increase}& \multicolumn{1}{l|}{}&  \textit{Increase}& \multicolumn{1}{l|}{}& \textit{Increase} & \multicolumn{1}{l|}{}\\
 & \multicolumn{2}{l|}{10}  & \multicolumn{2}{l|}{20}       & \multicolumn{2}{l|}{50}         \\ \hline
\end{tabular}
\caption{The hyperparameter setting for each combined challenge in increasing levels of difficulty}
\label{tab:challengehyperparams}
\end{table}

\begin{table}[h]
\centering
\begin{tabular}{lllll}
\toprule
{} & cartpole:swingup &     walker:walk &   quadruped:walk &   humanoid:walk \\
\midrule
       &    859.63 (5.68) &    983.24 (9.7) &    998.71 (0.32) &   934.0 (27.34) \\
Easy   &   482.32 (84.56) &  514.44 (70.21) &   787.73 (86.95) &  102.92 (22.47) \\
Medium &   175.47 (51.57) &   75.49 (16.94) &  268.01 (135.84) &     1.28 (0.99) \\
Hard   &    108.2 (57.97) &    59.85 (17.7) &  280.75 (123.21) &     1.27 (0.79) \\
\bottomrule
\end{tabular}
\caption{Mean D4PG performance ($\pm$ standard deviation) when incorporating all challenges into the system.}
\label{tab:d4pg_all_challenges}
\end{table}

\begin{table}[h]
\centering
\begin{tabular}{lllll}
\toprule
{} & cartpole:swingup &     walker:walk &   quadruped:walk &   humanoid:walk \\
\midrule
       &   859.06 (18.07) &   977.71 (14.5) &    998.35 (3.71) &  788.49 (33.88) \\
Easy   &   464.05 (89.11) &  474.44 (74.55) &  567.53 (210.54) &     1.33 (1.14) \\
Medium &   155.63 (35.81) &   64.63 (17.03) &    180.3 (92.41) &      1.27 (0.9) \\
Hard   &   138.06 (55.82) &   63.05 (18.71) &   144.69 (92.85) &      1.4 (0.82) \\
\bottomrule
\end{tabular}
\caption{Mean DMPO performance ($\pm$ standard deviation) when incorporating all challenges into the system.}
\label{tab:dmpo_all_challenges}
\end{table}

\subsection{Future Iterations}
In this paper, we have addressed 8 of the 9 challenges originally presented in~\citep{dulacarnold2019challenges}. The remaining challenge is explainability. Objectively evaluating explainability of a policy is not trivial, but we we hope this can be addressed in future iterations of this suite. We provide an overview of this challenge and possible approaches to creating explainable RL agents.

\paragraph{Explainability} Another essential aspect of real systems is that they are owned and operated by humans, who need to be reassured about the controllers' intentions and require insights regarding failure cases.
For this reason, policy explainability is important for real-world policies.
Especially in cases where the policy might find an alternative and unexpected approach to controlling a system, understanding the longer-term intent of the policy is important for obtaining stakeholder buy-in. In the event of policy errors, being able to understand the error's origins \textit{a posteriori} is essential.
%Again, we did not address explainability in this paper because it is outside the scope of modifying the tasks, but requires new experiment procedures and evaluations.
Previous work that is potentially well-suited to this challenge include options \citep{Sutton1999} that are well-defined hierarchical actions that can be composed together to solve a given task. Previous research in this area includes learning the options from scratch \citep{Mankowitz2016a,Mankowitz2016b,Bacon2017} as well as planning, given a pre-trained set of options \citep{Schaul2015,Mankowitz2018}. In addition, research has been done to develop a symbolic planning language that could be useful for explainability \citep{Konidaris2018,James2018}. 

\subsubsection{Possible additions to the nine challenges}
In addition to the nine challenges that have been defined there are a multitude of other challenges that are also up for consideration in future challenges. One such challenge is that of evolving state and action spaces. It is possible that the state space may evolve over time (e.g., adding new features to a system) as well as the action space (e.g., new capabilities are added to a robot). Instead of retraining the agent, it may be desirable to adapt the agent to the new state and action spaces. 

\subsubsection{Other challenges (e.g., infrastructure, societal etc)}
There are also other infrastructural, societal as well as problem-dependent challenges which are not in the scope of this work. This may include code modularization; how to best allocate compute when learning under a fixed resource budget; designing simple interfaces for people with limited RL knowledge such that they can solve real-world problems; how to identify when a problem is suitable for RL. All of these challenges are also preventing RL from scaling to real-world applications at an accelerated pace. We encourage researchers and practitioners to actively think about these issues as well.

\section{Additional Related Work}

 While we covered related work specific to each challenge in the sections above, there are a few other works that relate to ours, either through the goal of practical reinforcement learning or more generally by providing interesting benchmark suites.

 In general, the fact that machine learning methods have a tendency to overfit to their evaluation environments is well-recognized.  \cite{wagstaff2012machine} discusses the strong lack of real-world applications in ML conferences and the subsequent impact on research directions this can have.   \cite{henderson2018deep} investigate ways in which RL results can be made to be more reproducible and suggest guidelines for doing so.  Their paper ends by asking the question ``In what setting would [a given algorithm] be useful?'', to which we try to contribute by proposing a specific setting in which well-adapted work should hopefully stand out.
 
\citet{MLJ12-hester} similarly present a list of challenges for real world RL, but specifically for RL on robots. They present four challenges (sample efficiency, high-dimensional state and action spaces, sensor/actuator delays, and real-time inference), all of which we include in our set of challenges.  They do not include our other challenges such as satisfying constraints, multi-objective, non-stationarity and partial observability (e.g., noisy/stuck sensors). Their approach is to setup a real-time architecture for model-based learning where ensembles of models are learned to improve robustness and sample efficiency.
In a spirit similar to ours, the \texttt{bsuite} framework~\citep{Osband2019} proposes a set of challenges grounded in fundamental problems in RL such as memory, exploration, credit assignment etc. These problems are equally important and complementary to the more empirically founded challenges proposed in our suite. 
Recently, other teams have released real-world inspired environments, such as Safety Gym~\citep{Ray2019}, which extends a planar world with location-based safety constraints.  Our suite proposes a richer and more varied set of constraints, as well as an easy ability to add custom constraints,  which we believe provides a more general and difficult challenge for RL algorithms.

% \citet{rlblogpost} discusses a complementary series of difficulties of getting RL systems working on real systems.  

The Horizon platform~\citep{gauci2018horizon} and Decision Service~\citep{agarwal2016making}  provide software platforms for training, evaluation and deployment of RL agents in real-world systems. 
In the case of Decision Service, transition probabilities are logged to help make off-policy evaluation easier down the line, and both systems consider different approaches to off-policy evaluation.  We believe well-structured frameworks such as these are crucial to productionizing RL systems.
\cite{Kumar_ROBEL} propose a set of simple robot designs with corresponding simulators that have been tuned to be physically realistic, implementing safety constraints and various perturbations.

\cite{riedmiller201210} proposes a set of best practices for successfully solving typical real-world control tasks using RL. This is intended as a subjective report on how they tackle problems in practice. 

We emphasize in this work that the goal is enable RL on real-world products and systems, which may include recommender systems, physical control systems such as autonomous driving/navigation, warehouse automation etc). There are, of course, some real-world systems that have had success using RL as the algorithmic solution - mainly in robotics. For example, \citeauthor{gu2017deep} (\citeyear{gu2017deep}) perform off-policy training of deep Q functions to learn 3D manipulation skills as well as a door opening skill. \citeauthor{mahmood2018benchmarking} (\citeyear{mahmood2018benchmarking}) provide benchmarks using four off-the-shelf RL algoirthms and evaluate the performance on multiple commercially available robots. \cite{kalashnikov2018qt} introduce QT-Opt which is a self-supervised vision-based RL algoirthm that can learn a grasping skill that can generalize to unseen objects and handle perturbations. \cite{levine2016end} proposed an end-to-end learning algorithm that can map raw image observation's to torques at the robot's motors. This algorithm is able to complete a range of manipulation tasks requiring close coordination between vision and control, such as screwing a cap on a bottle. 

% Finally, \citeauthor{rlblogpost} discusses a complementary series of difficulties of getting RL systems working on real systems.
% and offer interesting directions for future research such as taking into consideration domain-specific priors, or densifying the reward signal. 

\section{Challenge Suite Overview}
Our open-sourced \suite contains: 

\begin{itemize}
\setlength\itemsep{0em}
    \item Seven real-world challenge wrappers (mentioned above) across $8$ DeepMind Control Suite tasks~\citep{tassa2018deepmind}: \\
    \texttt{cartpole:(swingup and balance)}, \texttt{walker:(walk and run)}, \\ \texttt{quadruped:(walk and run)}, \texttt{humanoid:(stand and walk)} 

    \item The flexibility to instantiate different variants of each challenge, as well as the ability to easily combine challenges together using a simple configuration language. See Appendix \ref{app:codebase} for more details.

    \item Examples of how to run RL agents on each challenge environment. 

    \item The ability to instantiate the ``Easy'', ``Medium'' and ``Hard'' combined challenges.

    \item A Jupyter notebook enabling an agent to be run on any of the challenges in a browser, as well as accompanying functions to plot the agent's performance. 

\end{itemize}

\paragraph{Evaluation environments.}  In this paper, we evaluate RL algorithms on a subset of four tasks from our suite, namely: \cartpoleswingup, \walkerwalk, \quadrupedwalk and \humanoidwalk. We chose these tasks to cover varying levels of task difficulty and dimensionality. It should be noted that \mujoco possesses an internal dynamics state and that only preprocessed observations are available to the agent~\citep{tassa2018deepmind}. We refer to state in this paper as in the typical MDP setting: the information available to the agent at time $t$. Since we provide all available observations as input to the agent, we use the term observation and state interchangeably in this paper. 
For each challenge, we have implemented environment wrappers that instantiate the challenge. These wrappers are parameterized such that the challenge can be ramped up from having no effect to being very difficult. For example, the amount of delay added onto the actuators can be set arbitrarily, varying the difficulty from slight to impossible. By implementing the challenges in this way, we can easily adapt them to other tasks and ramp them up and down to measure their effects.
Our goal with this task suite is to replicate difficulties seen in complex real systems in a more simplified setup, allowing for methodical and principled research.

\section{Discussion \& Conclusion}

To re-iterate from the Introduction, our contributions can be structured into four parts: (1) Identifying and defining a set of challenges; (2) Designing a set of experiments and analysing their effects on common RL agents; (3) Defining and benchmarking RWRL combined challenge tasks for easy algorithmic comparisons; and (4) Open-sourcing an environmental suite, \suite, which allows researchers and practitioners to easily replicate and extend the experiments we performed.

\paragraph{Identification and definition of real-world challenges} We believe that we provide a set of the most important challenges that RL algorithms need to succeed at before being ready for real-world application. In our own personal experience as well as that of our collaborators, we have been confronted ourselves numerous times with the often difficult task of applying RL to various real-world systems.  This set of challenges stems from these experiences, and we are convinced that finding solutions to them will likely provide promissing algorithms that are readily useable in real-world systems.  We are particularly interested in results in the off-line domain, as most large systems have a large amount of logs, but little to no tolerance for exploratory actions (datacenter cooling \& robotics being good examples of this). We also believe that algorithms able to reason  about environmental constraints will allow RL to move onto systems that were previously considered too fragile or expensive for learning-based approaches. Overall, we are excited about the directions that a lot of the cited research is taking and looking forward to interesting results in the near future.

\paragraph{Experiment design and analysis for each challenge}
 Additionally, the design of an experiment for each challenge demonstrates the independent effects of each challenge on an RL agent.  This allowed us to show which aspects of real-world tasks present the biggest difficulties for RL agents in a precise and reproducible manner.  
 In the case of \textit{learning on live systems from limited samples}, our proposed efficiency metrics (performance regret and stability) produced interesting findings, showing DMPO to be almost an order of magnitude worse in terms of regret, but significantly more stable once converged.  
 When \textit{dealing with system delays}, we saw that observation and action delays quickly degrade algorithm performance, but reward delays seem to be globally less impactful except on the \humanoidwalk task. 
 For \textit{high-dimensional continuous state \& action spaces}, we see that additional observation dimensions don't affect either DMPO or D4PG significantly, and that environments with more action dimensions are not necessarily harder to learn.   
 When \textit{reasoning about system constraints}, we argue that explicit reasoning about constraints is preferable to simply integrating them in the reward, and show that there is no natural way to express constraints in the standard MDP framework.  We provide a mechanism in \suite that can express constraints in the CMDP setting, and show that constraints can be violated in interesting ways, especially in tasks that have different regimes (e.g. \cartpoleswingup's `swing-up' and `balance' phases).  
 \textit{Partial observability \& non-stationarity} are often present in real systems, and can present clear problems for learning algorithms.  In small doses stuck sensors pose less of a problem than outright dropped signals however, even though the underlying information is the same. When it comes to non-stationary system dynamics, we can see that the effects depend greatly on the type of element that is varying.  Additionally, naive policies clearly degrade more quickly in the face of unstable system dynamics.
 \textit{Multi-objective rewards} can be difficult to optimize for when they are not well-aligned. By using safety-related constraints that weren't always compatible with the base task, we showed how naively reasoning about this trade-off can quickly degrade system performance, yet that compromising solutions are also possible.  We believe that expressing tasks beyond a single reward function is essential in tackling more complex problems and look forward to new methods able to do so.
 \textit{Real-time policies} are essential for high-frequency control loops present in robotics or low-latency responses necessary in software systems.  We showed the effects of both action and state delays on DMPO and D4PG, and showed that these approaches quickly degrade if the system's control frequency is higher than their response time and actions decorrelate too strongly from observations.
 Many real-world systems are hard to train on directly, and therefore RL agents need to be able to \textit{train off-line from fixed logs}.  It has long been known that this is not a trivial task, as situations that aren't represented in the data become difficult to respond to.  Especially in the case of off-policy td-learning methods, the $\argmax$ over-estimation issue quickly creates divergent value functions.  We showed that simply applying D4PG to data from a logged task is not sufficient to find a functional policy, but that offline-specific learning algorithms can deal with even small amounts of data.
 Finally, \textit{explainable policies} are often desirable (as are explainable machine learning models in general), but not easy to provide or even evaluate.  We provide a couple directions of current work in this area, and hope that future work finds clearer approach to this problem.

\paragraph{Define and baseline RWRL Combined Challenge Benchmark tasks}  By combining a well-tuned set of challenges into a single environment, we were able to generate 12 benchmark tasks (3 levels of difficulty and 4 tasks) which can serve as reference tasks for further research in real-world RL. The choice of challenge parameterizations for each level of difficulty was performed after careful analysis of the combined effects on the learning algorithms we experimented with. We also provided a first round of baselines on our benchmark tasks by running D4PG and DMPO on them: we find that D4PG seems to be slightly more robust for easy perturbations but, aside from the \quadrupedwalk task, quickly matches DMPO in poor performance.  By providing these baseline performance numbers for D4PG and DPMO on these task, we hope that followup work will have a good starting point to understand the quality of their proposed solutions. We encourage the research community to better our current set of RWRL combined challenge baseline results.

\paragraph{Open-source the \suite codebase} Finally, by implementing all our challenges in the open-sourced \suite, we provide a reference implementation for each challenge that allows easy performance comparisons between algorithms hoping to respond to these challenges.  By leveraging both the \suite and the performance baselines for each challenge presented in this paper, future researchers developing real-world RL algorithms can  easily compare their approach against common baselines to provide clear and objective evaluation.

We hope this body of works provides both encouragement to the reinforcement learning community to take up these challenges that are important holdups to bringing RL into real systems, as well as intuition to practitioners who have confronted themselves with attempting to apply RL methods on practical tasks.  We strongly believe that robust, dependable, safe, efficient, scalable RL algorithms are possible, and look forward to seeing the coming years of research in this area.

% We have presented a benchmarking suite called the \suite which is open-sourced and contains six of the nine real-world challenges presented in \citep{dulacarnold2019challenges}. These are implemented with wrappers across $8$ \mujoco tasks. We baseline performance of learning algorithms on each challenge and analyze the effect of each challenge. We also analyze the effect of combining all of the challenges together in varying levels of difficulty (``Easy'', ``Medium'', and ``Hard'' ) and present a leaderboard \todd{don't think the leaderboard was mentioned before} to baseline this performance. \todd{for these last two sentences, I would say something about what the baseline results were. Which challenges presented the most difficulty, major differences between DMPO and D4PG, and the effect of combining them} The purpose of this suite is to encourage and accelerate research progress along each of these challenge dimensions with the ultimate goal of deploying RL agents at scale on challenging real-world problems. For practitioners wishing to deploy RL to their own problems, we have pointed the reader to relevant references for each challenge that may guide them in deploying RL to a production system.  There have been many recent works on each of these challenges individually, but little work on approaches that address all nine challenges.

\bibliographystyle{abbrvnat}      % basic style, author-year citations
\bibliography{references}

\newpage
\null
% \newpage
\appendix
\section*{Appendix}

\section{Learning Algorithms}
\label{app:algorithms}

Parameters that were used for D4PG and DMPO can be found in Table~\ref{table:d4pg_hyperparameters} and Table~\ref{table:mpo_hyperparameters}, respectively.

\begin{table}[h]
\begin{center}
 \begin{tabular}{c||c} 
 Hyperparameters & D4PG \\
 \hline
 Policy net & 300-300-200 \\ 
 Number of actions sampled per state& 15\\
 Q function net & 400-400-300-100 \\
 $\sigma$ (exploration noise) & 0.1 \\
 vmim & -150 \\
 vmax & 150\\
 num atoms & 51\\
 n-step & 51\\
 Discount factor ($\gamma$) & 0.99 \\
 Adam learning rate & 0.0001 \\
 Replay buffer size & 2000000 \\
 Target network update period & 200\\
 Batch size & 512\\
 Activation function & elu\\
 Layer norm on first layer & Yes\\
 Tanh on output of layer norm & Yes\\
\end{tabular}
\end{center}
\caption{Hyperparameters for D4PG.}
\label{table:d4pg_hyperparameters}
\vspace{-0.2in}
\end{table}

\begin{table}[h]
\begin{center}
 \begin{tabular}{c||c} 
 Hyperparameters & DMPO \\
 \hline
 Policy net & 300-300-200 \\ 
 Number of actions sampled per state& 20\\
 Q function net & 400-400-300-100 \\
 $\epsilon$ & 0.1 \\
 $\epsilon_{\mu}$ & 0.005 \\
 $\epsilon_{\Sigma}$ & 0.000001\\
 Discount factor ($\gamma$) & 0.99 \\
 vmin & -150 \\
 vmax & 150 \\
 num atoms & 51 \\
 Adam learning rate & 0.0001 \\
 Replay buffer size & 1000000 \\
 Batch size & 256\\
 Activation function & elu\\
 Layer norm on first layer & Yes\\
 Tanh on output of layer norm & Yes\\
 Tanh on Gaussian mean & No \\
 Min variance & Zero \\
 Max variance & unbounded 
\end{tabular}
\end{center}
\caption{Hyperparameters for DMPO.}
\label{table:mpo_hyperparameters}
\end{table}

\newpage

% \textbf{Real-time Inference:} To deploy RL to a production system, policy inference must be done in real-time at the control frequency of the system. This may be on the order of milliseconds for a recommender system~\citep{covington2016deep} responding to a user request or the control of a physical robot, and up to the order of minutes for building control systems \citep{DM_Datacenter}. This constraint both limits us from running the task faster than real-time to generate massive amounts of data quickly~\citep{silver2016mastering,impala} and limits us from running slower than real-time to perform more computationally expensive approaches (e.g. some forms of model-based planning). We have not addressed this challenge in this paper because it also involves modifying the underlying structure of how the agent interacts with the environment, but we hope to address it in future work. 

\newpage

\section{Parameters}
\label{app:parameters}

The hyperparameters that were used for the individual challenges sweeps can be found in Table~\ref{app1:hyperparameters_sweeps}.

\begin{table}[h]
\centering
\begin{tabular}{|l|l|l|}
\hline
\textbf{Experiment}        & \multicolumn{2}{c|}{\textbf{Hyperparameter Sweep}}                     \\ \hline
\textbf{System Delays}     & \textbf{Delay (in timesteps)}      &                                   \\ \hline
Action Delay               & 0,3,6,9,12,15,18,20                &                                   \\ \hline
Observation Delay          & 0,3,6,9,12,15,18,20                &                                   \\ \hline
Rewards Delay              & 10,20,40,50,75,100            &                                   \\ \hline
\textbf{Noise}             & \textbf{Std. Deviation}            &                                   \\ \hline
Gaussian Action Noise      & 0.0,0.1,0.3,1.0,1.3,2.0,2.3        &                                   \\ \hline
Gaussian Observation Noise & 0.0,0.1,0.3,1.0,1.3,2.0,2.3        &                                   \\ \hline
Action Repetition Noise & 1,2,3,5,7,10,13,16,20        &                                   \\ \hline
                           & \textbf{Stuck/Dropped Probability} & \textbf{Stuck/Dropped steps}      \\ \hline
Stuck Sensor Noise         & 0.0,0.01,0.05,0.1,0.3,0.5,0.7      & 0,1,5,10,20,50                    \\ \hline
Dropped Sensor Noise       & 0.0,0.01,0.05,0.1,0.3,0.5,0.7      & 0,1,5,10,20,50                    \\ \hline
\textbf{}                  & \textbf{Perturbation Frequency}    & \textbf{Perturbation Schedule}    \\ \hline
\textbf{Perturbations}     & 1,2,5,10,50,100                    & uniform,cyclic\_pos               \\ \hline
\textbf{}                  & \textbf{State Dimension Increase}    &     \\ \hline
\textbf{High Dimensionality}     & 0,10,20,50,100                    &                \\ \hline
                           & \textbf{Safety Coefficient}        & \textbf{Safety Penalty Weighting} \\ \hline
\textbf{Safety}            & 1.0,0.8,0.5,0.2,0.1                               & N/A                              \\ \hline
\textbf{Multi-objective}            &     0.5                          & 1,0.8,0.5,0.2,0.1,0                              \\ \hline
\end{tabular}
\caption{Hyperparameter sweeps for each challenge experiment}
\label{app1:hyperparameters_sweeps}
\end{table}

\section{Codebase}
\label{app:codebase}

\subsection{Specifying Challenges}
\label{app:specifying_challenges}
Specifying task challenges is done by passing arguments to the \texttt{load} method of the environment (see examples in Appendix~\ref{appendix:code_snippets}). Comprehensive documentation is available in the codebase itself, however, for completeness we list the different arguments here.

\begin{itemize}
    \item \textbf{Constraints}
    \begin{itemize}
        \item \textit{Description}: Adds a set of constraints on the task. Returns an additional entry in the observations ('constraints') in the length of the number of the contraints, where each entry is True if the constraint is satisfied and False otherwise. In our implementation we used safety constraints as the constraints. The safety constraints per domain can be found in Table~\ref{tab:safe_envs_full}.
        \item \textit{Input argument}: \texttt{safety\_spec}, a dictionary that specifies the safety constraints specifications of the task. It may contain the following fields:
        \begin{itemize}
            \item \texttt{enable}, a boolean that represents whether safety specifications are enabled.
            \item \texttt{constraints}, a list of class methods returning boolean constraint satisfactions (default ones are provided).
            \item \texttt{limits}, a dictionary of constants used by the functions in 'constraints' (default ones are provided).
            \item \texttt{safety$\_$coeff}, a scalar between 1 and 0 that scales safety constraints, 1 producing the base constraints, and 0 likely producing an unsolveable task.
            \item \texttt{observations}, a default-True boolean that toggles the whether a vector of satisfied constraints is added to observations.
        \end{itemize}
    \end{itemize}
    \newpage
    \item \textbf{Delays}
    \begin{itemize}
        \item \textit{Description}: Adds actions, observations and rewards delays. Actions delay is the number of steps between passing the action to the environment when it is actually performed, and observation (reward) delay is the offset of freshness of the returned observation (reward) after performing a step.
        \item \textit{Input argument}: \texttt{delay\_spec}, a dictionary that specifies the delay specifications of the
        task. It may contain the following fields:
        \begin{itemize}
            \item \texttt{enable}, a boolean that represents whether delay specifications are enabled.
            \item \texttt{actions}, an integer indicating the number of steps actions are being delayed.
            \item \texttt{observations}, an integer indicating the number of steps observations are being delayed.
            \item \texttt{rewards}, an integer indicating the number of steps rewards are being delayed.
        \end{itemize}
    \end{itemize}
    
    \item \textbf{Noise}
    \begin{itemize}
        \item \textit{Description}: Adds action or observation noise. Different noise include: white Gaussian actions/observations, dropped actions/observations values, stuck actions/observations values, or repetitive actions.
        \item \textit{Input argument}: \texttt{noise\_spec}, a dictionary that specifies the noise specifications of the
        task. It may contains the following fields:
        \begin{itemize}
            \item \texttt{gaussian}, a dictionary that specifies the white Gaussian additive noise. It may contain the following fields:
            \begin{itemize}
                \item \texttt{enable}, a boolean that represents whether noise specifications are enabled.
                \item \texttt{actions}, a float indicating the standard deviation of a white Gaussian noise added to each action.
                \item \texttt{observations}, similarly, additive white Gaussian noise to each returned observation.
            \end{itemize}
            \item \texttt{dropped}, a dictionary that specifies the dropped values noise. It may contain the following fields:
            \begin{itemize}
                \item \texttt{enable}, a boolean that represents whether dropped values specifications are enabled.
                \item \texttt{observations$\_$prob}, a float in [0,1] indicating the probability of dropping each observation component independently.
                \item \texttt{observations$\_$steps}, a positive integer indicating the number of time steps of dropping a value (setting to zero) if dropped.
                \item \texttt{actions$\_$prob}, a float in [0,1] indicating the probability of dropping each action component independently.
                \item \texttt{actions$\_$steps}, a positive integer indicating the number of time steps of dropping a value (setting to zero) if dropped.
            \end{itemize}
            \item \texttt{stuck}, a dictionary that specifies the stuck values noise. It may contain the following fields:
            \begin{itemize}
                \item \texttt{enable}, a boolean that represents whether stuck values specifications are enabled.
                \item \texttt{observations$\_$prob}, a float in [0,1] indicating the probability of each observation component becoming stuck.
                \item \texttt{observations$\_$steps}, a positive integer indicating the number of time steps an observation (or components of) stays stuck.
                \item \texttt{actions$\_$prob}, a float in [0,1] indicating the probability of each action component becoming stuck.
                \item \texttt{actions$\_$steps}, a positive integer indicating the number of time steps an action (or components of) stays stuck.
            \end{itemize}
            \item \texttt{repetition}, a dictionary that specifies the repetition statistics. It may contain the following fields:
            \begin{itemize}
                \item \texttt{enable}, a boolean that represents whether repetition specifications are enabled.
                \item \texttt{actions$\_$prob}, a float in [0,1] indicating the probability of the actions to be repeated in the following steps.
                \item \texttt{actions$\_$steps}, a positive integer indicating the number of time steps of repeating the same action if it to be repeated.
            \end{itemize}
        \end{itemize}
    \end{itemize}
        
    \newpage
    \item \textbf{Perturbations}
    \begin{itemize}
        \item \textit{Description}: Perturbs physical quantities of the environment. These perturbations are non-stationary and are governed by a scheduler.
        \item \textit{Input argument}: \texttt{perturb\_spec}, a dictionary that specifies the perturbation specifications of the task. It may contain the following fields:
        \begin{itemize}
            \item \texttt{enable}, a boolean that represents whether perturbation specifications are enabled.
            \item \texttt{frequency}, an integer, number of episodes between updates perturbation updates.
            \item \texttt{param}, a string indicating which parameter to perturb (supporting multiple parameters, environment-dependent, see Table~\ref{tab:perturb_env_full}).
            \item \texttt{scheduler}, a string indicating the scheduler to apply to the perturbed parameter. Currently supporting:
            \begin{itemize}
                \item constant - constant value determined by the `start` argument.
                \item random$\_$walk - random walk governed by a white Gaussian process.
                \item drift$\_$pos - uni-directional (increasing) random walk which saturates. 
                \item drift$\_$neg - uni-directional (decreasing) random walk which saturates. 
                \item cyclic$\_$pos - uni-directional (increasing) random walk which resets once reaching the maximal value. 
                \item cyclic$\_$neg - uni-directional (decreasing) random walk which resets once reaching the minimal value.
                \item uniform - uniform sampling process within a bounded support.
                \item saw$\_$wave - alternating uni-directional random walks between minimal and maximal values.
            \end{itemize}
            \item \texttt{start}, a float indicating the initial value of the perturbed parameter.
            \item \texttt{min}, a float indicating the minimal value the perturbed parameter may be.
            \item \texttt{max}, a float indicating the maximal value the perturbed parameter may be.
            \item \texttt{std}, a float indicating the standard deviation of the white noise for the scheduling process.
        \end{itemize}
    \end{itemize}
    
    \item \textbf{Dimensionality}
    \begin{itemize}
        \item \textit{Description}: Adds extra dummy features to observations to increase dimensionality of the state space.
        \item \textit{Input argument}: \texttt{dimensionality\_spec}, a dictionary that specifies the added dimensions to the
        state space. It may contain the following fields:
        \begin{itemize}
            \item \texttt{num$\_$random$\_$state$\_$observations}, an integer indicating the number of random observations to add (defaults to zero).
        \end{itemize}
    \end{itemize}    
    
    \item \textbf{Multi-Objective Reward}
    \begin{itemize}
        \item \textit{Description}: Provides a reward that gets added onto the base reward and re-normalized to [0,1].
        \item \textit{Input argument}: \texttt{multiobj\_spec}, a dictionary that sets up the multi-objective challenge. The challenge works by providing an `Objective` object which describes both numerical objectives and a reward-merging method that allow to both observe the various objectives in the observation and affect the returned reward in a manner defined by the Objective object.
        \begin{itemize}
            \item \texttt{objective}, either a string which will load an `Objective` class from \\ utils.multiobj\_objectives.Objective, or an Objective object which subclasses it.
            \item \texttt{reward}, a boolean indicating whether to add the multiobj objective's reward to the environment's returned reward.
            \item \texttt{coeff}, a float in [0,1] that is passed into the Objective object to change the mix between the original reward and the Objective's rewards.
            \item \texttt{observed}, a boolean indicating whether the defined objectives should be added to the observation.
        \end{itemize}
    \end{itemize}

\end{itemize}

\newpage

\subsection{Code Snippets}\label{appendix:code_snippets}

Below is an example of using the OpenAI PPO baseline with our suite.

\begin{lstlisting}[language=Python]
from baselines import bench
from baselines.common.vec_env import dummy_vec_env
from baselines.ppo2 import ppo2
import example_helpers as helpers
import realworldrl_suite.environments as rwrl


def _load_env():
  """Loads environment."""
  raw_env = rwrl.load(
      domain_name='cartpole',
      task_name='realworld_swingup',
      safety_spec=dict(enable=True),
      delay_spec=dict(enable=True, actions=20),
      log_output='/tmp/path/to/results.npz',
      environment_kwargs=dict(log_safety_vars=True, flat_observation=True))
  env = helpers.GymEnv(raw_env)
  env = bench.Monitor(env, FLAGS.save_path)
  return env

env = dummy_vec_env.DummyVecEnv([_load_env])
ppo2.learn(env=env, network='mlp', lr=1e-3, total_timesteps=1000000,
           nsteps=100, gamma=.99)
\end{lstlisting}

Below is another example running a random policy.

\begin{lstlisting}[language=Python]
import numpy as np
import realworldrl_suite.environments as rwrl


def random_policy(action_spec):
  def _act(timestep):
    del timestep
    return np.random.uniform(low=action_spec.minimum,
                             high=action_spec.maximum,
                             size=action_spec.shape)
  return _act


env = rwrl.load(
    domain_name='cartpole',
    task_name='realworld_swingup',
    safety_spec=dict(enable=True),
    delay_spec=dict(enable=True, actions=20),
    log_output='/tmp/path/to/results.npz',
    environment_kwargs=dict(log_safety_vars=True, flat_observation=True))

policy = random_policy(action_spec=env.action_spec())

rewards = []
total_episodes = 100
for _ in range(total_episodes):
  timestep = env.reset()
  total_reward = 0.
  while not timestep.last():
    action = policy(timestep)
    timestep = env.step(action)
    total_reward += timestep.reward
  rewards.append(total_reward)
print('Random policy total reward per episode: {:.2f} +- {:.2f}'.format(
    np.mean(rewards), np.std(rewards)))
\end{lstlisting}

Below is an example of instantiating an environment with the `easy` challenge 

\begin{lstlisting}[language=Python]
import realworldrl_suite.environments as rwrl


env = rwrl.load(
    domain_name='cartpole',
    task_name='realworld_swingup',
    combined_challenge='easy',
    log_output='/tmp/path/to/results.npz',
    environment_kwargs=dict(log_safety_vars=True, flat_observation=True))
\end{lstlisting}

\end{document}